\documentclass{article}

\usepackage[final,sglblindworkshop]{neurips_2026}
\workshoptitle{AI for Accelerated Materials Design (AI4Mat)}
\makeatletter
\renewcommand{\@noticestring}{An earlier version of this manuscript was submitted to NeurIPS AI4Mat in August 2026.}
\makeatother

\usepackage[utf8]{inputenc}
\usepackage[T1]{fontenc}
\usepackage{amsmath}
\usepackage{amssymb}
\usepackage{booktabs}
\usepackage{tabularx}
\usepackage{longtable}
\usepackage{graphicx}
\usepackage{float}
\usepackage{placeins}
\usepackage{microtype}
\usepackage{xcolor}
\usepackage{colortbl}
\usepackage{caption}
\usepackage{hyperref}
\usepackage{url}

\newif\ifshowcomments
\showcommentstrue       % working PDF
\ifshowcomments
  \newcommand{\austin}[1]{\textcolor{red}{[Austin: #1]}}
  \newcommand{\jiaru}[1]{\textcolor{blue}{[Jiaru: #1]}}
\else
  \newcommand{\austin}[1]{}
  \newcommand{\jiaru}[1]{}
\fi

\definecolor{solblue}{HTML}{0072B2}
\definecolor{lunaorange}{HTML}{D55E00}
\definecolor{adaptedgreen}{HTML}{009E73}
\newcommand{\solhit}{\textcolor{solblue}{\raisebox{0.15ex}{\rule{0.72ex}{0.72ex}}}}
\newcommand{\lunahit}{\textcolor{lunaorange}{\raisebox{0.15ex}{\rule{0.72ex}{0.72ex}}}}
\newcommand{\adaptedhit}{\textcolor{adaptedgreen}{\raisebox{0.15ex}{\rule{0.72ex}{0.72ex}}}}
\newcommand{\monkeymiss}{\textcolor{black!12}{\raisebox{0.15ex}{\rule{0.72ex}{0.72ex}}}}
\newcommand{\faintrowrule}{\arrayrulecolor{black!10}\hline\arrayrulecolor{black}}
\newcommand{\boldtt}[1]{{\fontfamily{pcr}\fontseries{b}\selectfont #1}}

\title{Verifiable pXRD: Hill-climbing pXRD structure determination with LLM agents}

\author{%
  \parbox{\dimexpr\textwidth-2\tabcolsep\relax}{\centering
    Samuel Chen\textsuperscript{1}\quad
    Jiaru Bai\textsuperscript{2,3}\quad
    Austin Henry Cheng\textsuperscript{2,3}\\[0.15em]
    Varinia Bernales\textsuperscript{2}\quad
    Al\'an Aspuru-Guzik\textsuperscript{2,3,4,5,6}\\[0.3em]
    \normalfont\small
    \textsuperscript{1}Stanford University, Stanford, CA, USA\quad
    \textsuperscript{2}University of Toronto, Toronto, ON, Canada\\
    \textsuperscript{3}Vector Institute for Artificial Intelligence, Toronto, ON, Canada\\
    \textsuperscript{4}Acceleration Consortium, University of Toronto, Toronto, ON, Canada\\
    \textsuperscript{5}Canadian Institute for Advanced Research (CIFAR), Toronto, ON, Canada\\
    \textsuperscript{6}NVIDIA, Toronto, ON, Canada\\[0.3em]
    \texttt{sache@stanford.edu}\quad \texttt{jiaru.bai@utoronto.ca}\quad
    \texttt{austin@cs.toronto.edu}\\
    \texttt{bernales@cs.utoronto.ca}\quad \texttt{alan@aspuru.com}%
  }%
}

\hypersetup{
  pdftitle={Verifiable pXRD: Hill-climbing pXRD structure determination with LLM agents},
  pdfauthor={Samuel Chen, Jiaru Bai, Austin H. Cheng, Varinia Bernales, Alán Aspuru-Guzik}
}

\begin{document}

\maketitle
\vspace{-14pt}

% Keep main-text floats close to their discussion without forcing page breaks.
\raggedbottom
\setlength{\textfloatsep}{8pt plus 2pt minus 2pt}
\setlength{\floatsep}{8pt plus 2pt minus 2pt}
\setlength{\intextsep}{8pt plus 2pt minus 2pt}
\captionsetup{font=small,skip=4pt}
\setlength{\parskip}{5pt}

\begin{abstract}
Autonomous crystal-structure determination from powder X-ray diffraction (pXRD) remains a key bottleneck for self-driving materials laboratories. We formulate the task of pXRD structure determination as a verifiable agentic search problem: candidate structures can be assessed during inference through forward-simulated profile agreement and during evaluation through structural matching to a known reference. To benchmark progress, we introduce pXRD-Bench, a suite of 200 simulated single-phase problems spanning in-distribution structure determination, challenging and polymorphic structures, and controlled compositional and topological distribution shifts. Agent adaptation raises selected structural recovery from 27\% to 69\%, with further gains from repeated sampling and verifier-guided inference-time scaling. We extend this approach to experimental multiphase pXRD through a reflexive evolution framework that learns from self-generated synthetic problems. A replaceable laboratory specification defines the sample family and measurement conditions. An evolver agent uses the known generating structures and measurement parameters to diagnose solver failures and develop persistent guidance and executable tools, which are then provided to agents on measured patterns. This supplies verifiable training experience without requiring a collection of solved experimental measurements. On 15 measured patterns from solid-state synthesis, the evolved package improves average profile agreement and transfers an analogue-substitution procedure that recovers a phase whose structure is absent from the supplied reference library. These results demonstrate how learning from verifiable synthetic experience can improve agentic characterization of experimental samples whose structures are not known in advance.

\end{abstract}

\section*{Introduction}
\addcontentsline{toc}{section}{Introduction}
\label{sec:introduction}

Materials discovery and innovation hold promise for advancing many of the world's major challenges, including energy, compute, and sustainability \citep{tabor2018accelerating,depablo2019frontiers}. Certain materials have properties that enable functionalities like energy storage, electrical state control, and carbon sequestration. The distinctive properties of materials are largely determined by their atomic structure. Indeed, structure determination from diffraction and spectroscopy have been central to materials characterization and the broader materials research process, allowing materials researchers to connect synthesis, structure, and function. Insights derived from this connection have driven the improvements in solar cells, batteries, semiconductors, catalysts, and more in recent decades \citep{billingelewin2007,goodwin2019complex}.

Accelerated materials discovery through self-driving labs therefore requires autonomous structural characterization. Powder X-ray diffraction (pXRD) is a primary technique for structure determination that measures how crystalline materials diffract X-rays, producing patterns that encode their phase composition, crystal symmetry, lattice dimensions, and atomic arrangement. Efforts to automate its analysis have included crystallographic software packages \citep{david2008powder,toby2013gsasii}; database-based identification of known phases, including ML-assisted retrieval \citep{maffettone2021xca,fei2026dara}; physics-guided and global structure-search algorithms \citep{meredig2013fpass,anker2025autonomous,su2026abpxrd}; and, of particular recent interest, generative models in both multi-step pipelines and one-shot structure generation \citep{riesel2024crystalyze,choudhary2025diffractgpt,guo2025pxrdnet,li2025pxrdgen,johansen2026decifer,li2026realpxrd,yu2026xrdsol}. Despite this progress, reliable and accurate autonomous powder-diffraction structure determination remains elusive.

The recent development and rapid advancement of LLM agents have opened up a promising approach for autonomous characterization interpretation. Structure determination has historically been an adaptive investigation where scientists generate hypotheses, use established scientific tools, perform forward calculations, verify results, and form conclusions \citep{david2008powder,mccusker1999rietveld}. Rather than relying on algorithmic search techniques or generative models to complete the task end-to-end, AI agents have the potential to perform this adaptive investigation and tackle diverse, realistic problems with greater accuracy. Recent scientific-agent systems illustrate this approach by coupling LLM reasoning with specialized models, simulation codes, and laboratory interfaces \citep{boiko2023coscientist,bran2024chemcrow,zou2025elagente}. Related work has explored modular, reusable skills as a way to provide domain knowledge and computational capabilities without imposing rigid task-specific workflows \citep{perezsanchez2026quntur,deng2026atomisticskills}. Within materials characterization, agentic systems have begun to automate XRD phase identification \citep{maffettone2021xca,dartsi2026chemicalreasoning}, X-ray Rietveld refinement through AgentBuild \citep{shin2026agentbuild}, neutron Rietveld refinement \citep{li2026rongzai}, microscopy-to-theory workflows \citep{yao2025scilink}, and molecular structure elucidation \citep{morales2026nmr}. Here, we formulate pXRD structure determination as a verifiable task and demonstrate its suitability for modern agent adaptation and inference-time scaling techniques, as well as provide comprehensive benchmarking against existing baselines and strong general-purpose coding agents.

Following our initial submission to NeurIPS AI4Mat 2026, parallel works AutoXRD \citep{wu2026autoxrd} and Gan Jiang \citep{cao2026ganjiang} also demonstrated rigorous benchmarking and designed/learned skills for agents on pXRD, and Periodic Labs reported improvements in experimental multiphase interpretation through model post-training and increased inference-time computation \citep{periodiclabs2026nature}. \citet{woo2026pxrdbench} also introduced PXRD-Bench for evaluating generative structure models.

We identify two ways in which pXRD structure determination is verifiable—at inference-time and at evaluation-time—and apply multiple techniques to leverage this verifiability to improve performance.
While verifiability makes this problem suitable for reinforcement learning \citep{lambert2025tulu,guo2025deepseekr1,yu2025dapo}, we demonstrate meaningful performance improvements through agent adaptation \citep{jiang2025adaptation} and inference-time scaling techniques that can be applied easily to frontier closed-weight models. We benchmark our methods on a primary 200-problem benchmark of single-phase patterns and a 20-problem representative subset for the inference-time scaling techniques. Our adapted agent achieves the highest structure-recovery accuracy among the methods evaluated on pXRD-Bench, and inference-time compute scaling yields further gains on the 20-problem panel. We extend the applicability of our findings by introducing a reflexive evolution framework that learns from self-generated synthetic training
problems that achieves superior refinement and phase recovery capabilities on a 15-problem multiphase panel from the Precursor Genome \citep{walters2026precursorgenome}, as well as providing ablations that give useful insights about the value of evaluation-time verifiability, the limits of learned skills/guidance, and best practices for using agents to evaluate pXRD patterns. Code, benchmark data, and per-experiment result artifacts are available at: \url{https://github.com/sache07/verifiable-pxrd}.

\section*{pXRD structure determination as a verifiable task}
\addcontentsline{toc}{section}{pXRD structure determination as a verifiable task}
\label{sec:problem}

Structural characterization problems can be posed as inverse problems. Given characterization measurements and potentially some experimental parameters, the goal is to determine the underlying atomic structure. For pXRD structure determination, we consider the following inverse problem. Given a chemical composition $c$, an X-ray wavelength $\lambda$, and a measured intensity profile $y(2\theta)$, infer a periodic crystal structure $s$, described using a CIF. 

\begin{figure}[!tbp]
    \centering
    \setlength{\parskip}{0pt}
    \includegraphics[width=0.90\linewidth]{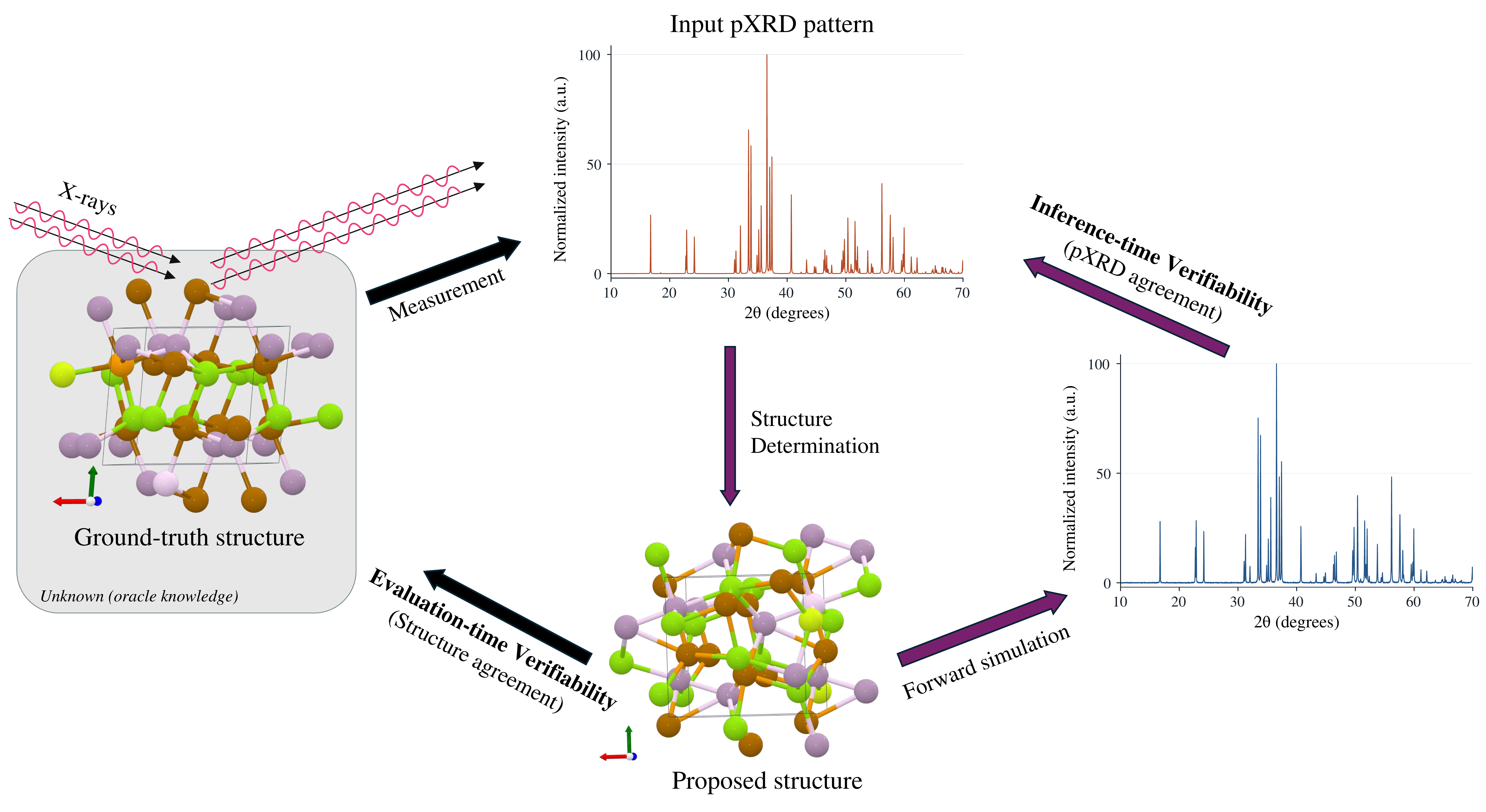}
    \caption{pXRD structure determination as a verifiable inverse problem. Ground truth: pXRD-Bench case \texttt{pxrd100-rep-010}. Proposed: \texttt{pxrd-codex} \texttt{selected.cif}. Structures rendered with Crystal Toolkit \citep{horton2023crystal}.}
    \label{fig:pxrd-verifiability}
\end{figure}

This problem is a verifiable task, making it particularly amenable for performance hill-climbing through both agent adaptation and inference-time scaling. A reliable, inexpensive verifier provides a stable objective for hill-climbing through updates to agent model policy, tools, or scaffolding \citep{shinn2023reflexion,yang2024sweagent}. This generate–verify–update loop has been central to the rapid advancement in LLM agent capabilities in coding and mathematics, where checking a candidate solution is often substantially easier than producing one \citep{li2022alphacode,hubert2026alphaproof}. During inference, it can rank parallel rollouts, guide targeted revisions, and allocate additional compute to insufficient solutions\citep{brown2024large,snell2025scaling,setlur2025scaling}. Figure~\ref{fig:pxrd-verifiability} summarizes the inverse problem and its inference- and evaluation-time verification pathways.

In particular, pXRD structure determination is verifiable in two distinct senses:
\begingroup
\setlength{\leftmargini}{1.5em}
\begin{enumerate}
\item \textbf{Inference-time verifiability:} When a diffraction measurement is taken of a sample, the sample's true structure diffracts X-rays in such a way as to produce the measured profile. This ground truth structure is unknown, so success can be verified by calculating the forward model $y_i=F(s^\star,\eta^\star;\lambda)(2\theta_i)+\epsilon_i$, where $y_i$ is the intensity at $2\theta_i$, \(\eta^\star\) denotes experimental and profile nuisance parameters, and $\epsilon_i$ captures measurement noise and physical effects not represented by the forward model. This forward calculation is compared with the measured profile, often by computing the weighted profile residual \(R_{\mathrm{wp}}\) \citep{rietveld1969}. A low residual establishes that a candidate is consistent with the measurement, a strong verifier of solution quality. It cannot, however, establish the absence of equally or more diffraction-consistent structures.

\item \textbf{Evaluation-time verifiability:} When evaluating the success of a pXRD structure solver, there is often a ground truth structure solution. Generative models are trained on pairs of diffraction profiles and the true structures that generated them, and benchmarking often evaluates the ability of a solver/solution-generator to generate a known reference structure given its diffraction profile. Success is verified by the similarity of the proposed structure and the ground truth structure. pymatgen's \texttt{StructureMatcher} is commonly used for this comparison, accepting structures as equivalent if, after applying a coordinate transformation that accounts for equivalent cell choices, basis transformations, and origin shifts, lattice-length mismatch is below \(l_{\mathrm{tol}}\), the lattice-angle mismatch is below \(\theta_{\mathrm{tol}}\), and the maximum normalized corresponding atomic site displacement is below \(s_{\mathrm{tol}}\).
\end{enumerate}
\endgroup

The agent-adaptation and inference-time scaling techniques we utilize and benchmark in our results take advantage of both types of verifiability in various ways to improve performance. We utilize the evaluation-time structure-matcher verification as our primary metric of success in our benchmarking, following precedent in the literature and to compare against baseline methods.

\section*{Results and Discussion}
\addcontentsline{toc}{section}{Results and Discussion}
\label{sec:results-discussion}

\section{Agents for single-phase pXRD: pXRD-Bench}
\label{sec:single-phase-results}

\subsection{pXRD-Bench}
\label{sec:pxrd-bench}

To establish a uniform measure of solver performance and enable comparison between existing methods, our presented agent frameworks, and future approaches, we present \textbf{pXRD-Bench}. pXRD-Bench comprises 200 single-phase structure-determination problems divided equally between an in-distribution corpus, pXRD-100, and an out-of-distribution corpus, pXRD-ood. This suite is distinct from the independently developed PXRD-Bench of \citet{woo2026pxrdbench}, which evaluates generative models under controlled training splits and scores candidates before downstream refinement. For each problem, the solver receives a reduced chemical composition, the X-ray wavelength, and a simulated intensity profile \(y(2\theta)\), and must return a complete periodic structure as a CIF. The reference structure, source identifier, lattice parameters, space group, formula-unit count, and benchmark category remain hidden during inference. All case-selection criteria, summarized below and detailed in Appendix~\ref{app:pxrd-bench-details}, were fixed before running any evaluated solver.

\begin{table*}[!tbp]
    \centering
    \setlength{\parskip}{0pt}
    \small
    \setlength{\tabcolsep}{2pt}
    \renewcommand{\arraystretch}{1.08}
    \caption{Composition of pXRD-Bench. pXRD-100 measures recovery of structures represented in the training distributions of existing generative methods, whereas pXRD-ood separates compositional extrapolation from structural-topology extrapolation.}
    \label{tab:pxrd-bench-composition}
    \begin{tabular}{llrp{0.61\textwidth}}
        \toprule
        Corpus & Category & \(n\) & Construction and intended test \\
        \midrule
        pXRD-100
            & Representative
            & 30
            & Distribution-matched sample of eligible MP-20 structures, balancing crystal system, primitive-cell size, chemical complexity, stability, density, and elemental prevalence. \\
        pXRD-100
            & Challenge
            & 30
            & Six structures selected along each of five predefined difficulty axes: rare symmetry, site and chemical complexity, peak congestion, scattering ambiguity, and large primitive cells. \\
        pXRD-100
            & Polymorph
            & 40
            & Twenty structurally distinct pairs with identical reduced compositions, stratified by diffraction profile similarity \\
        \midrule
        pXRD-ood
            & Substitution
            & 25
            & Composition-novel structures obtained by full-orbit isovalent substitution of an exposed structural parent. \\
        pXRD-ood
            & Ordered decoration
            & 25
            & Composition-novel structures obtained by ordered decoration of symmetry orbits in an exposed structural parent. \\
        pXRD-ood
            & Wyckoff enumeration
            & 25
            & Topology-novel structures with exposed compositions generated by symmetry-constrained Wyckoff enumeration. \\
        pXRD-ood
            & Crystal search
            & 25
            & Topology-novel structures with exposed compositions obtained from MACE-MH-1 crystal-search \\
        \bottomrule
    \end{tabular}
\end{table*}

The pXRD-100 corpus contains 100 ordered Materials Project structures drawn from the frozen MP-20 training split \citep{jain2013materialsproject,xie2022cdvae}. Every structure contains at most 20 sites in its primitive cell and is also an audited strict training-set match for Crystalyze, PXRDGen, deCIFer, and DiffractGPT, and plausibly exposed to the other frameworks. \citep{riesel2024crystalyze,li2025pxrdgen,johansen2026decifer,choudhary2025diffractgpt}. pXRD-100 is therefore intentionally an in-distribution benchmark rather than a held-out test of novelty. However, the challenge and polymorph panels were designed to stress-test the solvers, as recent work has shown that peak overlap, weak elemental contrast, and light-atom localization remain difficult for learned pXRD methods \citep{li2025pxrdgen}, while evaluation on same-composition polymorphs helps establish that predictions are conditioned on the diffraction signal rather than primarily on a composition--structure prior \citep{segal2026xrdiff}.

The pXRD-ood corpus instead tests two operationally distinct forms of extrapolation. Novelty is evaluated relative to a frozen exposure boundary containing the Materials Project, the Crystallography Open Database (COD), JARVIS, and the released training corpora or retrieval indices associated with the evaluated structure-determination systems \citep{jain2013materialsproject,grazulis2012cod,choudhary2020jarvis,riesel2024crystalyze,li2025pxrdgen,johansen2026decifer,choudhary2025diffractgpt,fei2026dara}. Table~\ref{tab:pxrd-bench-composition} summarizes the benchmark composition by category.

\begin{table*}[!tbp]
    \centering
    \setlength{\parskip}{0pt}
    \scriptsize
    \setlength{\tabcolsep}{2.5pt}
    \renewcommand{\arraystretch}{1.08}
    \caption{Exact structural recovery of existing pXRD structure-determination methods on pXRD-Bench. Category and corpus entries report \textbf{pass@1} \(\mid\) \textbf{pass@8} as the number of recovered structures.}
    \label{tab:pxrd-bench-baselines}
    \resizebox{\textwidth}{!}{%
    \begin{tabular}{l*{10}{r@{\(\,\mid\,\)}l}}
        \toprule
        & \multicolumn{6}{c}{pXRD-100 categories}
        & \multicolumn{8}{c}{pXRD-ood categories}
        & \multicolumn{6}{c}{Aggregate} \\
        \cmidrule(lr){2-7}\cmidrule(lr){8-15}\cmidrule(lr){16-21}
        Method
            & \multicolumn{2}{c}{Rep.}
            & \multicolumn{2}{c}{Chall.}
            & \multicolumn{2}{c}{Poly.}
            & \multicolumn{2}{c}{Sub.}
            & \multicolumn{2}{c}{Ord.}
            & \multicolumn{2}{c}{Wyck.}
            & \multicolumn{2}{c}{CSS}
            & \multicolumn{2}{c}{pXRD-100}
            & \multicolumn{2}{c}{pXRD-ood}
            & \multicolumn{2}{c}{Overall} \\
        \midrule
        Crystalyze
            & 5 & 9 & 2 & 4 & 4 & 6
            & 2 & 4 & 0 & 1 & 0 & 0 & 0 & 1
            & 11 & 19 & 2 & 6 & 6.5\% & 12.5\% \\
        DiffractGPT
            & 5 & -- & 0 & -- & 4 & --
            & 5 & -- & 1 & -- & 0 & -- & 0 & --
            & 9 & -- & 6 & -- & 7.5\% & -- \\
        Ab-PXRD-Solver
            & 21 & -- & 5 & -- & 16 & --
            & 9 & -- & 9 & -- & 1 & -- & 1 & --
            & 42 & -- & 20 & -- & 31.0\% & -- \\
        deCIFer\(^{\dagger}\)
            & 9 & 12 & 3 & 9 & 6 & 11
            & 4 & 5 & 3 & 5 & 0 & 0 & 2 & 2
            & 18 & 32 & 9 & 12 & 13.5\% & 22.0\% \\
        PXRDGen\(^{\ddagger}\)
            & 21 & 30 & 11 & 22 & 15 & 32
            & 10 & 12 & 7 & 13 & 0 & 0 & 0 & 1
            & 47 & 84 & 17 & 26 & 32.0\% & 55.0\% \\
        RealPXRD-Solver\(^{\ddagger}\)
            & 20 & 26 & 8 & 15 & 13 & 21
            & 12 & 16 & 7 & 10 & 0 & 0 & 1 & 2
            & 41 & 62 & 20 & 28 & 30.5\% & 45.0\% \\
        GPT-5.6 Sol (high)
            & 15 & -- & 6 & -- & 9 & --
            & 11 & -- & 6 & -- & 2 & -- & 5 & --
            & 30 & -- & 24 & -- & 27.0\% & -- \\
        \bottomrule
    \end{tabular}%
    }

    \vspace{2pt}
    \parbox{\textwidth}{\footnotesize
    \(^{\dagger}\) deCIFer receives the known conventional-cell formula-unit count \(Z\), matching its native conditioning.
    \(^{\ddagger}\) PXRDGen and RealPXRD-Solver receive the known primitive-cell \(Z\); these are oracle-input comparisons.}
\end{table*}

When benchmarking other solvers for this comparison, the aim was to allow them to perform to the best of their ability. The diffraction inputs given to each system were calculated from the same reference CIFs in their respective native representations using inputs that reproduce each method's documented training and/or evaluation preprocessing (see Appendix~\ref{app:baseline-protocols}). This includes wavelength convention, angular or reciprocal-space grid, profile model, broadening, noise model, and discretization pulled from the papers or source code. When benchmarking our agent frameworks, we forward-simulated the diffraction inputs with GSAS-II using a fixed, benchmark-specific profile intended to represent an idealized, background-subtracted laboratory measurement\citep{toby2013gsasii}. The simulation uses monochromatic Cu K\(\alpha_1\) radiation with \(\lambda=1.5406~\text{\AA}\) and span \(5^\circ\) to \(90^\circ\) in \(2\theta\) at \(0.01^\circ\) intervals, yielding 8,501 intensity values. The constant-wavelength pseudo-Voigt profile uses \(U=2\), \(V=-2\), \(W=5\), \(X=Y=0\), polarization \(0.7\), and \(SH/L=0.002\), together with isotropic broadening from a \(1~\mu\mathrm{m}\) crystallite size and 1000 ppm microstrain (parameter conventions are defined in Appendix~\ref{app:profile-parameters}). Preferred orientation, sample displacement, and background are disabled. Each clean profile is scaled to a maximum of 5,000 counts, sampled with Poisson noise, augmented with Gaussian noise of standard deviation five counts, and finally normalized to a maximum intensity of 100.

We evaluate submitted structures using pymatgen's \texttt{StructureMatcher} \citep{ong2013pymatgen}. We use pymatgen’s default geometric tolerances, \(l_{\mathrm{tol}}=0.2\), \(\theta_{\mathrm{tol}}=5^\circ\), and \(s_{\mathrm{tol}}=0.3\), while disabling lattice rescaling to require recovery of the unit-cell dimensions.

As expected, exact recovery was consistently higher on pXRD-100 than on pXRD-ood. On pass@1, PXRDGen, Ab-PXRD-Solver, and RealPXRD-Solver form a narrow leading cluster at 30.5--32.0\% overall, although PXRDGen and RealPXRD-Solver receive the oracle primitive-cell \(Z\). Ab-PXRD-Solver and DiffractGPT do not have pass@8 results because Ab-PXRD-Solver carries out substantial search and comparison producing one final candidate, while DiffractGPT uses a greedy decoding algorithm that results in identical samples. However, the other methods show significant gains from best-of-8 sampling. Notably, GPT-5.6 Sol remains competitive at 27.0\% overall and 24\% pXRD-ood on pass@1 with a simple prompt, no access to code/tools, and no oracle \(Z\) knowledge. This suggests that frontier LLMs possess strong crystallographic knowledge and reasoning, and in the rest of the paper we explore techniques utilizing them to reach even higher performance.

\subsection{Agent adaptation}
\label{sec:agent-adaptation}

Agent adaptation improves a fixed language model by changing the harness around it—its action interface, accessible tools and data, and reusable task knowledge---rather than its parameters. Prior work shows that interleaving reasoning with environment actions and designing agent-facing computer interfaces can materially change what an LLM can accomplish \citep{yao2023react,yang2024sweagent}.
Complementary approaches use task feedback to distill experience into persistent linguistic memory without weight updates \citep{shinn2023reflexion,zhao2024expel}. pXRD structure determination is especially amenable to this form of adaptation: established crystallographic programs provide precise operations for indexing, simulation, and refinement, while exact structure matching turns completed attempts into unambiguous supervision. We therefore construct a controlled adaptation ladder around the same frontier model. Starting from tool-free prediction relying solely on LLM reasoning, we add a coding-agent harness with a general materials-science Python environment and crystallographic database access (\texttt{matsci-codex}), then access to GSAS-II and DICVOL14 (\texttt{pxrd-codex}), and finally a compact frozen capability synthesized from externally verified attempts on a disjoint 50-problem curriculum (\texttt{adapted-codex}). This design separates improvements due to agent scaffolding, domain tools, and learned nonparametric capability from changes to the underlying model.

All conditions received the same simple prompt, with the exception of GPT-5.6 Sol (high) receiving ``from the below pattern'' instead of ``from \texttt{pattern.csv}'' and the complete diffraction profile as prompt text because the no-tools Responses API condition truncates .csv files, and \texttt{adapted-codex} receiving a prefix saying to use its capability package. 

\begingroup
\centering
\begin{minipage}{0.92\linewidth}
    \small
    \setlength{\parskip}{3pt}
    \hrule
    \vspace{0.4em}
    \textbf{Benchmark prompt}
    \vspace{0.2em}

    Solve the crystal structure from \texttt{pattern.csv}. The composition is \(\langle\textit{reduced composition}\rangle\).

    The X-ray wavelength is \(\langle\textit{wavelength}\rangle\) angstrom (\(\langle\textit{radiation}\rangle\)).

    Save your primary structure as \texttt{selected.cif} and, if necessary, up to a few surviving alternatives as \texttt{alternative-<n>.cif}.
    \vspace{0.4em}
    \hrule
\end{minipage}
\par
\endgroup

Moving from tool-free GPT-5.6 Sol (high) to \texttt{matsci-codex} increases selected-structure recovery from 27.0\% to 52.0\%, a gain of 25 percentage points. Because the model weights are fixed, this comparison measures the joint value of a stateful coding-agent substrate, a scientific Python environment, and crystallographic database access. In place of a single model response, the agent can maintain workspace state, inspect intermediate results, and synthesize case-specific programs that compose established materials-science libraries. This improvement requires substantially more test-time computation: the mean reported input-plus-output usage was \(6.09\) million tokens over a mean wall-clock duration of 25.3 minutes among completed \texttt{matsci-codex} rollouts that didn't hit the 1-hour limit (\(n=187\)), compared with 161,293 tokens and 6.94 minutes across the 200 single-call, tool-free GPT-5.6 Sol (high) condition solves. Adding crystallography-specific software in \texttt{pxrd-codex} yields a further five-point increase in overall recovery, from 52.0\% to 57.0\%. This gain is concentrated in pXRD-ood, where recovery rises from 44\% to 56\%, while pXRD-100 changes from 60\% to 58\%, within the range of stochastic variance. This moderate gain demonstrates how access pXRD structure determination software is beneficial for agent performance, but also how coding agents are already able to consistently carry out complex crystallographic tasks like cell-indexing, charge flipping, and Rietveld refinement using custom scripts.

\begin{table}[!tbp]
    \centering
    \setlength{\parskip}{0pt}
    \scriptsize
    \setlength{\tabcolsep}{2.5pt}
    \renewcommand{\arraystretch}{1.08}
    \caption{Exact structural recovery across the agent-adaptation ladder on pXRD-Bench. Entries report \textbf{selected exact} \(\mid\) \textbf{any exact}, where any exact permits a match from either \texttt{selected.cif} or \texttt{alternative-<n>.cif}.}
    \label{tab:agent-adaptation}
    \resizebox{\textwidth}{!}{%
    \begin{tabular}{l*{10}{r@{\(\,\mid\,\)}l}}
        \toprule
        & \multicolumn{6}{c}{pXRD-100 categories}
        & \multicolumn{8}{c}{pXRD-ood categories}
        & \multicolumn{6}{c}{Aggregate} \\
        \cmidrule(lr){2-7}\cmidrule(lr){8-15}\cmidrule(lr){16-21}
        Condition
            & \multicolumn{2}{c}{Rep.}
            & \multicolumn{2}{c}{Chall.}
            & \multicolumn{2}{c}{Poly.}
            & \multicolumn{2}{c}{Sub.}
            & \multicolumn{2}{c}{Ord.}
            & \multicolumn{2}{c}{Wyck.}
            & \multicolumn{2}{c}{CSS}
            & \multicolumn{2}{c}{pXRD-100}
            & \multicolumn{2}{c}{pXRD-ood}
            & \multicolumn{2}{c}{Overall} \\
        \midrule
        GPT-5.6 Sol (high)
            & 15 & 16 & 6 & 6 & 9 & 11
            & 11 & 12 & 6 & 6 & 2 & 2 & 5 & 5
            & 30 & 33 & 24 & 25 & 27.0\% & 29.0\% \\
        \texttt{matsci-codex}
            & 25 & 26 & 14 & 14 & 21 & 21
            & 15 & 15 & 12 & 12 & 7 & 7 & 10 & 10
            & 60 & 61 & 44 & 44 & 52.0\% & 52.5\% \\
        \texttt{pxrd-codex}
            & 22 & 23 & 12 & 12 & 24 & 24
            & 20 & 20 & 16 & 18 & 8 & 8 & 12 & 12
            & 58 & 59 & 56 & 58 & 57.0\% & 58.5\% \\
        \texttt{adapted-codex}
            & 27 & 27 & 22 & 22 & 28 & 28
            & 19 & 19 & 21 & 21 & 10 & 12 & 11 & 12
            & 77 & 77 & 61 & 64 & 69.0\% & 70.5\% \\
        \bottomrule
    \end{tabular}%
    }
\end{table}

A modern paradigm in agent adaptation is experience-driven self-evolution: an agent executes a curriculum of tasks, receives external feedback on its rollouts, and distills recurring successes and failures into persistent prompts, memories, skills, or scaffold code that changes its future behavior without updating the base model. Reflexion and ExpeL \citep{shinn2023reflexion,zhao2024expel} established this pattern for linguistic memory, Voyager \citep{wang2023voyager} accumulated reusable executable skills through environment interaction, and more recent systems such as GEPA, Agentic Context Engineering (ACE), and the Self-Improving Coding Agent (SICA) \citep{agrawal2026gepa,zhang2025ace,robeyns2025sica} optimize increasingly general forms of agent state---from prompts and evolving playbooks to the agent implementation itself---using evaluated trajectories. Verifiability allows for easier learning during these processes by allowing richer and more direct feedback from trajectories.

We instantiate this experience-driven loop to create \texttt{adapted-codex} using a trainer--trainee framework and a disjoint 50-problem curriculum sampled from the same distribution as pXRD-Bench. The trainer launches a blind \texttt{pxrd-codex} trainee on each curriculum problem, analyzes their trajectories and outcomes, designs a capability package comprising natural-language skills and optional executable tool adapters, and iteratively reruns problems and revises the capability pack to optimize performance. The trainer had access to the ground-truth structures as well as the structure-matcher verifier, meaning \textit{evaluation-time verifiability} was utilized as a learning signal. The resulting \texttt{adapted-codex} capability pack contained just (1) a 139-word skill and (2) a DICVOL adapter that improves the agent's understanding of raw DICVOL input—a minimal surface as intended by our trainer prompt. More details on the training process and resulting package are provided in Appendix~\ref{app:adapted-codex-self-evolution}. 

\texttt{adapted-codex} achieved 69.0\% exact matches from \texttt{selected.cif} on pXRD-Bench, significantly higher than \texttt{pxrd-codex} and the other conditions. Our framework is a relatively simple and efficient implementation of self-evolution, in which we find limiting the package surface achieved better performance than allowing the package to sprawl and attempt to correct for each observed mistake. In practice, a significant portion performance gain can be attributed to trainer-written guidance in the skills about the nature of the task and verifier; this guidance leads behavior to be better aligned with verifier scoring,  analogous to verifier-guided prompt optimization and reinforcement learning with verifiable rewards \citep{agrawal2026gepa,lambert2025tulu,guo2025deepseekr1}. These results with our proof-of-concept approach establish pXRD structure determination as a suitable domain for experience-driven agent adaptation and motivate more involved reflective optimizers and post-training techniques.

\subsection{Inference-time scaling}
\label{sec:inference-time-scaling}

Inference-time scaling improves a fixed agent by allocating additional computation to each problem rather than modifying its parameters or persistent capabilities. The simplest approach samples independent solution trajectories and aggregates or ranks them after generation \citep{cobbe2021training,wang2023selfconsistency,brown2024large}. More structured methods make this computation adaptive: they extend or revise active trajectories, branch and backtrack over partial solutions, and direct search using learned value functions, executable feedback, or formal verifiers \citep{muennighoff2025s1,yao2023tree,hao2023rap,zhou2024lats,snell2025scaling,setlur2025scaling}. We demonstrate the potential of inference-time scaling on agentic pXRD structure determination through repeated sampling on a 20 problem representative subset of pXRD-Bench, which we denote \textbf{pXRD-Bench-20} (stored as \texttt{inference-scaling-20}), as well as results demonstrating the value of LLMs as verifiers/judges for adjudication and coverage-directed continuation. Full experimental details including panel construction, rollout, verifier and token-accounting protocols are given in Appendix~\ref{app:inference-scaling-details}.

\begin{figure}[!b]
    \centering
    \setlength{\parskip}{0pt}
    \scriptsize
    \setlength{\tabcolsep}{0.55pt}
    \renewcommand{\arraystretch}{0.54}
    \begin{minipage}[t]{0.495\linewidth}
        \vspace{0pt}
        \centering
        \textbf{\strut (a) \boldtt{pxrd-codex}: GPT-5.6 Sol (high)}

        \vspace{2pt}
        \resizebox{\linewidth}{!}{%
        \begin{tabular}{c*{20}{c}c}
            \toprule
            & \multicolumn{20}{c}{pXRD-Bench-20 problem number} & \\
            \cmidrule(lr){2-21}
            Pass & 1 & 2 & 3 & 4 & 5 & 6 & 7 & 8 & 9 & 10 & 11 & 12 & 13 & 14 & 15 & 16 & 17 & 18 & 19 & 20 & Exact \\
            \midrule
            1 & \monkeymiss & \solhit & \solhit & \solhit & \monkeymiss & \monkeymiss & \solhit & \solhit & \solhit & \monkeymiss & \monkeymiss & \monkeymiss & \monkeymiss & \monkeymiss & \monkeymiss & \solhit & \solhit & \solhit & \monkeymiss & \solhit & 10 \\
            \faintrowrule
            2 & \monkeymiss & \monkeymiss & \monkeymiss & \solhit & \monkeymiss & \monkeymiss & \solhit & \solhit & \solhit & \monkeymiss & \solhit & \solhit & \solhit & \monkeymiss & \monkeymiss & \solhit & \solhit & \solhit & \monkeymiss & \solhit & 11 \\
            \faintrowrule
            3 & \monkeymiss & \monkeymiss & \solhit & \solhit & \monkeymiss & \monkeymiss & \monkeymiss & \monkeymiss & \solhit & \monkeymiss & \solhit & \monkeymiss & \solhit & \monkeymiss & \monkeymiss & \solhit & \solhit & \monkeymiss & \solhit & \solhit & 9 \\
            \faintrowrule
            4 & \monkeymiss & \solhit & \solhit & \solhit & \monkeymiss & \monkeymiss & \solhit & \solhit & \solhit & \monkeymiss & \monkeymiss & \monkeymiss & \monkeymiss & \monkeymiss & \monkeymiss & \solhit & \solhit & \monkeymiss & \monkeymiss & \solhit & 9 \\
            \faintrowrule
            5 & \monkeymiss & \solhit & \monkeymiss & \solhit & \monkeymiss & \monkeymiss & \monkeymiss & \solhit & \solhit & \monkeymiss & \solhit & \solhit & \solhit & \monkeymiss & \monkeymiss & \solhit & \solhit & \monkeymiss & \monkeymiss & \solhit & 10 \\
            \faintrowrule
            6 & \monkeymiss & \solhit & \solhit & \solhit & \monkeymiss & \monkeymiss & \solhit & \solhit & \solhit & \monkeymiss & \monkeymiss & \solhit & \monkeymiss & \monkeymiss & \monkeymiss & \solhit & \solhit & \solhit & \solhit & \monkeymiss & 11 \\
            \faintrowrule
            7 & \monkeymiss & \solhit & \monkeymiss & \solhit & \solhit & \monkeymiss & \solhit & \solhit & \solhit & \monkeymiss & \solhit & \solhit & \monkeymiss & \monkeymiss & \monkeymiss & \solhit & \solhit & \solhit & \solhit & \monkeymiss & 12 \\
            \faintrowrule
            8 & \monkeymiss & \monkeymiss & \solhit & \solhit & \monkeymiss & \monkeymiss & \solhit & \solhit & \solhit & \monkeymiss & \solhit & \solhit & \solhit & \monkeymiss & \monkeymiss & \solhit & \solhit & \monkeymiss & \solhit & \solhit & 12 \\
            \bottomrule
        \end{tabular}%
        }
    \end{minipage}%
    \hfill
    \begin{minipage}[t]{0.495\linewidth}
        \vspace{0pt}
        \centering
        \textbf{\strut (b) \boldtt{pxrd-codex}: GPT-5.6 Luna (max)}

        \vspace{2pt}
        \resizebox{\linewidth}{!}{%
        \begin{tabular}{c*{20}{c}c}
            \toprule
            & \multicolumn{20}{c}{pXRD-Bench-20 problem number} & \\
            \cmidrule(lr){2-21}
            Pass & 1 & 2 & 3 & 4 & 5 & 6 & 7 & 8 & 9 & 10 & 11 & 12 & 13 & 14 & 15 & 16 & 17 & 18 & 19 & 20 & Exact \\
            \midrule
            1 & \monkeymiss & \monkeymiss & \lunahit & \lunahit & \monkeymiss & \monkeymiss & \monkeymiss & \monkeymiss & \lunahit & \monkeymiss & \monkeymiss & \monkeymiss & \lunahit & \monkeymiss & \monkeymiss & \lunahit & \lunahit & \lunahit & \lunahit & \lunahit & 9 \\
            \faintrowrule
            2 & \monkeymiss & \monkeymiss & \lunahit & \lunahit & \monkeymiss & \monkeymiss & \monkeymiss & \lunahit & \lunahit & \monkeymiss & \monkeymiss & \monkeymiss & \monkeymiss & \monkeymiss & \monkeymiss & \lunahit & \lunahit & \monkeymiss & \monkeymiss & \lunahit & 7 \\
            \faintrowrule
            3 & \monkeymiss & \monkeymiss & \monkeymiss & \lunahit & \monkeymiss & \monkeymiss & \monkeymiss & \monkeymiss & \lunahit & \monkeymiss & \monkeymiss & \lunahit & \monkeymiss & \monkeymiss & \monkeymiss & \lunahit & \monkeymiss & \monkeymiss & \monkeymiss & \lunahit & 5 \\
            \faintrowrule
            4 & \monkeymiss & \monkeymiss & \monkeymiss & \lunahit & \monkeymiss & \monkeymiss & \monkeymiss & \lunahit & \lunahit & \monkeymiss & \monkeymiss & \monkeymiss & \lunahit & \monkeymiss & \monkeymiss & \lunahit & \monkeymiss & \monkeymiss & \monkeymiss & \lunahit & 6 \\
            \faintrowrule
            5 & \monkeymiss & \monkeymiss & \monkeymiss & \lunahit & \monkeymiss & \monkeymiss & \monkeymiss & \lunahit & \lunahit & \monkeymiss & \monkeymiss & \monkeymiss & \monkeymiss & \monkeymiss & \monkeymiss & \lunahit & \monkeymiss & \monkeymiss & \monkeymiss & \lunahit & 5 \\
            \faintrowrule
            6 & \monkeymiss & \monkeymiss & \lunahit & \lunahit & \monkeymiss & \monkeymiss & \lunahit & \monkeymiss & \lunahit & \monkeymiss & \monkeymiss & \monkeymiss & \lunahit & \monkeymiss & \monkeymiss & \lunahit & \monkeymiss & \monkeymiss & \monkeymiss & \monkeymiss & 6 \\
            \faintrowrule
            7 & \monkeymiss & \monkeymiss & \monkeymiss & \lunahit & \monkeymiss & \monkeymiss & \lunahit & \lunahit & \lunahit & \monkeymiss & \monkeymiss & \monkeymiss & \monkeymiss & \monkeymiss & \monkeymiss & \lunahit & \monkeymiss & \monkeymiss & \monkeymiss & \lunahit & 6 \\
            \faintrowrule
            8 & \monkeymiss & \monkeymiss & \lunahit & \lunahit & \monkeymiss & \monkeymiss & \monkeymiss & \lunahit & \lunahit & \monkeymiss & \monkeymiss & \monkeymiss & \monkeymiss & \monkeymiss & \monkeymiss & \lunahit & \monkeymiss & \lunahit & \monkeymiss & \lunahit & 7 \\
            \bottomrule
        \end{tabular}%
        }
    \end{minipage}%

    \vspace{4pt}

    \begin{minipage}[t]{0.495\linewidth}
        \vspace{0pt}
        \centering
        \textbf{(c) \boldtt{adapted-codex}: GPT-5.6 Sol (high)}

        \vspace{2pt}
        \resizebox{\linewidth}{!}{%
        \begin{tabular}{c*{20}{c}c}
            \toprule
            & \multicolumn{20}{c}{pXRD-Bench-20 problem number} & \\
            \cmidrule(lr){2-21}
            Pass & 1 & 2 & 3 & 4 & 5 & 6 & 7 & 8 & 9 & 10 & 11 & 12 & 13 & 14 & 15 & 16 & 17 & 18 & 19 & 20 & Exact \\
            \midrule
            1 & \adaptedhit & \adaptedhit & \monkeymiss & \adaptedhit & \monkeymiss & \monkeymiss & \adaptedhit & \adaptedhit & \adaptedhit & \adaptedhit & \monkeymiss & \adaptedhit & \monkeymiss & \monkeymiss & \adaptedhit & \adaptedhit & \adaptedhit & \adaptedhit & \monkeymiss & \adaptedhit & 13 \\
            \faintrowrule
            2 & \monkeymiss & \adaptedhit & \adaptedhit & \adaptedhit & \monkeymiss & \monkeymiss & \adaptedhit & \adaptedhit & \adaptedhit & \adaptedhit & \monkeymiss & \adaptedhit & \adaptedhit & \monkeymiss & \monkeymiss & \adaptedhit & \adaptedhit & \adaptedhit & \monkeymiss & \adaptedhit & 13 \\
            \faintrowrule
            3 & \adaptedhit & \adaptedhit & \monkeymiss & \adaptedhit & \monkeymiss & \monkeymiss & \adaptedhit & \adaptedhit & \adaptedhit & \adaptedhit & \adaptedhit & \adaptedhit & \adaptedhit & \monkeymiss & \adaptedhit & \adaptedhit & \monkeymiss & \adaptedhit & \monkeymiss & \adaptedhit & 14 \\
            \faintrowrule
            4 & \adaptedhit & \adaptedhit & \monkeymiss & \adaptedhit & \monkeymiss & \monkeymiss & \adaptedhit & \adaptedhit & \adaptedhit & \adaptedhit & \monkeymiss & \adaptedhit & \adaptedhit & \monkeymiss & \adaptedhit & \adaptedhit & \adaptedhit & \adaptedhit & \monkeymiss & \adaptedhit & 14 \\
            \faintrowrule
            5 & \adaptedhit & \adaptedhit & \monkeymiss & \adaptedhit & \adaptedhit & \monkeymiss & \adaptedhit & \adaptedhit & \adaptedhit & \adaptedhit & \adaptedhit & \adaptedhit & \adaptedhit & \monkeymiss & \adaptedhit & \adaptedhit & \adaptedhit & \adaptedhit & \monkeymiss & \adaptedhit & 16 \\
            \faintrowrule
            6 & \monkeymiss & \monkeymiss & \adaptedhit & \adaptedhit & \adaptedhit & \monkeymiss & \adaptedhit & \adaptedhit & \adaptedhit & \adaptedhit & \monkeymiss & \adaptedhit & \adaptedhit & \monkeymiss & \adaptedhit & \adaptedhit & \adaptedhit & \monkeymiss & \adaptedhit & \monkeymiss & 13 \\
            \faintrowrule
            7 & \monkeymiss & \adaptedhit & \adaptedhit & \adaptedhit & \adaptedhit & \monkeymiss & \adaptedhit & \adaptedhit & \adaptedhit & \adaptedhit & \monkeymiss & \adaptedhit & \adaptedhit & \monkeymiss & \adaptedhit & \adaptedhit & \adaptedhit & \monkeymiss & \monkeymiss & \adaptedhit & 14 \\
            \faintrowrule
            8 & \adaptedhit & \adaptedhit & \monkeymiss & \adaptedhit & \adaptedhit & \monkeymiss & \adaptedhit & \adaptedhit & \adaptedhit & \adaptedhit & \monkeymiss & \adaptedhit & \monkeymiss & \monkeymiss & \monkeymiss & \adaptedhit & \adaptedhit & \adaptedhit & \monkeymiss & \adaptedhit & 13 \\
            \bottomrule
        \end{tabular}%
        }
    \end{minipage}%
    \hfill
    \begin{minipage}[t]{0.495\linewidth}
        \vspace{0pt}
        \centering
        \textbf{(d) Inference-time scaling}

        \vspace{-10pt}
        \includegraphics[width=\linewidth]{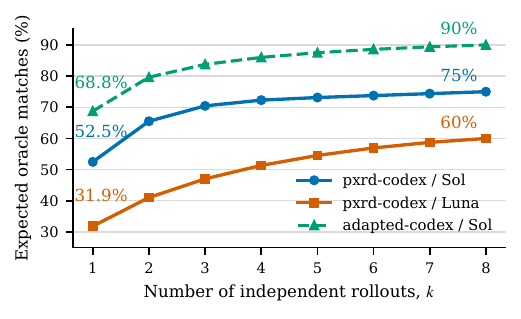}
    \end{minipage}
    \caption{Repeated-sampling results on pXRD-Bench-20. Panels (a)--(c) show candidate-set exact \texttt{StructureMatcher} recovery for \texttt{pxrd-codex} Sol, \texttt{pxrd-codex} Luna, and \texttt{adapted-codex} Sol, respectively. A colored square denotes a candidate-structure match; the rightmost Exact column counts these matches. Panel (d) shows order-invariant expected oracle pass@\(k\), computed by averaging cumulative candidate-set coverage over all \(\binom{8}{k}\) subsets of the observed rollouts.}
    \label{fig:monkey-scaling}
\end{figure}

Panels (a), (b), and (c) in Figure \ref{fig:monkey-scaling} show the results of sampling 8 solution trajectories for each of the 20 pXRD-Bench-20 problems. Some problems are consistently solved or consistently missed, but many are occasionally solved, which allows repeated sampling to consistently recover a correct solution. Our results in Figure \ref{fig:monkey-scaling}(d) shows expected candidate-set match rate scaling with repeated sampling across three initial-performance tiers, with \texttt{adapted-codex} Sol (high) reaching 90\% at pass@8 vs 68.8\% pass@1. These gains require substantial but highly parallelizable test-time compute. Across 160 \texttt{pxrd-codex} attempts per model, Sol averaged 18.5 minutes per rollout (49.4 aggregate agent-hours; 2 timeouts), whereas Luna averaged 29.8 minutes (79.5 agent-hours; 29 timeouts). Each benchmark worker was allocated 2 vCPUs and 5 GB of memory and capped at one hour. Notably, the expected GPT-5.6 Luna (max) pass@5 performance exceeded GPT-5.6 Sol (high) expected pass@1 performance at an estimated 30\% lower expected token cost.

pXRD structure determination is particularly amenable to inference-time scaling because of its \textit{inference-time verifiability}—the measured pattern provides a reference-blind selection signal. Repeated sampling alone improves only oracle coverage---the probability that at least one trajectory produces the correct structure. Realizing this gain requires an inference-time verifier that can recognize the correct candidate without access to the hidden reference. In domains without reliable verifiers, majority-vote and learned-reward selection can saturate even while oracle coverage continues to increase, and scaling best-of-\(N\) selection against an imperfect learned reward can eventually improve the proxy score while degrading ground-truth performance \citep{brown2024large,gao2023scaling}. Here, every proposed structure induces a physical prediction that can be compared directly with the observed diffraction profile. Across eight independent \texttt{adapted-codex} Sol trajectories on our 20-problem panel, the candidate pool contained an exact structure in 18 cases, and a lowest-\(R_{\mathrm{wp}}\) selector with bounded nuisance-profile fitting ranked an exact structure first in all 18. Thus, on this panel, \(R_{\mathrm{wp}}\)-based verification converted the complete oracle pass@8 gain into top-1 performance.

\begin{figure}[!tbp]
    \centering
    \setlength{\parskip}{0pt}
    \begin{minipage}[t]{0.55\linewidth}
        \vspace{0pt}
        \centering
        {\scriptsize\bfseries (a) Inference-time verifier decisions\par}

        \vspace{5pt}
        \resizebox{\linewidth}{!}{%
            \begin{tabular}{@{}lccc@{}}
                \toprule
                Method & Accuracy & False stops & False continues \\
                \midrule
                \(R_{\mathrm{wp}} \leq 5.924\%\) & 127/200 (63.5\%) & 1/59 (1.7\%) & 72/141 (51.1\%) \\
                102-word seed & 126/200 (63.0\%) & 1/59 (1.7\%) & 73/141 (51.8\%) \\
                \(R_{\mathrm{wp}} \leq 6.66\%\) & 149/200 (74.5\%) & 7/59 (11.9\%) & 44/141 (31.2\%) \\
                Final judge-v1 & 155/200 (77.5\%) & 7/59 (11.9\%) & 38/141 (27.0\%) \\
                \bottomrule
            \end{tabular}%
        }
    \end{minipage}%
    \hfill
    \begin{minipage}[t]{0.43\linewidth}
        \vspace{0pt}
        \centering
        {\scriptsize\bfseries (b) Verifier-gated inference scaling\par}

        \vspace{-8pt}
        \includegraphics[width=\linewidth]{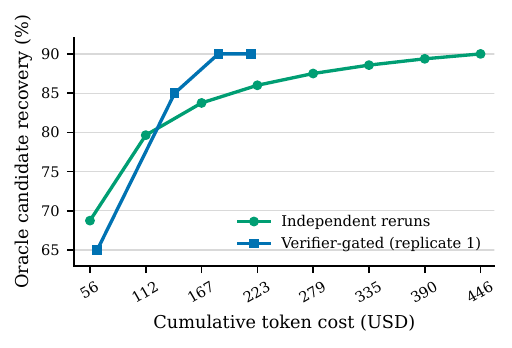}
    \end{minipage}
    \caption{Inference-time verifier accuracy and token-cost-scaled recovery. (a) Decisions on 200 \texttt{adapted-codex} pXRD-Bench trajectories. False stops are incorrect structures judged complete; false continues are exact structures allocated another solve. The \(R_{\mathrm{wp}}\) thresholds were selected retrospectively to match the two LLM policies' false-stop counts. (b) Verifier-gated continuation framework plotted against the 8 independent \texttt{adapted-codex} Sol reruns. The eight independent reruns points are placed at expected token cost for \(k=1,\ldots,8\) rollouts. The verifier-gated curve shows cumulative recovery after cumulative token cost following the initial blind pass, a coverage-directed continuation, and two verifier-authorized fresh solves.}
    \label{fig:verifier-gated-scaling}
\end{figure}

We also utilize an LLM verifier to scale inference-time compute more efficiently. Despite \(R_{\mathrm{wp}}\) being a powerful discriminator between solutions, there is no clear \(R_{\mathrm{wp}}\) cutoff that determines if a solution is correct \citep{toby2006rfactors}. We use an LLM agent to review the provided structures and traces of an initial solve attempt and determine if the structures are likely to be valid solutions, determining \(R_{\mathrm{wp}}\) as well as performing more involved, qualitative analysis. The verifier was tuned using GEPA-style prompt evolution \citep{agrawal2026gepa}, in which a seed policy given in the prompt that emphasized fit quality, physical plausibility, and unresolved structural ambiguity was iteratively revised through reflections on labeled training errors, with the optimization retaining a Pareto frontier of candidate prompts across case-level performance. The final policy was selected on held-out validation trajectories and frozen before benchmark evaluation; full details are provided in Appendix~\ref{app:inference-scaling-details}.

On the 200 \texttt{adapted-codex} pXRD-Bench solves, the evolved-prompt verifier correctly decided stop on 103/141 correct solves and continue on 52/59 missed solves. Compared with the seed prompt, the evolved verifier reduced the false-continue rate from 51.8\% to 27.0\%, while increasing the false-stop rate from 1.7\% to 11.9\% (Figure~\ref{fig:verifier-gated-scaling}a). Retroactively-picked cutoffs on the same bounded-fit \(R_{\mathrm{wp}}\), chosen to match the false stop rate of the LLM verifiers, produce more false continues, despite being generous comparisons rather than realistic de novo verifiers, demonstrating the LLM verifiers make decisions more nuanced than what only considering scalar \(R_{\mathrm{wp}}\) can afford. Figure~\ref{fig:fixedparams-profile-residuals} in Appendix~\ref{app:profile-residual-distributions} shows the presence of structure misses with lower shared fixed-parameter \(R_{\mathrm{wp}}\) than many successful matches.

Additionally, we tested verifier-gated continuation as a more efficient way to leverage inference-time compute than naively repeating solves. Starting from the first 20 \texttt{adapted-codex} solutions on pXRD-Bench-20, cumulative recovery rose from 13/20 to 17/20 after a verifier-authorized coverage-directed solve (given the \texttt{reason} field of the verifier output) and to 18/20 after one verifier-authorized fresh solve; a second fresh solve added no further exact candidate. Including the initial 20 solves, their reviews, and 12 coverage-directed continuations, the first two gated stages cost \$28.97 more than two complete independent passes. As correctly solved problems were subsequently stopped, the third and fourth stages cost \$12.16 and \$23.29 less, respectively, than allocating one additional solve to all 20 problems.

% Experimental multiphase results and discussion.
\FloatBarrier
\section{Agents for experimental multiphase pXRD: learning from self-generated synthetic problems}
\label{sec:experimental-multiphase-adaptation}

In addition to the simulated single-phase problems in pXRD-Bench, we also evaluate agents on experimental multiphase patterns. These patterns arise during powder synthesis experiments and are relevant to materials discovery efforts. In a powder synthesis experiment, the compositions of the starting materials (precursors) are known, but the product of the solid-state reaction can contain contain multiple phases, including the intended product (but not necessarily), residual precursors, intermediates, competing phases, etc. Its characterization must establish which crystalline phases are present, construct defensible atomic models, and estimate their proportions. This information determine whether a synthesis succeeded and what should change in the next experiment, an important piece of feedback in the materials discovery loop.

We study this setting using 15 measured patterns from the Precursor Genome, a collection of solid-state reactions with associated experimental metadata and deposited refinements \citep{walters2026precursorgenome}. The selected family comprises reactions between \(\mathrm{WO_{3}}\) and sources of Mg, Ca, Sr, Ba, Pb, Zn, Cu, Co, Ni, or Mn, fired at \(200\)--\(900\,^\circ\mathrm{C}\). Sources include oxides, carbonates, and a peroxide. The measurements span largely unreacted powders, partially reacted mixtures, and product-dominated samples; 13 of the 15 deposited interpretations are multiphase. We refer to this panel as \texttt{divalent-tungstates-PG15}.

The solving agent receives the measured counts, the two precursor formulas, the firing temperature, and the radiation and instrument description, but not the product composition, the number of phases, a calibrated peak-profile function, or the deposited interpretation. This is intended to mimic the information known to a materials scientist / crystallographer when analyzing a pXRD measurement, a substantial change from the single-phase benchmark: the agent must infer the phase inventory while estimating instrument and specimen effects that influence which phases appear to explain the pattern. Its deliverables are phase CIFs, crystalline weight fractions, a calculated profile, a report, and the final GSAS-II project \citep{toby2013gsasii}. Appendix~\ref{app:multiphase-evolution} gives the baseline solver configuration and task prompt.

\subsection{A reflexive evolution framework that learns from self-generated synthetic problems}
\label{sec:multiphase-framework}

Similar to the agent adaptation to create \texttt{adapted-codex} on pXRD-Bench, we ask whether reflexive evolution can improve agent performance. However, unlike pXRD-Bench problems where representative training problems could be sampled from the same distribution as the evaluation problems, experimental pXRD datasets do not necessarily come with additional representative patterns nor would those patterns come with ground-truth solutions. Real samples don't have ground truth CIFs—an expert-determined structural model, if available, is the closest substitute and may still be imperfect. We develop a framework for this setting, allowing agents to improve performance beyond their initial baselines and the deposited expert-determined solutions. We ask whether synthetic problems, for which the generating structures and measurement parameters are known, allow an evolver/trainer to utilize its access to the ground-truth solutions to learn prose guidance and create executable tools that improve analysis of experimental mixtures when provided to agents. Our framework learns persistent guidance and optional code from these problems while keeping the language-model weights fixed.

\begin{figure}[tbp]
  \centering
  \includegraphics[width=\textwidth]{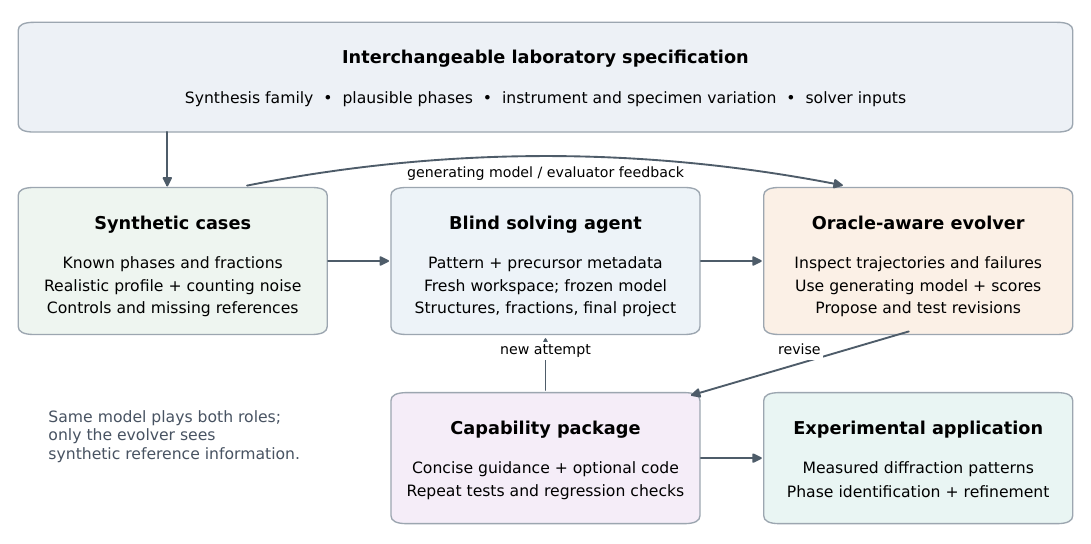}
  \caption{\textbf{Synthetic experience for experimental characterization.} A laboratory specification defines the scientific setting and permitted synthetic variation. Fresh solver workspaces receive only public measurements and metadata. A persistent evolver can inspect trajectories and use synthetic reference information to diagnose failures, revise a capability package, and test it on further solves. The frozen package is subsequently applied to measured patterns. Ground-truth access is confined to synthetic research and evaluation; experimental fits retain the solving agent's own model and calibration.}
  \label{fig:multiphase-framework}
\end{figure}

The framework separates the adaptation procedure from a replaceable laboratory specification (Figure~\ref{fig:multiphase-framework}). Groups of samples synthesized and characterized in a (potentially autonomous) materials research lab often form a well-defined family, linked by a shared structural prototype, a common synthesis route, or a target chemical class—as with the divalent metal tungstates in \texttt{divalent-tungstates-PG15}. The laboratory specification in our framework allows a researcher to define the sample family, precursors, chemically plausible outcomes, measurement geometry, ranges for instrument / specimen effects, and information available to a solver. For the tungstate family, it describes nominally 1:1 metal-to-tungsten precursor mixtures, a single firing step, and laboratory Cu K\(\alpha\) Bragg--Brentano measurements. Paired with a synthetic generator, an evolver can curate and create synthetic training structural models within the laboratory scope. In the case of our experiments, the synthetic generator combines COD structures \citep{grazulis2012cod} into chemically balanced mixtures and simulates patterns with the researcher-defined counting noise and varied instrument and specimen effects. By changing the laboratory specification and sample generator, agents can undergo a reflexive learning process for a new experimental setting and sample family without rewriting the evolution mechanisms.

A persistent evolver agent curates six initial problems and runs three independent baseline attempts on each. It reads the resulting scripts, projects, trajectories, and verifier diagnostics, then proposes a small package of guidance and, when useful, executable helpers. Its role extends beyond rewriting a prompt: it can inspect failed numerical procedures, perform controlled tests on completed projects, and use the known generating structural model and scorer (evaluation-time verifiability) to identify what the solver missed. The solver workspaces remain reference-blind. Both roles use GPT-5.6 Sol at maximum reasoning effort; a stronger teacher model is not introduced, meaning our results should apply to the frontier of released models.

The adaptation protocol screens evolved packages with one solve per active problem. Cases that succeed can be set aside temporarily while unresolved cases receive further revisions, followed by regression checks. A promising package is evaluated with three solves per case, a choice made due to observed rollout variability. The initial run allows 30 development solves per cycle, with additional baseline, transfer, and evaluation repeats accounted for separately. At budget exhaustion, a package is promoted based on its improvement on the scores compared to the baseline or incumbent package, and new synthetic problems are curated for further evolution and learning. Details about the evolution/algorithm can be found in the repo.

\paragraph{What the synthetic verifier measures.}
The quality of a solution in the setting of pXRD can be assessed both in terms of profile agreement—how well the structural solution's predicted diffraction pattern reproduces the input—and structural agreement—how close the structures and phase weights are to the generating ground-truth. These criteria are complementary. Weak sensitivity to light-atom positions, overlapping reflections, and weak minor-phase contributions can leave multiple models consistent with the available diffraction information. Evaluation based solely on correspondence with the generating model may therefore penalize plausible alternatives that the measurement cannot distinguish. Conversely, a lower profile residual does not necessarily indicate a more accurate description of the sample: additional phases or flexible background, texture, and structural parameters can absorb noise or compensate for an incorrect model. Our scorer design assesses profile agreement alongside recovery of the known synthetic solution and the physical plausibility of the inferred sample model. 

The verifier evaluates the submitted GSAS-II project without further refinement, retaining the solver's structures, background, and instrument and specimen parameters. Profile agreement is measured against the hidden noise-free generating pattern using mean Poisson deviance \(D\); a structural score \(S\in[0,1]\) assesses phase identities, atomic structures, and weight fractions. The combined score is \(Q=100[0.6/(1+D)+0.4S]\), subject to physical-validity checks. While the solver receives only noisy observations, the evolver can use these reference-based diagnostics to distinguish missing phases and structural errors from inadequate profile modeling.

% Reserve the bottom float before the subsection heading so small text edits do
% not defer the evolution figure to a separate float-only page.
\begin{figure}[!b]
  \centering
  \includegraphics[width=0.94\textwidth]{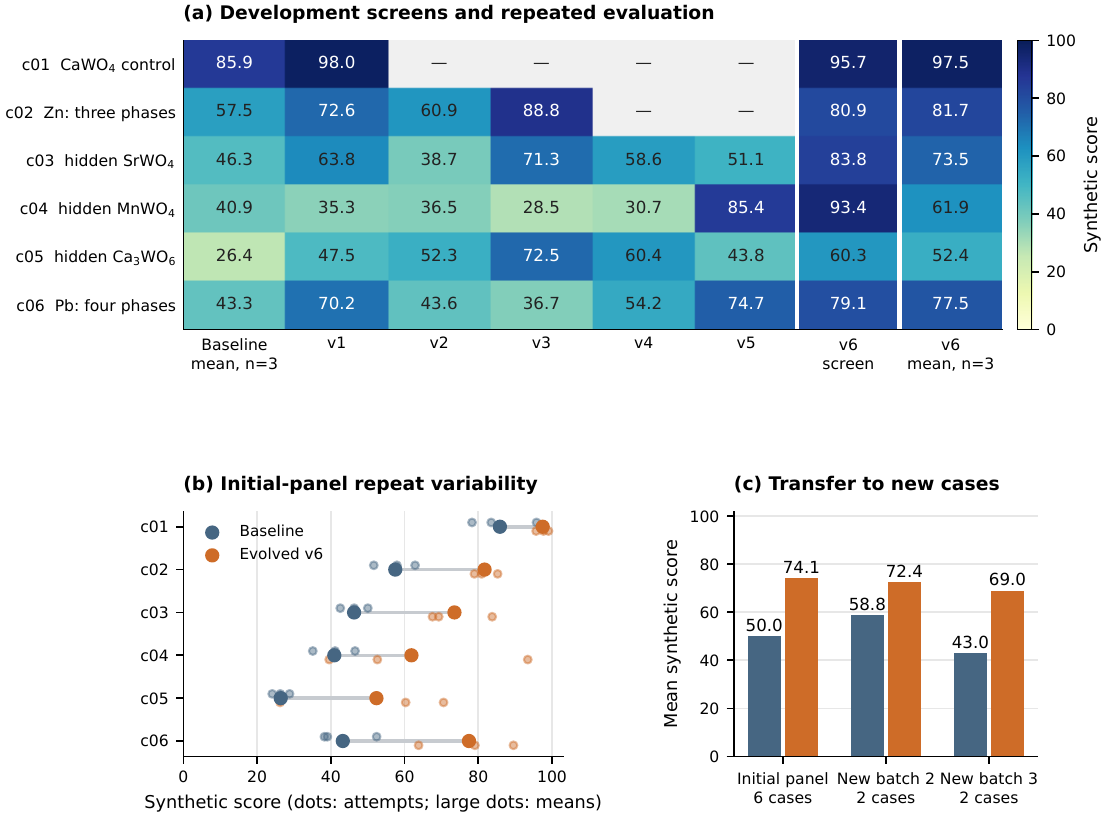}
  \caption{\textbf{Evolution improved the initial curriculum but was neither monotonic nor uniformly reliable.} (a) Case-level scores for the baseline, single-attempt revision screens, and the retained v6 package. Gray cells were not evaluated under that revision. The v6 mean includes its initial screen and two additional attempts per case; it is a repeated development evaluation, not an independent held-out test. (b) Individual baseline and v6 scores, with larger dots denoting case means. (c) Initial-panel means and prospective comparisons on two subsequent pairs of new synthetic cases, each with three attempts per condition and no package edits before that comparison. Scores are the composite \(Q\).}
  \label{fig:multiphase-evolution}
\end{figure}

\subsection{Evolution of phase-recovery and refinement procedures}
\label{sec:multiphase-november}

The initial evolver-curated curriculum comprised six synthetic problems spanning single- and multiphase samples, low counts, weak residual phases, and deliberately withheld reference structures. Across the full run, the evolver completed 135 solver attempts on ten synthetic problems and produced 15 package revisions.

The first revision addressed a phase-identification failure: the baseline missed the withheld \(\mathrm{Ca_{3}WO_{6}}\) intermediate in all three attempts. The evolver introduced a helper that searches for chemically analogous structural prototypes, allowing the solver to construct candidates when an exact reference is unavailable. Subsequent revisions addressed coupled errors in phase selection and profile modeling. Ground-truth-assisted tests identified misassigned broadening as a major source of error, but translating this diagnosis into reliable solver guidance required several revisions and initially caused regressions. The retained package, v6, combined the prototype-search helper with 331 words of guidance on source-structure selection, reconsideration of weak phases, instrument and specimen effects, and validation of saved solutions.

On the six development problems, v6 increased the mean composite score from 50.04 to 74.07 across three attempts per problem, improving every case mean and recovering the complete generating phase inventory in 17 of 18 attempts (Figure~\ref{fig:multiphase-evolution}). Its initial single-attempt screen had scored 82.19, illustrating the variability obscured by evaluating each revision only once. v6 was promoted at the 30-solve development-budget limit.

The next 2 batches each introduced two new problems, evaluated with the unchanged Baseline and Evolved v6 conditions. Mean scores increased from 58.79 to 72.44 and from 43.02 to 68.97, providing evidence of transfer beyond the initial curriculum. Further revisions produced local gains and regressions, but none replaced v6. Details on the evolution framework including scoring formula, generator implementation, and package promotion rules are provided in Appendix~\ref{app:multiphase-evolution}; evolver-written notes, execution logs, and the entire implementation are available in the repository.

\FloatBarrier
\subsection{Evolved agent performance on \texttt{divalent-tungstates-PG15}}
\label{sec:multiphase-transfer}

We evaluated the frozen v6 package on the 15 measured patterns, with three independent attempts per measurement. The Baseline and Evolved v6 conditions use the same GPT-5.6 Sol model alias at maximum reasoning effort, baseline task prompt, Codex harness, crystallographic software, and reference library; Evolved v6 additionally receives the v6 guidance and prototype-search helper. Each condition therefore contributes 45 attempts. We compare their results with one deposited PG solution per measurement, withheld from both agents.

% Queue the transfer plot early enough to share the opening results page.
\begin{figure}[!t]
  \centering
  \includegraphics[width=\textwidth]{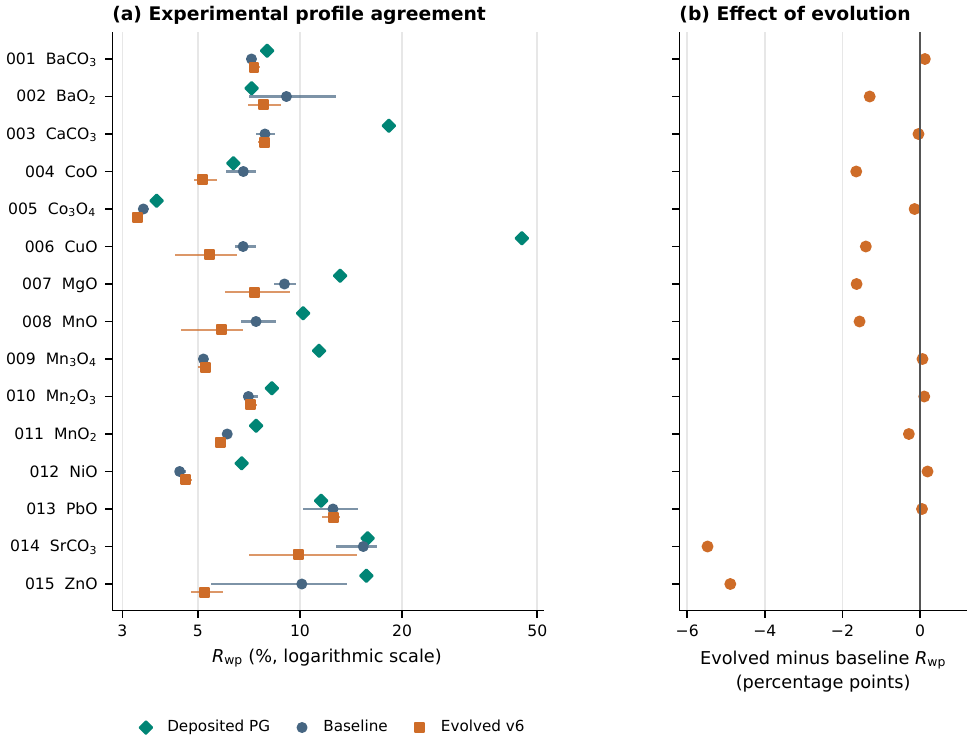}
  \caption{\textbf{Experimental transfer of the v6 package.} (a) Agent points show the mean verified \(R_{\mathrm{wp}}\) for each measurement; horizontal lines span the observed attempts, not a confidence interval. Each PG point scores the saved deposited profile. The horizontal axis is logarithmic. Labels give the case and second precursor; every reaction also contains \(\mathrm{WO_{3}}\). (b) Change in the measurement-level mean from baseline to v6; negative values indicate improved profile agreement. Only the initial three attempts per measurement are included.}
  \label{fig:multiphase-experimental}
\end{figure}

\paragraph{Uniform evaluation.}
For each attempt, the evaluator reopens the final GSAS-II project and calculates its profile without further refinement, retaining the submitted structures, background, calibration, and specimen parameters. We refer to the independently recomputed residual as ``verified'' \(R_{\mathrm{wp}}\); this verifies the reported profile agreement, not the structural interpretation. We score these replayed profiles and the saved PG calculated profiles against the same measured counts over \(10.1^\circ\)--\(99.9^\circ\) in \(2\theta\), using \(R_{\mathrm{wp}}\) with \(w_i=1/\max(y_i,1)\). The comparison measures profile agreement under each analysis's fitted model. The deposited models provide a useful baseline and comparison but not a structural ground truth. For each agent condition, we average verified attempts within a measurement before calculating the mean and median across the 15 measurements. Appendix~\ref{app:multiphase-experimental-results} gives the per-attempt scores and evaluation details.

% Generated by scripts/plot_experimental_transfer.py.
\newcommand{\experimentalTransferPrimaryRows}{%
Deposited PG & 12.59 & 10.22 & \textemdash & \textemdash & \textemdash \\
Baseline & 7.90 & 7.20 & 6.92 & 1.04 & 27.0 \\
Evolved v6 & 6.71 & 5.86 & 6.04 & 0.80 & 50.2 \\
}

\begin{table}[htbp]
  \centering\small
  \setlength{\tabcolsep}{4pt}
  \caption{\textbf{Transfer from synthetic experience to measured patterns.} Mean and median summarize the 15 measurement-level mean residuals. Best-of-three averages the lowest \(R_{\mathrm{wp}}\) among three attempts on each measurement. Within-scan SD is the mean sample standard deviation across measurements, in percentage points (pp). Time is mean wall time per attempt. Lower residuals indicate better profile agreement.}
  \label{tab:multiphase-experimental}
  \begin{tabularx}{\textwidth}{@{}>{\raggedright\arraybackslash}Xccccc@{}}
    \toprule
    Condition & \shortstack{Mean \(R_{\mathrm{wp}}\)\\(\%)}
      & \shortstack{Median\\(\%)}
      & \shortstack{Mean best-of-3\\(\%)}
      & \shortstack{Within-scan\\SD (pp)}
      & \shortstack{Mean time\\(min)} \\
    \midrule
    \experimentalTransferPrimaryRows
    \bottomrule
  \end{tabularx}
\end{table}

The evolved package lowers mean \(R_{\mathrm{wp}}\) from 7.90\% to 6.71\%, a relative reduction of 15.1\%, and improves the measurement-level mean on 10 of 15 scans (Table~\ref{tab:multiphase-experimental}, Figure~\ref{fig:multiphase-experimental}). The largest improvements occur on Sr and Zn. Mean duration increases from 27.0 to 50.2 minutes, so the package changes both the analysis and the amount of computation the agent devotes to it. The deposited profiles have a mean residual of 12.59\% and a median of 10.22\%; their mean is particularly affected by the Cu deposit's 45.05\% residual. v6 has a lower mean residual than the deposit on 13 of the 15 measurements, although the phase inventories and fitted parameters also differ.

\begin{figure}[p]
  \centering
  \includegraphics[width=\textwidth]{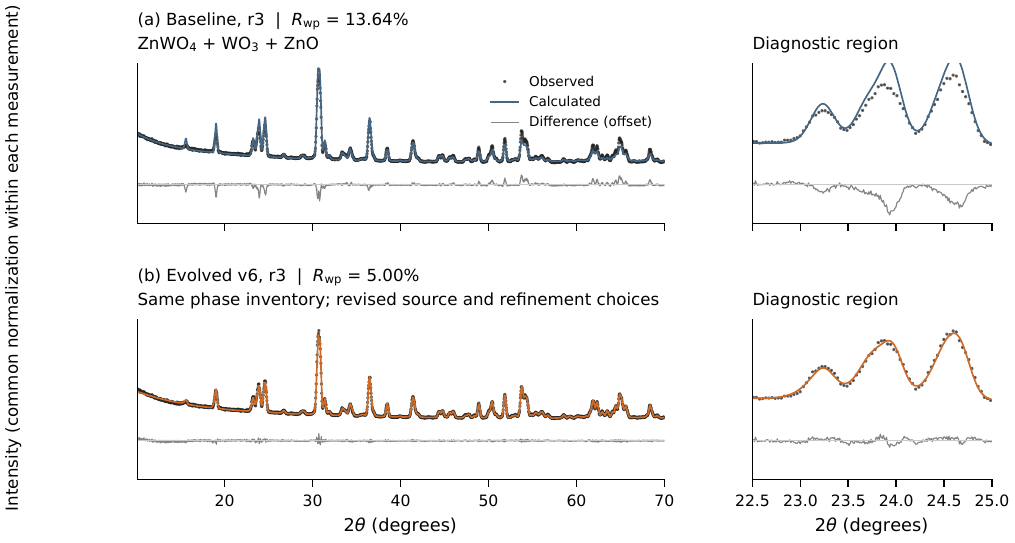}
  \caption{\textbf{Same phase inventory, improved profile agreement: the Zn measurement.} Baseline and evolved attempt 3 on scan 015, shown as an illustrative trajectory comparison rather than a best-of-three selection. All curves are taken from saved independent project replays. Observations are black points, calculations are colored lines, and differences are gray with a vertical offset. The right panels resolve the \(\mathrm{WO_{3}}\)-rich region. Both fits contain \(\mathrm{ZnWO_{4}}\), \(\mathrm{WO_{3}}\), and \(\mathrm{ZnO}\); the evolved attempt changes the source structure and profile treatment. Intensities share the observed maximum as a display normalization, while reported \(R_{\mathrm{wp}}\) uses raw counts over the full evaluation window.}
  \label{fig:multiphase-zn}
\end{figure}

\paragraph{Refinement of an established phase inventory.}
The Zn measurement (015) illustrates how v6 changes the refinement procedure (Figure~\ref{fig:multiphase-zn}). Both agents identify \(\mathrm{ZnWO_{4}}\), \(\mathrm{WO_{3}}\), and residual \(\mathrm{ZnO}\). The baseline's three residuals span 5.49--13.64\%, whereas v6's span 4.82--5.89\%. In the illustrated v6 attempt, the agent replaces a coarse \(\mathrm{WO_{3}}\) source containing a short O--O contact with a neutron-derived record, uses an eight-coefficient background, and tests crystallite size and microstrain separately. It retains a \(\mathrm{WO_{3}}\) microstrain term and removes an unstable size contribution. These changes follow the package's guidance to screen source structures and distinguish common profile errors from phase-specific effects. The trajectory supports a more systematic refinement procedure and better profile agreement.

% Keep the detailed Zn profiles and phase-inventory table on one float page.

\paragraph{Phase recovery and disagreements with the deposited interpretations.}
The Sr measurement (014) provides a different kind of gain. The deposited \(\mathrm{Sr_{2}WO_{5}}\) structure is absent from the supplied reference library. In two of its three attempts, v6 constructs a matching model by substituting into a \(\mathrm{Ba_{2}ReO_{5}}\) prototype retrieved with the learned helper. These attempts assign 24.9--25.9 wt\% to the phase, close to the deposited 24.8 wt\%, and achieve residuals of 7.12\% and 7.88\%. The baseline recovers it in none of its three attempts. 

% Generated by scripts/plot_experimental_transfer.py.
\newcommand{\experimentalTransferInventoryRows}{%
003 (Ca) & CaCO$_{3}$: 13.5 wt\%\newline $R_{\mathrm{wp}}=18.28\%$ & CaCO$_{3}$: 3/3\newline 27.7--33.8 wt\%\newline $\overline{R}_{\mathrm{wp}}=7.89\%$ & CaCO$_{3}$: 3/3\newline 28.4--30.2 wt\%\newline $\overline{R}_{\mathrm{wp}}=7.84\%$ \\
\addlinespace[4pt]
006 (Cu) & 7 phase models; CuWO$_{4}$: 68.1 wt\%\newline $R_{\mathrm{wp}}=45.05\%$ & 2 phase models in 3/3\newline CuWO$_{4}$: 97.7--97.9 wt\%\newline $\overline{R}_{\mathrm{wp}}=6.80\%$ & 2 phase models in 3/3\newline CuWO$_{4}$: 96.8--97.5 wt\%\newline $\overline{R}_{\mathrm{wp}}=5.40\%$ \\
\addlinespace[4pt]
007 (Mg) & MgO: absent\newline $R_{\mathrm{wp}}=13.12\%$ & MgO: 3/3\newline 7.0--7.6 wt\%\newline $\overline{R}_{\mathrm{wp}}=9.00\%$ & MgO: 3/3\newline 7.1--7.4 wt\%\newline $\overline{R}_{\mathrm{wp}}=7.37\%$ \\
\addlinespace[4pt]
008 (Mn) & WO$_{3}$: absent\newline $R_{\mathrm{wp}}=10.22\%$ & WO$_{3}$: 3/3\newline 2.0--2.1 wt\%\newline $\overline{R}_{\mathrm{wp}}=7.42\%$ & WO$_{3}$: 3/3\newline 2.0--3.3 wt\%\newline $\overline{R}_{\mathrm{wp}}=5.86\%$ \\
\addlinespace[4pt]
013 (Pb) & Pb$_{3}$WO$_{6}$: 15.3 wt\%\newline $R_{\mathrm{wp}}=11.54\%$ & Pb$_{3}$WO$_{6}$: 0/3\newline $\overline{R}_{\mathrm{wp}}=12.51\%$ & Pb$_{3}$WO$_{6}$: 0/3\newline $\overline{R}_{\mathrm{wp}}=12.56\%$ \\
\addlinespace[4pt]
014 (Sr) & Sr$_{2}$WO$_{5}$: 24.8 wt\%\newline $R_{\mathrm{wp}}=15.81\%$ & Sr$_{2}$WO$_{5}$: 0/3\newline $\overline{R}_{\mathrm{wp}}=15.37\%$ & Sr$_{2}$WO$_{5}$: 2/3\newline 24.9--25.9 wt\%\newline $\overline{R}_{\mathrm{wp}}=9.89\%$ \\
\addlinespace[4pt]
015 (Zn) & ZnO: absent\newline $R_{\mathrm{wp}}=15.70\%$ & ZnO: 3/3\newline 6.9--7.6 wt\%\newline $\overline{R}_{\mathrm{wp}}=10.13\%$ & ZnO: 3/3\newline 7.1--7.5 wt\%\newline $\overline{R}_{\mathrm{wp}}=5.24\%$ \\
\addlinespace[4pt]
}

\begin{table}[p]
  \centering\footnotesize
  \setlength{\tabcolsep}{4pt}
  \renewcommand{\arraystretch}{1.08}
  \caption{\textbf{Selected differences in phase assignment and quantity.} Agent entries give the number of initial attempts containing the indicated phase and the range of its reported crystalline weight fraction where present. \(\overline{R}_{\mathrm{wp}}\) averages all three attempts, including those without the phase. The Cu row instead gives the total number of phase models and the dominant-phase fraction. Sr and Pb presence requires a structural match to the deposited reference structure; other rows use the submitted phase formula. The table highlights substantial disagreements rather than listing every trace phase.}
  \label{tab:multiphase-inventories}
  \begin{tabularx}{\textwidth}{@{}l *{3}{>{\raggedright\arraybackslash}X}@{}}
    \toprule
    Case & Deposited PG & Baseline & Evolved v6 \\
    \midrule
    \experimentalTransferInventoryRows
    \bottomrule
  \end{tabularx}
\end{table}

Table~\ref{tab:multiphase-inventories} summarizes selected differences in phase inventory or fraction among the baseline, v6, and PG deposits. On Cu, both agents consistently fit two phase models, dominated by \(\mathrm{CuWO_{4}}\), whereas the deposit contains seven. Appendix Figure~\ref{fig:multiphase-cu-deposited} compares their calculated profiles. On Mg and Zn, both agents repeatedly include residual precursor oxides absent from the deposits; on Ca, all three interpretations include carbonate, but the agents assign about twice its deposited fraction. Conversely, neither baseline nor v6 recovers the deposited \(\mathrm{Pb_{3}WO_{6}}\) structure. Differences between recovered phase inventories don't indicate for certain which (or if any) of the solutions are correct, but a lower \(\mathrm{R_{wp}}\) provides a useful signal.

\FloatBarrier
\subsection{Framework access and disentangling the value of reflexive evolution}
\label{sec:multiphase-mechanisms}

Despite the encouraging gains resulting from our reflexive evolution on self-generated problems framework, we run additional experiments to understand what gains can be attributed to the learning process and what gains can be attributed to the additional information included in the evolution framework but unavailable to the baseline solvers. The \emph{Framework access} condition gives the baseline read-only access to the framework's starting directory: the laboratory description, generator and scorer code, and other process specification / infrastructure. Two further conditions separate combining these resources from evolving the framework-access agent itself. \emph{Framework + v6} supplies framework access together with the original v6 package. \emph{Evolved framework v5} supplies framework access with a 205-word v5 package from a separate evolution run that began from the framework-access baseline. That package contains no additional helper and was not promoted by its development gate; its experimental evaluation is exploratory. All five conditions share the model alias, reasoning effort, baseline task prompt, harness, scientific software, measurements, and reference library.

\begin{table}[tbp]
  \centering\small
  \setlength{\tabcolsep}{5pt}
  \caption{\textbf{Scientific context and evolved guidance.} Summary conventions follow Table~\ref{tab:multiphase-experimental}. Best-of-three averages the lowest \(R_{\mathrm{wp}}\) among three attempts on each of the 15 measurements; the later Pb/Sr repeats are excluded.}
  \label{tab:multiphase-conditions}
  \begin{tabularx}{\textwidth}{@{}Xcccc@{}}
    \toprule
    Condition & \shortstack{Mean \(R_{\mathrm{wp}}\)\\(\%)}
      & \shortstack{Median\\(\%)}
      & \shortstack{Mean best-of-3\\(\%)}
      & \shortstack{Mean time\\(min)} \\
    \midrule
    Baseline & 7.90 & 7.20 & 6.92 & 27.0 \\
    Evolved v6 & 6.71 & 5.86 & 6.04 & 50.2 \\
    Framework access & 5.98 & 5.49 & 5.47 & 66.1 \\
    Framework + v6 & 6.49 & 6.36 & 5.68 & 67.0 \\
    Evolved framework v5$^\dagger$ & 6.44 & 6.15 & 5.65 & 43.8 \\
    \bottomrule
  \end{tabularx}
  \begin{minipage}{\textwidth}\footnotesize
  \vspace{4pt}
  \(^{\dagger}\)v5 was evaluated despite not meeting its synthetic promotion criteria. Experimental runs used the same model alias on different dates and were not constrained to equal wall time.
  \end{minipage}
\end{table}

\begin{figure}[tbp]
  \centering
  \includegraphics[width=\textwidth]{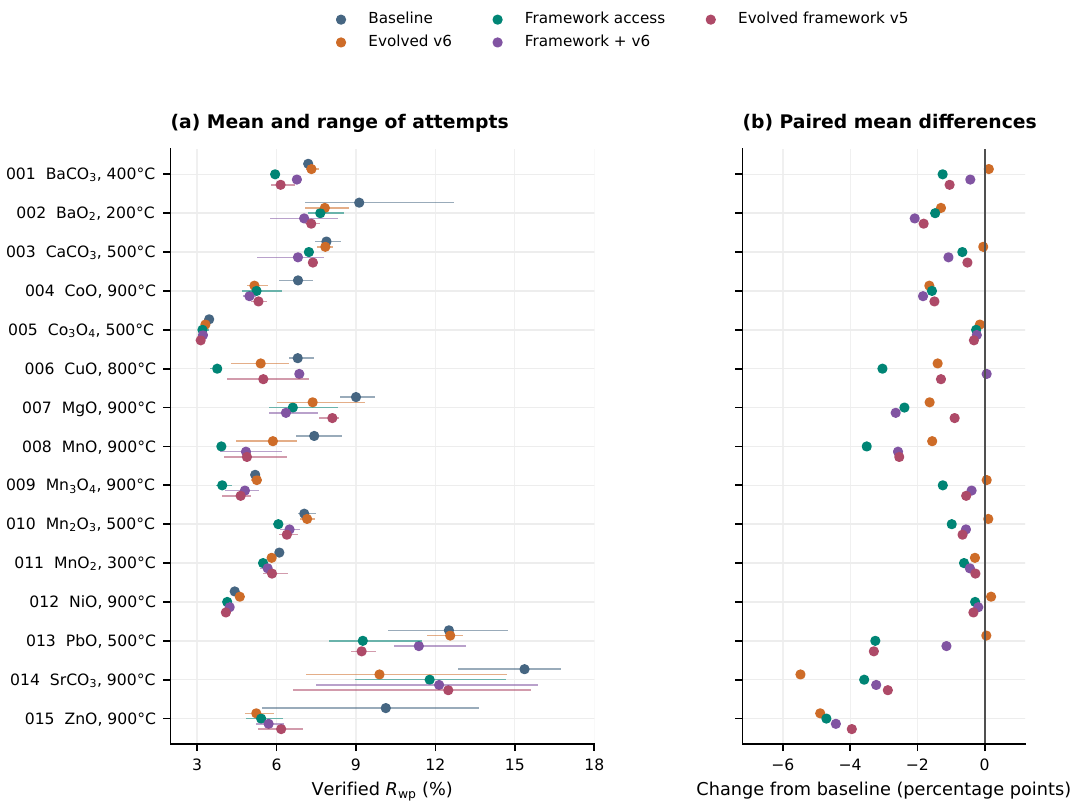}
  \caption{\textbf{The effect of framework access varies across measurements.} (a) Points show the mean verified \(R_{\mathrm{wp}}\); horizontal lines span the observed attempts. Labels give the second precursor and firing temperature; every reaction also contains \(\mathrm{WO_{3}}\). (b) Differences from the baseline's measurement-level mean. The framework-access condition improves on baseline on every scan, but is not the best condition on every scan. Only the initial three attempts per measurement are included.}
  \label{fig:multiphase-ablation}
\end{figure}

Framework access lowers mean \(R_{\mathrm{wp}}\) to 5.98\%; it improves on the baseline on all 15 measurements and on v6 alone on 12 (Table~\ref{tab:multiphase-conditions}, Figure~\ref{fig:multiphase-ablation}). Adding the v6 guidance/tools to framework access raises the mean to 6.49\%, and the separately evolved v5 yields 6.44\%. Both improve on framework access on only four measurements. Thus the best average fit in this comparison comes from direct access to the scientific context and code, without an evolved package. We first examine the framework-access gains and their origin. Section~\ref{sec:multiphase-phase-recovery} then highlights a phase-recovery capability elicited by evolution despite poorer average profile agreement, and Section~\ref{sec:multiphase-nonpromotion} discusses the value of reflexive evolution for pXRD and in general, considering the relative results of each evaluated condition. 

The framework-access trajectories show active use of the supplied material, including inspection of the simulator and evaluator and execution of reference-blind replay utilities. On several measurements the final phase inventory is much the same as in the baseline, but the agent tests a more adequate profile or structural model. For the Cu measurement (006), mean \(R_{\mathrm{wp}}\) falls from 6.80\% for Baseline to 5.40\% for Evolved v6 and 3.76\% with framework access. The dominant assignment remains \(\mathrm{CuWO_{4}}\) with a small \(\mathrm{WO_{3}}\) contribution. On Mn measurements 008 and 009, framework access similarly reaches means of 3.92\% and 3.95\%, with changes involving atomic parameters, broadening, and low-angle intensity corrections.

The final GSAS-II projects provide a second view of these behavioral changes. The median number of background coefficients is 15 for Baseline, 8 for Evolved v6, and 16 for framework access. The axial-asymmetry parameter \(SH/L\) differs from its standard 0.002 starting value in 12/45, 17/45, and 44/45 projects, respectively. These are observations about submitted models, not isolated causal effects: a parameter can be fitted and then fixed, and a more flexible background can reduce \(R_{\mathrm{wp}}\) without resolving a missing phase. Nevertheless, the two interventions lead to visibly different refinement choices. v6 consistently favors a restrained background and explicit tests of physical contributions, possibly due to the synthetic patterns being constrained to what the generator can create, leading to tendencies that don't transfer positively to real patterns; framework access more often encourages broad exploration of the available model.

The framework-access results suggest that much of the improvement over the baseline comes from giving the solver access to scientific knowledge already encoded in the experimental setup, rather than learning new procedures through evolution. The generator exposes implementations of background, peak broadening, and specimen effects, while the replay utilities provide a way to check the resulting fits. The trajectories show agents inspecting and using this code to test changes to their structural and profile models. v6 distills some of the same information, together with experience from synthetic rollouts, into a short set of recommendations. This improves the baseline, but may also favor refinement choices suited to the synthetic curriculum rather than the measured patterns. Direct framework access leaves the solver more freedom to decide which assumptions and operations are useful for each measurement while still providing useful context about the samples and success measures. Having access to this information induces longer solves and more persistence/care, even more so than the guidance learned by the reflexive evolution process.

\subsubsection{Transfer of a missing-phase recovery procedure}
\label{sec:multiphase-phase-recovery}

\begin{figure}[tbp]
  \centering
  \includegraphics[width=\textwidth]{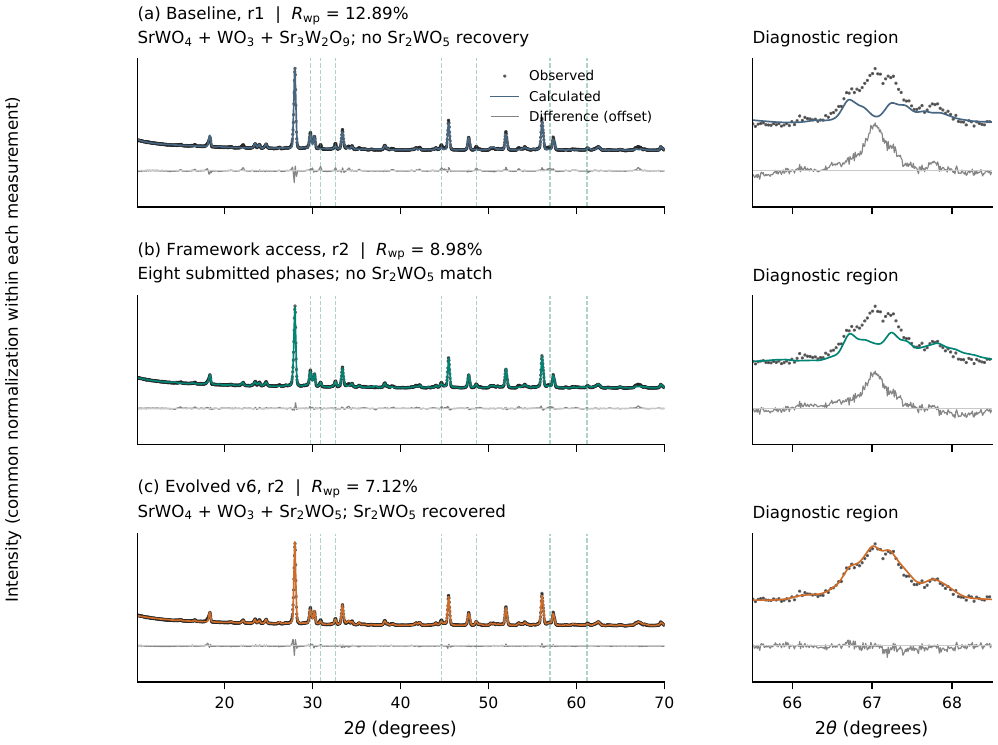}
  \caption{\textbf{The Sr measurement exposes a difference in the hypotheses explored.} The illustrated baseline r1 and framework-access r2 are their lowest-\(R_{\mathrm{wp}}\) attempts in the original three-repeat panels; evolved r2 is the successful prototype-transfer example discussed in the text. Framework access fits several weak features using a larger phase set but does not recover \(\mathrm{Sr_{2}WO_{5}}\). The evolved attempt constructs \(\mathrm{Sr_{2}WO_{5}}\) from a \(\mathrm{Ba_{2}ReO_{5}}\) prototype and fits a three-phase mixture. Dashed lines mark diagnostic reflection positions identified in that trajectory, not a separate reference-assisted fit. All plotted profiles are frozen project replays; display and residual conventions follow Figure~\ref{fig:multiphase-zn}. The repeat frequencies in Table~\ref{tab:multiphase-recovery} provide context for these selected examples.}
  \label{fig:multiphase-sr}
\end{figure}

Average profile agreement does not capture the clearest additional contribution of evolution. The deposited PG solution for the Sr measurement (014) includes a \(\mathrm{Sr_{2}WO_{5}}\) phase whose exact structure is absent from the supplied reference library. Neither the original baseline nor framework-access attempts recover a structurally matching \(\mathrm{Sr_{2}WO_{5}}\) model, but solvers with packages learned from the evolution process were able to discover this phase and achieve significantly lower \(R_{\mathrm{wp}}\) compared to even the best solutions without the \(\mathrm{Sr_{2}WO_{5}}\) phase. In a successful evolved attempt, coherent residual reflections motivate a search across Sr-rich tungstate stoichiometries. The agent uses the learned helper to retrieve orthorhombic \(\mathrm{Ba_{2}ReO_{5}}\) (COD 1529756), substitutes Ba\(^{2+}\) with Sr\(^{2+}\) and Re\(^{6+}\) with W\(^{6+}\), and tests the resulting \(\mathrm{Sr_{2}WO_{5}}\) model. Its scale is positive before cell refinement. The final three-phase interpretation contains approximately 66.1 wt\% \(\mathrm{SrWO_{4}}\), 24.9 wt\% \(\mathrm{Sr_{2}WO_{5}}\), and 9.0 wt\% \(\mathrm{WO_{3}}\), with verified \(R_{\mathrm{wp}}\) of 7.12\% (Figure~\ref{fig:multiphase-sr}). The submitted \(\mathrm{Sr_{2}WO_{5}}\) structure matches the deposited reference structure under the evaluator's structural criterion.

This is a transfer of a procedure, not memorization of the experimental phase. The synthetic failure that motivated the helper involved a \(\mathrm{Ca_{3}WO_{6}}\) intermediate and a \(\mathrm{Ca_{3}OsO_{6}}\) parent; the experimental solve requires a different stoichiometry and a different substituted prototype. The helper makes a composition-led route through the structural library explicit. It neither supplies the target CIF nor resolves the mixture without further hypothesis testing and refinement. The agent's own report also distinguishes the strongly supported phase and cell from prototype-inherited coordinates and reports that its fraction uncertainty exceeds the formal least-squares error.

% Keep the Sr repeat table at the top of the following page.
\suppressfloats[t]
\begin{table}[t]
  \centering\small
  \setlength{\tabcolsep}{3pt}
  \caption{\textbf{Profile agreement and phase recovery on the Sr measurement (014).} Entries are verified \(R_{\mathrm{wp}}\) values (\%). Bold entries indicate a submitted structure matching the deposited \(\mathrm{Sr_{2}WO_{5}}\) phase; Total counts matches among the nine completed solves. Initial attempts belong to the full-panel experiment; added attempts were collected after inspecting those results. All repeats concern the same specimen.}
  \label{tab:multiphase-recovery}
  % Initial values: data/experimental_transfer/records.json, case-014.
  % Added values and matches: analysis/per_attempt.json in the raw PG15 runs
  % pxrd-pg15-recovery-supplement-20261109,
  % pxrd-pg15-recovery-supplement2-20261109, and
  % pxrd-pg15-recovery-supplement3-20261111.
  \begin{tabularx}{\textwidth}{@{}>{\raggedright\arraybackslash}X*{10}{c}@{}}
    \toprule
    & \multicolumn{3}{c}{Initial attempts} & \multicolumn{6}{c}{Added attempts} & \\
    \cmidrule(lr){2-4}\cmidrule(lr){5-10}
    Condition & 1 & 2 & 3 & 4 & 5 & 6 & 7 & 8 & 9 & Total \\
    \midrule
    Baseline & 12.89 & 16.48 & 16.72 & 13.99 & 13.53 & 12.54 & 18.68 & 16.71 & 15.27 & 0/9 \\
    Evolved v6 & 14.66 & \textbf{7.12} & \textbf{7.88} & 12.97 & 12.90 & 14.67 & \textbf{7.32} & 15.77 & 15.34 & 3/9 \\
    Framework access & 14.65 & 8.98 & 11.73 & 11.13\(^{\dagger}\) & 11.14 & 11.36 & 13.28 & 10.21 & 12.90 & 0/9 \\
    Framework + v6 & 15.85 & 13.04 & \textbf{7.53} & 13.44 & \textbf{7.19} & \textbf{6.82} & \textbf{6.93} & \textbf{13.40} & 13.08 & 5/9 \\
    Evolved framework v5 & \textbf{6.64} & 15.23 & 15.59 & 15.22 & 14.38 & 12.97 & 13.07 & 13.18 & 12.93 & 1/9 \\
    \bottomrule
  \end{tabularx}
  \begin{minipage}{\textwidth}\footnotesize
  \vspace{4pt}
  \(^{\dagger}\)For framework access, a 10th attempt replaces attempt 4, which produced no submission.
  \end{minipage}
\end{table}

\paragraph{Targeted repeats.}
After observing recovery differences in the initial panels, we collected six additional attempts for every condition on the Sr measurement, reported on Table \ref{tab:multiphase-recovery}. The repeats demonstrate how trends in the initial 3 attempts are not fully prescriptive—v6 recovers the phase in 3/9 attempts rather than the initial 2/3, the combined framework-plus-v6 condition recovers it in 5/9, the independently evolved framework-access v5 recovers it in 1/9 despite the initial 1/3. However, a clear behavioral trend persists: the baseline and framework access agents both recover the \(\mathrm{Sr_{2}WO_{5}}\) phase in 0/9 attempts, with no attempts reaching below 10 in \(\mathrm{R_{wp}}\) (besides an outlier 8.98 that fit 8 phases), while each of conditions that included evolved packages were able to discern the \(\mathrm{Sr_{2}WO_{5}}\) phase at varying rates and produced solutions with \(\mathrm{Sr_{2}WO_{5}}\) clustered around 6-8 \(\mathrm{R_{wp}}\) (besides an outlier at 13.40 that got trapped in a poor lattice/mixture during refinement).

This points to the reflexive learning from self-generated synthetic problems process being able to elicit helpful behaviors that extend beyond the additional context/knowledge embedded in the evolution framework. Oracle-knowledge of answers to the synthetic problems allows the same model as the solvers (equal intelligence) to reason about why the solvers fail and design guidance and tools to push them towards the approaches that would lead to the right solution. However, the worse overall profile-agreement performance of the conditions with evolved packages available to them compared to the framework-access condition demonstrates important limitations of reflexive evolution.

\subsubsection{What reflexive evolution adds to an informed agent for pXRD}
\label{sec:multiphase-nonpromotion}

Framework + v6 evaluates whether the evolved package can combine its phase-recovery capability with the stronger refinement performance of framework access. Recovery of \(\mathrm{Sr_{2}WO_{5}}\) increases from zero to five of nine completed attempts, but mean \(R_{\mathrm{wp}}\) across the experimental panel rises from 5.98\% to 6.49\% (Tables~\ref{tab:multiphase-conditions} and~\ref{tab:multiphase-recovery}). The package therefore encourages a productive phase-search route while also changing the analysis in ways that worsen average profile agreement. Interpreting this tradeoff requires examining what the package contributes beyond the information already available in the framework.

\begin{figure}[tbp]
  \centering
  \includegraphics[width=\textwidth]{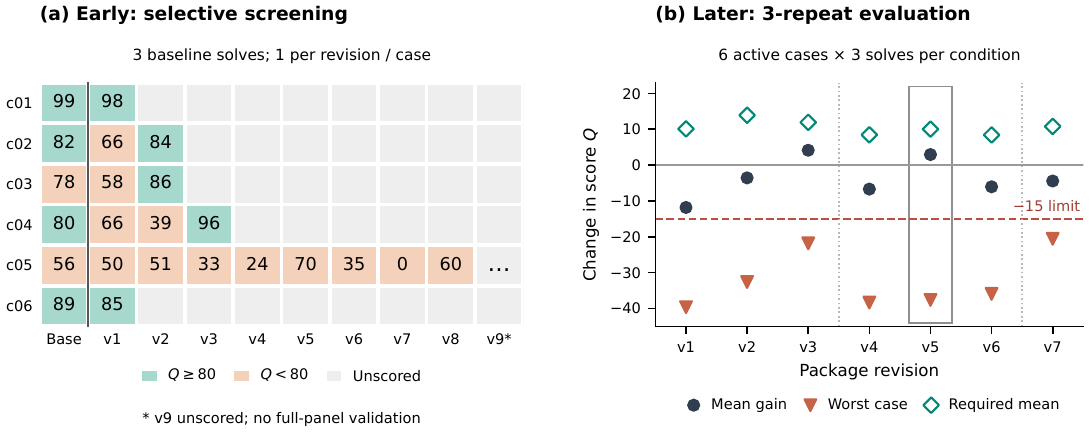}
  \caption{\textbf{Two framework-access evolution campaigns retained their starting agent.} (a) Early selective screening: baseline cells average three solves; revision cells show single solves. Gray cells were not scored. Passing cases were set aside; combining different revisions does not validate a single package. (b) Later 3-repeat evaluation: each revision has three solves on six cases. Changes are relative to the baseline on the same cases. Mean gains must reach the plotted threshold and every case must remain above the dashed regression limit. Vertical separators mark panel changes; the boxed v5 was also tested experimentally. Appendix~\ref{app:multiphase-framework-evolution} gives the protocols and scores; further details can be found in the \texttt{verifiable-pxrd} repo.}
  \label{fig:multiphase-framework-evolution}
\end{figure}

The two framework-evolution campaigns assess whether guidance developed specifically for this better-informed framework-access agent can improve its performance. Neither produced a promoted package (Figure~\ref{fig:multiphase-framework-evolution}). The earlier campaign used the same screening and promotion criterion in the evolution process as evolved v6, revising guidance on unresolved cases while setting aside successful cases. It produced revisions through v8 but never completed a successful validation of one package across the panel; v9 remained unscored when the run was stopped. The later campaign with a slightly modified screening/promotion process in the evolution framework evaluated every revision three times on each active case, completing 156 solves across baseline evaluations and seven revisions. Every revision exceeded the permitted case-level regression. Local improvements were repeatedly accompanied by losses elsewhere.

The unpromoted v5 from the second campaign illustrates this difficulty in generalizing improvements across cases. Its synthetic mean increased by 2.96 score points, but a 38.50-point gain on a missing-\(\mathrm{Mn_{3}WO_{6}}\) problem was offset by a 37.62-point regression on a difficult Zn refinement. Its exploratory experimental evaluation on \texttt{divalent-tungstates-PG15} reproduced the broader tradeoff: mean \(R_{\mathrm{wp}}\) increased to 6.44\% relative to framework access, although one of nine Sr attempts recovered the missing phase. Thus, both adding the original v6 package and evolving a new package from framework access yielded useful phase-recovery behavior without improving overall fitting performance.

The contrasting outcomes of baseline and framework-access evolution suggest that a substantial part of the initial opportunity lies in making existing scientific knowledge and task requirements usable by the solver. The framework contains descriptions of the sample family and expected challenges, implementations of relevant physical effects, and explicit evaluation criteria. Framework-access trajectories show agents inspecting and using this material to improve their analyses. Evolution can distill some of that information into guidance, whereas direct access allows the agent to consult the underlying context as the analysis develops. A related pattern appears in the single-phase experiments: the compact \texttt{adapted-codex} package significantly improved performance by clarifying the submission contract and improving the DICVOL14 interface, without providing guidance on solution pathways. Failures arising from missing task context or tool-interface difficulties can admit compact corrections that apply across many problems.

Improving an agent that already has this context presents a different optimization problem. Decisions about phase search, background flexibility, atomic coordinates, and specimen effects depend on the current fit and on one another. Guidance derived from a particular failure must remain useful when those circumstances change. A restrained background can preserve evidence of a weak phase in one pattern and underfit the baseline in another; further prototype search can resolve a missing phase or divert effort from refining an adequate inventory. The evolved packages impose preferences on these decisions. Evolved v6 consistently favors a lower-dimensional background, while the framework-evolved v5 limits optional search and explicitly discourages inspection of the generator and scorer. These prescriptions provide a plausible explanation for why guidance that improves the original baseline can constrain the more informed agent.

These results suggest that the benefit of reflective adaptation depends on which failures remain amenable to instruction changes, a result reflected in the reflexive evolution literature. \citet{akbar2026frontier} report substantially smaller gains from GEPA for a stronger model with higher initial performance, attributing this pattern to reduced opportunity for prompt-level correction. Analyses of learned agent skills also identify gains from restoring workflow instructions and supplying task-specific knowledge absent from the initial configuration \citep{kozyrev2026skillissue}. Once this information is available, further improvement requires guidance that generalizes across heterogeneous decisions. \citet{gao2026p1} show why this can be difficult: rollout variability can obscure differences between prompts, while changes beneficial to some problems harm others. Our framework-access evolution campaigns exhibit these patterns, managing to improve performance with 3-rollout evaluations during evolution, but struggling to find generalizable instructions when the low-hanging fruit of task-alignment and providing context disappear. Together, these observations support distinguishing adaptation that clarifies how to perform and submit the task from adaptation that improves the scientific procedures themselves.

For agentic pXRD analysis in real laboratories, our work suggests providing informative scientific context and evaluating multiple independent solves before investing in extensive prompt evolution (just selecting the lowest-\(R_{\mathrm{wp}}\) baseline solve from three independent attempts reduces mean \(R_{\mathrm{wp}}\) from 7.90\% to 6.92\% on \texttt{divalent-tungstates-PG15} Table~\ref{tab:multiphase-experimental}). Additional reasoning effort and model choice also influence performance: mean \(R_{\mathrm{wp}}\) decreases from 8.98\% for the earlier Sol-high baseline to 7.90\% for Sol-max, while the Opus comparison reaches 6.32\%. Appendix~\ref{app:multiphase-experimental-results} reports the differing runtimes, harnesses, and repeat counts. Framework access and evolution provide additional means of improving a fixed model when a higher reasoning level or a stronger model is unavailable.

The distinctive contribution of synthetic experience is the ability to investigate failures using information unavailable during experimental analysis. Knowledge of the generating phases allows the evolver to establish that a solution omitted a component, identify why its search failed, and test an alternative procedure. In this study, that process produced an analogue-substitution procedure that transferred from a synthetic calcium intermediate to the measured Sr mixture. An expert could also have suggested that route when presented with the baseline solve traces and high \(R_{\mathrm{wp}}\)'s. The broader opportunity arises when promising routes are not known in advance: self-generated, verifiable problems can provide experience from which to discover and validate them, a motivation shared with Absolute Zero \citep{zhao2025absolutezero}.

\FloatBarrier

% Keep the conclusion together after the experimental discussion.
\clearpage
\section{Conclusion}
\label{sec:conclusion}
LLM agents can perform end-to-end pXRD structure determination by combining scientific reasoning, executable analysis, and feedback from physical models. On the 200-problem pXRD-Bench, adding a coding harness, crystallographic software, and verifier-informed adaptation raises selected exact recovery from 27\% to 52\%, 57\%, and 69\%, respectively. On the 20-problem inference-scaling panel, repeated adapted-agent solves supply an exact candidate for 18 problems, and profile-based selection recovers all 18. These results demonstrate the complementary roles of evaluation-time structural verification in adaptation and inference-time profile verification in search and selection.

Experimental multiphase characterization requires a broader notion of success: identifying phases, constructing plausible structures, estimating their fractions, and explaining instrument and specimen effects. Our laboratory-conditioned synthetic framework supplies verifiable experience for this setting without requiring ground-truth experimental solutions. The evolved v6 package improves mean profile agreement over its baseline and transfers an analogue-substitution procedure to a measured mixture whose \(\mathrm{Sr_{2}WO_{5}}\) structure is absent from the supplied library. Direct access to the framework produces still better average fits, while further evolution exposes a tradeoff between phase recovery and routine refinement. Together, these results distinguish two sources of improvement: scientific context helps an agent use its existing capabilities more effectively, while verifiable synthetic experience can develop specific procedures that context alone does not reliably elicit.

The multiphase experiments extend our central argument beyond problems with known reference solutions. Although experimental measurements rarely provide definitive ground truth, synthetic problems supply the information needed to diagnose failures, test alternative strategies, and learn procedures that can transfer to measured samples. The recovery of an absent-library phase in the Sr mixture demonstrates this possibility. For laboratory deployment, our results support providing scientific context and comparing independent solves, then directing adaptation toward persistent failures. A replaceable laboratory specification makes this approach applicable to recurring challenges within a sample family without requiring a large collection of solved experimental measurements. Its scientific value ultimately rests on improving phase identification and structural interpretation together with agreement to the measured diffraction pattern.

\clearpage
\appendix
\flushbottom
\captionsetup{font=normalsize,skip=7pt}
\setlength{\parskip}{5.5pt}
\setlength{\textfloatsep}{20pt plus 2pt minus 4pt}
\setlength{\floatsep}{12pt plus 2pt minus 2pt}
\setlength{\intextsep}{12pt plus 2pt minus 2pt}
\section{Additional experimental details}
\label{app:additional-material}

\subsection{pXRD-Bench construction}
\label{app:pxrd-bench-details}

\paragraph{pXRD-100: controlled training exposure.}
pXRD-100 controls for documented training exposure. Candidates came from the \href{https://github.com/txie-93/cdvae/tree/f857f598d6f6cca5dc1ea0582d228f12dcc2c2ea/data/mp_20}{MP-20 training split} at frozen CDVAE commit \texttt{f857f598d6f6cca5dc1ea0582d228f12dcc2c2ea} and were required to have a numeric Materials Project identifier, full ordered occupancy, 1--20 primitive-cell sites, energy above hull at most \(0.08~\mathrm{eV\,atom^{-1}}\), and formation energy below \(2~\mathrm{eV\,atom^{-1}}\) \citep{xie2022cdvae}. Each selected structure also matched a released Crystalyze training record, a public deCIFer training record, and the later official DiffractGPT training split, yielding 3,649 structures jointly exposed to these systems and MP-20-trained PXRDGen \citep{riesel2024crystalyze,li2025pxrdgen,johansen2026decifer,choudhary2025diffractgpt}. DiffractGPT exposure is a proxy because its paper-era split is unavailable. RealPXRD-Solver was trained on structures from the Materials Project; however, because its exact training split was unavailable, we could not assess case-level overlap with pXRD-Bench. Ab-PXRD-Solver instead conducts crystallographic search rather than learning a structure generator, so exact membership in a structure-training split is not defined for its core solver, although its learned auxiliary models may still be sensitive to distribution shift.

After retaining the lowest-energy eligible structure per formula, the representative panel was moment-matched to MP-20 over symmetry, cell size, chemical complexity, stability, density, volume per atom, and elemental prevalence. The challenge panel contains six cases from the top decile of each of five prespecified difficulty scores: rare symmetry, site and chemical complexity, peak congestion, scattering ambiguity, and primitive-cell size. The polymorph panel contains 20 same-composition pairs stratified by diffraction-profile similarity into five easy, ten medium, and five hard pairs. Eligibility and selection were fixed without using evaluated-model outputs.

\paragraph{A frozen public-exposure boundary for pXRD-ood.}
Out-of-distribution status is certified relative to a specified public boundary, not asserted universally. Before selection, we hash-locked a 3,279,483-record SQLite index comprising the released Crystalyze, CDVAE/PXRDGen, deCIFer, DiffractGPT, and DARA corpora together with frozen JARVIS-DFT, COD, and Materials Project releases \citep{jain2013materialsproject,grazulis2012cod,choudhary2020jarvis,riesel2024crystalyze,li2025pxrdgen,johansen2026decifer,choudhary2025diffractgpt,fei2026dara}. When checkpoint provenance was incomplete, we conservatively indexed a public superset, including complete NOMA, JARVIS, and COD snapshots. Source versions, counts, and hashes are retained in the benchmark manifests. Inaccessible data, notably DARA's licensed ICSD collection, remain outside this certificate.

\paragraph{Separating compositional and topological extrapolation.}
Every pXRD-ood candidate first had to be absent as a species-aware structure from the exposure index under \texttt{StructureMatcher} with \(l_{\mathrm{tol}}=0.20\), \(\theta_{\mathrm{tol}}=5^\circ\), and \(s_{\mathrm{tol}}=0.30\), primitive-cell reduction, and lattice rescaling. Rescaling prevents a volume-only perturbation from establishing novelty. For the 50 composition-novel cases, the reduced composition is absent from every source while the species-erased geometry matches an exposed parent under stricter tolerances \((0.15,3^\circ,0.20)\). For the 50 topology-novel cases, the composition occurs in all five learned-model sources and at least one retrieval source, but \texttt{StructureMatcher.fit\_anonymous} finds no match among indexed structures with the same anonymized formula. These certificates respectively isolate unfamiliar chemistry on an exposed scaffold and unfamiliar topology at familiar chemistry.

\paragraph{Generation, filtering, and final selection.}
We generated 496 candidates by isovalent substitution (100), ordered decoration (100), PyXtal Wyckoff enumeration (140), and MACE-MH-1-relaxed crystal search (156); 403 passed structural-validity, diffraction-quality, and novelty checks \citep{fredericks2021pyxtal,batatia2022mace,batatia2025crosslearning}. A deterministic mixed-integer program selected 25 cases per stream, balanced each stream across five primitive-cell-size bins, enforced unique compositions and parents, and limited repeated crystal systems, space groups, diffraction profiles, and species-erased geometries. Formulas are disjoint from pXRD-100, and neither selection nor filtering used baseline performance. An independent audit then recomputed all certificates from the released CIFs and frozen index. The certificates establish operational novelty and well-formed targets, not stability, synthesizability, or absence from every inaccessible database; simulator hashes and noise seeds are retained in the manifests.

\paragraph{GSAS-II profile-parameter conventions.}
\label{app:profile-parameters}
For constant-wavelength profiles, GSAS-II stores the Gaussian variance coefficients \(U,V,W\) in squared centidegrees and the Lorentzian width coefficients \(X,Y\) (and constant term \(Z\)) in centidegrees, where one centidegree is \(0.01^\circ\) in \(2\theta\). \(SH/L\) is the dimensionless Finger--Cox--Jephcoat axial-divergence parameter controlling peak asymmetry. Crystallite size is specified in \(\mu\mathrm{m}\), and microstrain in parts per million of relative reciprocal-spacing broadening; thus 1000 ppm corresponds to 0.1\% \citep{toby2013gsasii}.

\paragraph{Recovery evaluation.}
Unlike the novelty check, final recovery scoring disables lattice rescaling because cell dimensions are part of the solution. It otherwise uses \(l_{\mathrm{tol}}=0.20\), \(\theta_{\mathrm{tol}}=5^\circ\), and \(s_{\mathrm{tol}}=0.30\), with primitive-cell reduction, supercell matching, no subset matching, \texttt{ElementComparator}, and evaluation in both comparison directions \citep{ong2013pymatgen}. Exact@1 evaluates the declared primary structure; exact@\(k\) is an oracle match among the first \(k\) candidates. Missing or unparseable CIFs are failures.

\subsection{Specialist pXRD baseline protocols}
\label{app:baseline-protocols}

The specialist comparison holds the reference structures and exact-recovery evaluator fixed, but preserves each framework's native diffraction representation, conditioning variables, and generation procedure. Recalculating the same reference CIF under each released preprocessing pipeline avoids evaluating a trained checkpoint on an artificial common input distribution. Consequently, the rows in Table~\ref{tab:baseline-protocols} are matched evaluation experiments, not equal-information ablations.

\begin{table*}[htbp]
    \centering
    \scriptsize
    \setlength{\tabcolsep}{2.0pt}
    \renewcommand{\arraystretch}{1.15}
    \caption{Reproduction protocol for the specialist pXRD baselines. Oracle inputs are derived from the hidden reference structure and therefore provide more information than the reduced-composition benchmark prompt.}
    \label{tab:baseline-protocols}
    \begin{tabular}{@{}l>{\raggedright\arraybackslash}p{2.55cm}>{\raggedright\arraybackslash}p{2.9cm}c>{\raggedright\arraybackslash}p{4.2cm}@{}}
        \toprule
        Method & Native diffraction input & Structural side information & Candidates & Generation and reporting protocol \\
        \midrule
        Crystalyze
            & Cu K\(\alpha\) Caglioti--Gaussian encoder profile
            & Element set; stoichiometric counts not enforced
            & 8
            & Released diffusion sampler with allowed-element constraints; samples are reported in generation order. \\
        DiffractGPT
            & Released 180-bin nearest-bin vector
            & Reduced composition
            & 1
            & One greedy decode with a 1,024-token generation limit; no stochastic best-of-eight policy is introduced. \\
        Ab-PXRD-Solver
            & Released noisy GSAS-II profile
            & Reduced composition; space group inferred
            & 1
            & Deterministic search with Halton quasi-random sampling and a four-hour CPU limit per case; a candidate emitted before timeout remains eligible. \\
        deCIFer
            & Clean 1,000-bin \(Q\)-space pseudo-Voigt profile
            & Full conventional-cell composition (oracle \(Z\))
            & 8
            & Stochastic autoregressive sampling at temperature 1 and top-\(k=50\); raw samples are reported in generation order. \\
        PXRDGen
            & Frozen GSAS-II Cu K\(\alpha_1/\alpha_2\) profile
            & Exact primitive-cell composition and atom count (oracle \(Z\))
            & 8
            & 200 diffusion steps with step size 5.0; the reported structures are raw generated samples without the paper's subsequent Rietveld stage. \\
        RealPXRD-Solver
            & Variable-length native sparse peak list
            & Exact primitive-cell composition and atom count (oracle \(Z\)); no lattice
            & 20 (first 8 reported)
            & 200 diffusion steps with step size 5.0; candidate order is released sampler order because the public GSAS-II ranking/refinement stage is unavailable. \\
        \bottomrule
    \end{tabular}
\end{table*}

\paragraph{Candidate generation and scoring.}
For Crystalyze, deCIFer, and PXRDGen, exact@1 is the first generated sample and exact@8 is oracle candidate-set recall over samples 1--8. RealPXRD-Solver generates 20 structures per case, but the main comparison uses the same first-eight cutoff; its best-of-20 result is retained only as an auxiliary oracle diagnostic. DiffractGPT follows its released deterministic greedy helper, so repeated calls under the frozen input produce the same output and no exact@8 value is reported. Ab-PXRD-Solver performs an internal search and emits at most one final structure. No method is reranked with the shared \(R_{\mathrm{wp}}\) verifier, and missing, invalid, or unparsable outputs count as failures on the original 200-case denominator.

\paragraph{Output completeness and validity.}
\label{app:baseline-output-completeness}
Table~\ref{tab:baseline-output-validity} distinguishes missing or invalid outputs from structures with the wrong composition. Basic validity requires a parseable CIF, cell volume at least \(0.1~\text{\AA}^3\), and minimum periodic interatomic distance at least \(0.5~\text{\AA}\); these checks do not establish physical plausibility. Composition matching compares reduced elemental proportions. Crystalyze's element-set conditioning does not enforce stoichiometric counts: all 1,600 candidates pass the validity checks, but only 351 match the required composition. Conversion or parsing fails for 41 DiffractGPT outputs and 41 deCIFer samples. PXRDGen and RealPXRD-Solver return valid, composition-matching structures in nearly all evaluated slots, so their exact-recovery shortfall predominantly reflects incorrect structures rather than missing or invalid outputs.

\begin{table}[htbp]
    \centering
    \small
    \setlength{\tabcolsep}{7pt}
    \renewcommand{\arraystretch}{1.12}
    \caption{Output completeness and validity before additional refinement, pooled across all 200 problems. Counts refer to candidate slots, not recovered problems. We use one slot per problem for DiffractGPT and Ab-PXRD-Solver, and the first eight for the other methods, including RealPXRD-Solver. Missing outputs remain in the slot denominator.}
    \label{tab:baseline-output-validity}
    \begin{tabular}{@{}lccc@{}}
        \toprule
        Method & Candidate slots & Valid structures & \shortstack{Valid and\\composition-matched} \\
        \midrule
        Crystalyze & 1,600 & 1,600 & 351 \\
        DiffractGPT & 200 & 146 & 138 \\
        Ab-PXRD-Solver & 200 & 160 & 160 \\
        deCIFer & 1,600 & 1,500 & 1,406 \\
        PXRDGen & 1,600 & 1,592 & 1,592 \\
        RealPXRD-Solver & 1,600 & 1,587 & 1,587 \\
        \bottomrule
    \end{tabular}
\end{table}

Ab-PXRD-Solver produces structures for 160 problems (77 pXRD-100 and 83 pXRD-ood). Of the remaining 40, 23 reach the four-hour limit without a candidate (15 pXRD-100 and eight pXRD-ood), and 17 terminate without a viable cell--space-group solution (eight and nine, respectively). Another 56 runs reach the limit after producing a candidate; these structures remain eligible and are included among the 160. The 40 missing outputs remain failures in the 200-problem recovery denominator; only the 160 available structures can enter the additional refinement experiment below.

\paragraph{Reproduction-specific caveats.}
DiffractGPT states that Gaussian broadening was used but does not specify its width or release the paper-era preprocessing code; we therefore use the later released 180-bin nearest-bin representation, spanning \(0^\circ\)--\(89.5^\circ\) in \(0.5^\circ\) increments. Ab-PXRD-Solver's regex-only composition parser required parenthesized formulas to be flattened to chemically equivalent integer element counts in 17 cases from each corpus; neither composition nor stoichiometry changed. For deCIFer, a reduced-composition diagnostic recovered only 2/200 cases at exact@8, whereas the paper-native full-cell conditioning reported in the main text supplies the hidden conventional-cell \(Z\). PXRDGen and RealPXRD-Solver likewise receive the hidden primitive-cell formula and atom count; 54/100 pXRD-100 targets have a primitive-cell formula that differs from the reduced formula, making these materially oracle-assisted conditions. Exact code revisions, checkpoint hashes, random seeds, per-case inputs, generated CIFs, and verifier outputs are retained in the released experiment bundles.

\FloatBarrier
\subsubsection{Method-specific refinement of specialist predictions}
\label{app:specialist-refinement}

We refine the saved specialist predictions to measure changes in exact structural recovery and profile agreement. Each method uses its original diffraction input and its own refinement procedure. The completed experiments include eight candidates per problem for Crystalyze and PXRDGen, 20 for RealPXRD-Solver, and the 160 available Ab-PXRD-Solver predictions described above. Reference structures are used only for post-hoc scoring.

\paragraph{Refinement procedures.}
All three CPU procedures use GSAS-II revision 5806 \citep{toby2013gsasii}. For PXRDGen, we execute the released two-pass refinement functions, fitting cell, scale, and coordinates, with conventional-cell conversion between passes and symmetry detection at \texttt{symprec=0.5} \citep{li2025pxrdgen}. We adapt the file paths and supply the benchmark observation, excluding the example's reference-structure analysis. Each refinement call allows 100 cycles, with a 30-minute limit per candidate. For Ab-PXRD-Solver, we use its released staged optimizer with 20 cycles per call and a one-minute limit \citep{su2026abpxrd}. Its early-exit rule skips atomic and microstrain refinement if the profile/cell fit remains above \(R_{\mathrm{wp}}=50\%\). We add an export of the fitted phase because the released routine returns the input CIF path. Both methods retain the default GSAS-II uncertainty-based CIF formatting, including export failures caused by rounding.

For RealPXRD-Solver, we implement the sequence described in Supplementary Figure~S4 of \citet{li2026realpxrd}: background, scale, profile, cell, zero shift, coordinates, then joint refinement. This is our adapter; the inspected release did not include the authors' refinement implementation. Saved peak lists are broadened with \(0.10^\circ\)-FWHM Gaussians on a \(5^\circ\)--\(80^\circ\) grid at \(0.01^\circ\) spacing, using \(\lambda=1.54184~\text{\AA}\). We use unit weights, P1, fixed occupancies and \(U_{\mathrm{iso}}=0.01~\text{\AA}^2\), and one fixed highest-atomic-number atom to set the origin. Each stage allows 20 cycles, with 15 minutes per candidate. A stage is rolled back if the recalculated residual worsens or the geometry/profile bounds are violated.\footnote{Bounds are \([0.75,4/3]\) times the initial volume, minimum interatomic distance \(0.5~\text{\AA}\), zero shift \(\pm0.15^\circ\), and Gaussian FWHM \(0.02^\circ\)--\(0.5^\circ\); composition must be preserved. The residual tolerance is \(10^{-5}\) percentage points. The adapter exports at 12 significant digits.}

\paragraph{Crystalyze / StructSnap.}
The GPU experiment refines eight saved candidates per problem using the first stage of their released differentiable routine \citep{riesel2024crystalyze}. It fits cell lengths and peak width but not coordinates, minimizing negative cosine similarity against the unmasked 8,500-point target. This stage permits 30,001 iterations with native early stopping; each candidate has a one-hour CUDA limit. We repair an obsolete pymatgen import, save the best peak-width tensor with the best structure, and run one candidate per batch. Since we report only the first refinement stage and omit the subsequent fallback schedules with alternative initializations, this should be considered an exploratory experiment, not a reproduction of their released StructSnap routine. Including a second coordinate-refinement stage lowered exact matches from 60 to 23 among 1,067 candidates, but we did not evaluate the additional lattice-refinement schedules alluded to in the paper because of their computational cost.

\paragraph{Recovery and failure handling.}
We apply the benchmark's composition, validity, and symmetric \texttt{StructureMatcher} checks to the exported CIFs. A completed refinement replaces its original candidate even if it loses an exact match. On an execution or export failure, the original is retained. Table~\ref{tab:specialist-refinement-recovery} reports this refinement-with-fallback procedure using the original candidate order and all 100 problems per corpus. Pass@8 is oracle candidate-set coverage, not \(R_{\mathrm{wp}}\)-ranked selection.

\begin{table}[htbp]
    \centering
    \small
    \setlength{\tabcolsep}{7pt}
    \renewcommand{\arraystretch}{1.12}
    \caption{Exact recovery before \(\to\) after method-specific refinement, out of 100 problems per corpus. Execution/export failures retain the original candidate. Crystalyze results use the first StructSnap stage.}
    \label{tab:specialist-refinement-recovery}
    \begin{tabular}{@{}llcc@{}}
        \toprule
        Method & Corpus & Pass@1 & Pass@8 \\
        \midrule
        Crystalyze / StructSnap & pXRD-100 & \(11\to12\) & \(19\to20\) \\
        & pXRD-ood & \(2\to2\) & \(6\to6\) \\
        PXRDGen & pXRD-100 & \(47\to32\) & \(84\to67\) \\
        & pXRD-ood & \(17\to6\) & \(26\to17\) \\
        Ab-PXRD-Solver & pXRD-100 & \(42\to41\) & --- \\
        & pXRD-ood & \(20\to23\) & --- \\
        RealPXRD-Solver adapter & pXRD-100 & \(41\to29\) & \(62\to62\) \\
        & pXRD-ood & \(20\to18\) & \(28\to27\) \\
        \bottomrule
    \end{tabular}
\end{table}

First-candidate recovery increases from 13/200 to 14/200 for Crystalyze, decreases from 64/200 to 38/200 for PXRDGen and from 61/200 to 47/200 for the RealPXRD-Solver adapter, and increases from 62/200 to 64/200 for Ab-PXRD-Solver. Among all 20 RealPXRD-Solver candidates, oracle recovery changes from 100/200 to 97/200.

\paragraph{Execution failures and retained matches.}
Table~\ref{tab:specialist-refinement-failures} separates matching refined outputs from matches retained only because refinement failed. PXRDGen's 38 first-candidate successes comprise 27 refined matches and 11 retained originals (nine pXRD-100 and two pXRD-ood). Neither Ab-PXRD-Solver nor the RealPXRD-Solver adapter gains a first-candidate match through fallback.

\begin{table}[htbp]
    \centering
    \small
    \setlength{\tabcolsep}{4pt}
    \renewcommand{\arraystretch}{1.12}
    \caption{Execution and fallback accounting across both corpora. Export/failure counts cover all available candidate slots. The final two columns count exact first candidates out of 200 problems and sum to post-refinement pass@1. Export completion does not establish structural correctness or optimizer convergence.}
    \label{tab:specialist-refinement-failures}
    \begin{tabular}{@{}lcccc@{}}
        \toprule
        & \multicolumn{2}{c}{All candidate slots} & \multicolumn{2}{c}{Exact first candidates} \\
        \cmidrule(lr){2-3}\cmidrule(lr){4-5}
        Method & \shortstack{Exported/\\available} & Failures & Refined & \shortstack{Retained\\original} \\
        \midrule
        Crystalyze / StructSnap & 1,600/1,600 & 0 & 14 & 0 \\
        PXRDGen & 1,225/1,600 & 375 & 27 & 11 \\
        Ab-PXRD-Solver & 159/160 & 1 & 64 & 0 \\
        RealPXRD-Solver adapter & 3,975/4,000 & 25 & 47 & 0 \\
        \bottomrule
    \end{tabular}
\end{table}

PXRDGen has 273 unreadable exports, 87 inter-pass symmetry-conversion failures, 13 CIF-import failures, one numerical failure, and one composition-changing export. Ab-PXRD-Solver has one unreadable export. The RealPXRD-Solver adapter has 24 unreadable inputs and one structure with no reflections in the observation range. Rejected individual RealPXRD-Solver stages restore the last accepted state and are recorded separately from full failures. A readable Ab-PXRD-Solver export also loses one match because rounding changes its parsed site multiplicity; this is scored as a completed but invalid structure rather than replaced by the original.

\paragraph{Paired profile agreement.}
We independently simulate each original/refined CIF pair against the method-specific observation using identical fixed profile and displacement parameters before and after. Only a nonnegative scale and four Chebyshev background coefficients are fitted, with unit weights. For Crystalyze, PXRDGen, and Ab-PXRD-Solver the profile width is initialized from observed peaks; the RealPXRD-Solver adapter uses the fixed \(0.10^\circ\) width. These residuals measure exported-structure fit separately from the optimizer's reported objective. Absolute values should be compared within a method, since observations and forward models differ. Unlike the clean-reference distributions in Appendix~\ref{app:profile-residual-distributions}, this diagnostic compares structures against the observation and fits scale/background.

\begin{table}[htbp]
    \centering
    \small
    \setlength{\tabcolsep}{5pt}
    \renewcommand{\arraystretch}{1.12}
    \caption{Profile agreement before \(\to\) after refinement. Here \(n\) counts successfully exported first candidates with both residuals available; the same \(n\) candidates contribute on each side. Failed refinements and missing originals are excluded from this paired table, but remain in recovery denominators.}
    \label{tab:specialist-refinement-fit}
    \begin{tabular}{@{}llccc@{}}
        \toprule
        Method & Corpus & \(n\) & Mean \(R_{\mathrm{wp}}\) (\%) & Median \(R_{\mathrm{wp}}\) (\%) \\
        \midrule
        Crystalyze / StructSnap & pXRD-100 & 100 & \(93.35\to80.93\) & \(95.66\to86.23\) \\
        & pXRD-ood & 100 & \(94.57\to87.68\) & \(95.40\to92.66\) \\
        PXRDGen & pXRD-100 & 76 & \(89.47\to51.10\) & \(93.79\to51.60\) \\
        & pXRD-ood & 71 & \(93.10\to82.10\) & \(96.17\to92.09\) \\
        Ab-PXRD-Solver & pXRD-100 & 76 & \(41.47\to30.50\) & \(36.54\to17.70\) \\
        & pXRD-ood & 83 & \(54.66\to46.12\) & \(53.08\to30.67\) \\
        RealPXRD-Solver adapter & pXRD-100 & 100 & \(94.71\to85.52\) & \(97.68\to96.34\) \\
        & pXRD-ood & 99 & \(96.98\to94.02\) & \(98.06\to98.04\) \\
        \bottomrule
    \end{tabular}
\end{table}

Improved profile fit does not consistently improve structural recovery. For PXRDGen on pXRD-100, the median residual falls from 93.79\% to 51.60\%, while first-candidate recovery falls from 47 to 32. These outcomes reflect the specified refinement procedures, including their numerical and export behavior. Across all 1,600 Crystalyze candidates, the first StructSnap stage increases median native cosine similarity from 0.235 to 0.427, but first-candidate recovery increases by only one problem and oracle pass@8 changes from 25/200 to 26/200.

\paragraph{Reproducibility.}
Source versions, dependency pins, runner snapshots, original/refined CIFs, per-candidate scores, and failure logs are retained under \path{experiments/appendix/native-refinement} and \path{results/appendix/native-refinement}. For 508 RealPXRD-Solver candidates, the independent profile diagnostic failed after refinement; recalculating it in a fresh process recovered the residuals without changing the CIFs or rerunning refinement. Fourteen successful-diagnostic controls agreed within \(10^{-6}\) percentage points. The earlier precision-patched pilot is retained separately and is not pooled with these results.

\FloatBarrier

\subsection{Agent-adaptation experimental details}
\label{app:agent-adaptation-details}

All four adaptation-ladder conditions use GPT-5.6 Sol at high reasoning effort and the same reduced composition, wavelength, diffraction profile, and output contract. The tool-free condition receives the complete 8,501-row profile inline in one user message. Coding-agent conditions receive it as \texttt{pattern.csv} in an isolated persistent workspace and may return \texttt{selected.cif} plus a small number of explicitly named alternatives. Network and web access are disabled for every coding-agent rollout. The shared coding environment used Python 3.11.15 with NumPy 2.4.6, SciPy 1.17.1, pymatgen 2026.5.4, ASE 3.29.0, spglib 2.7.0, and PyXtal 1.1.4; the pXRD-specific conditions additionally exposed GSAS-II revision 5837 and a hash-locked DICVOL14 executable. All coding conditions received read-only access to the 62,098-entry COD subset defined by DARA's \texttt{cod\_filtered\_info\_2024} index \citep{grazulis2012cod,fei2026dara}. Under the benchmark’s structural-equivalence criterion, this COD subset overlaps with 7 of the 100 pXRD-100 reference structures and none of the 100 pXRD-ood references.

\begin{table*}[htbp]
    \centering
    \scriptsize
    \setlength{\tabcolsep}{2.0pt}
    \renewcommand{\arraystretch}{1.15}
    \caption{Interfaces and scientific environments in the agent-adaptation ladder. The capability column describes persistent task-specific material supplied before a rollout; scripts written by an agent during its own rollout are not persistent across cases.}
    \label{tab:agent-condition-protocols}
    \begin{tabular}{@{}l>{\raggedright\arraybackslash}p{2.5cm}>{\raggedright\arraybackslash}p{3.8cm}>{\raggedright\arraybackslash}p{2.9cm}>{\raggedright\arraybackslash}p{1.6cm}@{}}
        \toprule
        Condition & Interface & Scientific environment & Persistent pXRD capability & Per-case limit \\
        \midrule
        GPT-5.6 Sol (high)
            & Single Responses API call; no workspace or tools
            & None
            & None
            & 65,536 output tokens \\
        \texttt{matsci-codex}
            & Stateful Codex coding agent
            & Scientific Python, including NumPy, SciPy, pymatgen, ASE, spglib, and PyXtal; read-only local COD
            & None
            & 2 vCPU, 5 GiB, 1 hour \\
        \texttt{pxrd-codex}
            & Stateful Codex coding agent
            & \texttt{matsci-codex} environment plus GSAS-II revision 5837 and bare DICVOL14
            & None
            & 2 vCPU, 5 GiB, 1 hour \\
        \texttt{adapted-codex}
            & Stateful Codex coding agent
            & Same pXRD environment as \texttt{pxrd-codex}
            & Frozen 139-word skill and neutral DICVOL14 adapter
            & 2 vCPU, 5 GiB, 1 hour \\
        \bottomrule
    \end{tabular}
\end{table*}

\begin{table*}[htbp]
    \centering
    \scriptsize
    \setlength{\tabcolsep}{2.0pt}
    \renewcommand{\arraystretch}{1.12}
    \caption{Compute and token accounting for the 200-case agent-adaptation evaluations. Wall time includes model-service latency, agent reasoning, and local command execution. For attempts without a final usage record, token counts are estimated from wall-clock duration using the mean tokens per minute of recorded attempts under the same condition; cached input is a subset of input rather than an additional token category.}
    \label{tab:agent-compute-accounting}
    \begin{tabular}{@{}lcccccc@{}}
        \toprule
        Condition
            & Attempts
            & \shortstack{Timeouts /\\ other incomplete}
            & \shortstack{Mean / median\\ wall (min)}
            & \shortstack{Total\\ wall-hours}
            & \shortstack{Usage\\ records}
            & \shortstack{Mean input / cached / output\\ tokens} \\
        \midrule
        GPT-5.6 Sol (high)
            & 200 & 0 / 0 & 6.94 / 6.95 & 23.1 & 200 & 0.140M / 0 / 0.021M \\
        \texttt{matsci-codex}
            & 200 & 10 / 3 infrastructure & 25.28 / 21.32 & 84.3 & 187 & 6.488M / 6.294M / 0.041M \\
        \texttt{pxrd-codex}
            & 200 & 1 / 7 nonzero exits & 16.30 / 12.09 & 54.3 & 192 & 7.112M / 6.976M / 0.039M \\
        \texttt{adapted-codex}
            & 200 & 3 / 0 & 21.21 / 16.16 & 70.7 & 197 & 4.203M / 4.045M / 0.036M \\
        \bottomrule
    \end{tabular}
\end{table*}

The coding agents completed a mean of 43.8, 48.9, and 43.2 local command executions per attempt for \texttt{matsci-codex}, \texttt{pxrd-codex}, and \texttt{adapted-codex}, respectively. Their archived event streams preserve command boundaries but not timestamps for individual model and command events, so the total wall time cannot be decomposed reliably into LLM-service and local CPU components. The three \texttt{matsci-codex} infrastructure failures were memory-limit exits and remain denominator misses. The seven non-timeout \texttt{pxrd-codex} nonzero exits terminated before a final usage event; any CIF already written remained eligible for evaluation. The \texttt{matsci-codex} and \texttt{adapted-codex} evaluations used machine-local ChatGPT authentication, whereas the tool-free and \texttt{pxrd-codex} evaluations used API authentication. The corresponding Codex standalone versions were 0.145.0, 0.147.0, and 0.149.0 for \texttt{matsci-codex}, \texttt{pxrd-codex}, and \texttt{adapted-codex}; the ladder should therefore be interpreted as a systems comparison rather than a strict single-variable ablation.

\subsubsection{\texttt{adapted-codex} experience-driven evolution}
\label{app:adapted-codex-self-evolution}

\texttt{adapted-codex} was constructed through a bounded, experience-driven self-evolution procedure. The trainer operated over a fixed curriculum of 50 development problems comprising 25 pXRD-100-style and 25 pXRD-ood-style cases, with structural categories and primitive-cell sizes chosen to approximate the benchmark distribution. The curriculum is disjoint from pXRD-Bench by reduced composition, structure, source material, and OOD generation parent. Its composition and ordering were fixed from pre-run metadata, without access to solver outcomes or verifier scores.

The trainer first launched a blind \texttt{pxrd-codex} rollout on each of the 50 curriculum problems. Trainees received only the reduced composition and diffraction profile; after each trajectory terminated, the trainer gained access to the submitted structures, reference structure, and structure-matcher outcome. The initial system recovered 30/50 reference structures, yielding an experience corpus containing both successful and unsuccessful trajectories together with evaluation-time verifier feedback.

The trainer then synthesized a persistent capability update from recurring trajectory-level patterns. Adaptation was restricted to the agent's capability layer: the underlying model, scientific environment, task interface, and compute allocation remained unchanged. The resulting package consisted of a compact 139-word skill and a neutral adapter exposing DICVOL14 through a reliable programmatic interface. Together, these components encode transferable task and verifier priors while improving access to existing crystallographic computation, without introducing reference structures, case-specific solution policies, or an additional scientific ranking procedure.

The revised capability was behaviorally validated in a single blind wave over all 20 initial misses and three predeclared successful cases chosen as regression controls. It recovered 6/20 previous misses while preserving exact recovery on all three controls; two of the 23 attempts reached the one-hour limit. The capability package was then frozen and evaluated once on the 200 held-out pXRD-Bench problems. This protocol separates experience collection and capability synthesis from the final benchmark evaluation, allowing the resulting performance gain to measure transfer of the learned agent behavior rather than direct optimization against pXRD-Bench.

Our trainer protocol prescribed a minimal capability package surface, with no prescribed workflow or case-specific corrections to preserve agent autonomy. Thus, the trainer agent accepted the 6-case capability gain and did not attempt to further optimize performance against the difficult remaining cases. In our experience, we found that allowing self-evolution to continue unbounded led to sprawling packages with stalled performance gains as the tools and skills over-corrected for observed mistakes and struggled to generalize. pXRD structure determination is a complex task requiring adaptable strategies and crystallographic reasoning. Discovering policies that improve general crystallographic capabilities and lead to improvements beyond verifier alignment may require substantially more rollouts than we allotted and potentially updates to the weights rather then only the harness. 

\paragraph{Frozen pXRD skill.}
The complete 139-word \texttt{pxrd-solve} instruction body bundled with the frozen \texttt{adapted-codex} capability is reproduced below verbatim apart from Markdown-to-LaTeX formatting.

\begin{quote}
\footnotesize
\textbf{pXRD submission contract}

Only \texttt{selected.cif} and \texttt{alternative-<n>.cif} are graded. Keep a parseable \texttt{selected.cif} current once a concrete model exists; it must be the best-supported model.

A candidate is an element-specific periodic realization with explicit cell scale, coordinates, occupancies, and ordering. A prototype, indexed cell, unmaterialized site permutation, or averaged/disordered site is a different candidate. If a few distinct element-to-site orderings, supercells, or periodic models genuinely survive, write each as a separate declared CIF instead of collapsing them into mixed occupancy or leaving them in scripts. Alternatives provide coverage; they do not replace the primary.

Round-trip every declared CIF and confirm its intended reduced formula, occupancies, cell, and coordinates.

\texttt{dicvol14-adapter} is a neutral interface to the installed DICVOL14 binary. Its \texttt{-{}-help} describes explicit peak-list, input-kind, crystal-system, bound, provenance, and timeout options; it performs no peak picking or solution ranking.
\end{quote}

\FloatBarrier

\subsection{Inference-time scaling protocols}
\label{app:inference-scaling-details}

\paragraph{Panel construction.}
All inference-time scaling experiments use the same frozen pXRD-Bench-20 panel. The panel was selected without consulting solver outcomes, fit values, or trajectory contents: three cases lacking a completed neutral pilot for infrastructure reasons were first excluded, and a seeded mixed-integer selection then chose 10 pXRD-100 cases (three representative, three challenge, and four polymorph) and 10 pXRD-ood cases (two substitution, two ordered-decoration, three Wyckoff-enumeration, and three crystal-search). Constraints balanced primitive-cell size and crystal system, retained two complete polymorph pairs and three challenge axes, and prevented repeated OOD parents. The panel is disjoint from the 50-case adaptation curriculum by reduced composition, structure, source material, and OOD generation parent. Selection seed \(20260817\) controlled only deterministic tie-breaking; the model API exposed no deterministic sampling seed. The manifest and selection record were hash-locked before outcomes were opened. Because this panel was subsequently inspected while developing the inference-time procedures, it should be regarded as a development panel rather than an untouched estimate of final-test performance.

\paragraph{Independent repeated sampling.}
Each condition in Figure~\ref{fig:monkey-scaling} comprises eight independent invocations for each of the 20 problems. Every invocation began with a fresh model session and isolated workspace, received the same task prompt and scientific environment, and shared no state or candidates with another invocation. Workers were limited to 2 vCPUs, 5~GiB of memory, and one hour; timeouts, process failures, and missing or unparsable submissions remained failures. The \texttt{pxrd-codex} Sol and Luna campaigns used Codex 0.147.0, while the later \texttt{adapted-codex} Sol campaign used Codex 0.149.0. Replicate numbers identify independent API calls rather than reproducible pseudorandom streams.

For a problem with \(c_i\) candidate-set successes among \(n=8\) observed rollouts, we report the order-invariant estimator
\begin{equation}
\widehat{\operatorname{pass@}k}
=
\frac{1}{20}\sum_{i=1}^{20}
\left[
1-\frac{\binom{n-c_i}{k}}{\binom{n}{k}}
\right].
\label{eq:inference-scaling-pass-at-k}
\end{equation}
Here a candidate-set success means that either \texttt{selected.cif} or a declared \texttt{alternative-<n>.cif} passes the benchmark matcher. Thus, the curves measure expected oracle coverage of a size-\(k\) subset of the eight observed rollouts; they do not describe a deployable selector or constitute confidence intervals over new problems.

\begin{table*}[htbp]
    \centering
    \scriptsize
    \setlength{\tabcolsep}{4.0pt}
    \renewcommand{\arraystretch}{1.12}
    \caption{Compute and token accounting for the repeated-sampling campaigns. Each row contains 160 attempted rollouts. Token totals sum attempts with a final usage event and are therefore lower bounds; cached input is included in, rather than additional to, total input.}
    \label{tab:inference-scaling-compute}
    \begin{tabular}{@{}lrrrrrr@{}}
        \toprule
        Condition
            & \shortstack{Usage\\records}
            & \shortstack{Input\\tokens}
            & \shortstack{Cached-input\\subset}
            & \shortstack{Output\\tokens}
            & \shortstack{Wall\\hours}
            & Timeouts \\
        \midrule
        \texttt{pxrd-codex}, Sol (high)
            & 156/160 & 1{,}110.30M & 1{,}089.35M & 6.28M & 49.4 & 2 \\
        \texttt{pxrd-codex}, Luna (max)
            & 112/160 & 1{,}827.73M & 1{,}792.60M & 12.02M & 79.5 & 29 \\
        \texttt{adapted-codex}, Sol (high)
            & 156/160 & 621.02M & 605.17M & 5.56M & 44.9 & 3 \\
        \bottomrule
    \end{tabular}
\end{table*}

\paragraph{Reference-blind profile selection.}
The \(R_{\mathrm{wp}}\) experiment pooled every \texttt{selected.cif} and declared alternative from the eight \texttt{adapted-codex} Sol rollouts. Of 160 attempts, 157 emitted at least one CIF, giving 213 candidates; all 213 completed the same bounded GSAS-II fit. Candidate structures were expanded to an equivalent P1 representation for backend robustness, while their lattices, coordinates, species, and occupancies remained fixed. We fitted only phase scale, four Chebyshev background coefficients, a zero shift bounded to \(\pm0.15^\circ\), and Gaussian Caglioti \(U\) and \(W\), with \(V=0\); all remaining profile terms were shared and fixed. Because the supplied profiles contain no counting uncertainties, all profile weights were set to one. The candidate with the lowest post-fit residual was selected using only the observed pattern, wavelength, and submitted CIFs. Reference structures and matcher labels were opened only after these selections were sealed. This is a controlled nuisance-profile comparison, not a structural Rietveld refinement or a general proof of structural identity.

\paragraph{Blind stop/continue verifier.}
The LLM verifier receives the composition, wavelength, measured pattern, chronological solver trace, final response, and submitted CIFs. The reference CIF and \texttt{StructureMatcher} result are excluded from its workspace, and the reviewer can only write a JSON \texttt{stop}/\texttt{continue} decision and reason. We evolved its policy with GEPA using 50 nonbenchmark \texttt{adapted-codex} trajectories, split by problem into 30 training, 10 validation, and 10 sealed development cases; each 10-case holdout contained seven exact candidate sets and three misses. GPT-5.6 Sol (high) served as reviewer and reflector. Under a 180-call budget and optimizer seed 0, GEPA generated 10 policies with utility 1 for a correct decision, 0.5 for an unnecessary continuation, and 0 for a false stop or invalid output. Candidate 3 was selected using two validation repeats before the development set was opened, then evaluated with three development repeats. Oracle evidence became available only to the post-decision reflection step during optimization. The selected policy was frozen before its evaluation on the 200 pXRD-Bench first-pass trajectories shown in Figure~\ref{fig:verifier-gated-scaling}(a).

\paragraph{Verifier-gated continuation.}
Figure~\ref{fig:verifier-gated-scaling}(b) starts from replicate 1 of the \texttt{adapted-codex} Sol campaign. The judge first reviews each blind solve. A \texttt{continue} decision launches an evidence-conditioned solve in a fresh workspace supplied with the original problem, first-pass trace and CIFs, and the judge's reason; a \texttt{stop} decision preserves the existing submission without another solver call. Surviving cases are reviewed again, then receive at most two ordinary fresh solves that see only the original problem. Candidates accumulate across completed stages, but the hidden reference is used only for retrospective scoring. Judges and solvers were capped at 30 and 60 minutes, respectively, with at most 16 concurrent calls. The USD axis reconstructs per-call token charges and calibrates them to the contemporaneous \$446.15 API bill for the source eight-replicate campaign: the independent baseline assigns \(k/8\) of that bill to \(k\) complete passes, while the gated curve sums the initial solves, all reviews, and only the solver calls authorized at each stage. Later model-price changes are not applied retroactively.

\begin{table*}[htbp]
    \centering
    \scriptsize
    \setlength{\tabcolsep}{2.15pt}
    \renewcommand{\arraystretch}{1.12}
    \caption{Case-level execution of verifier-gated inference-time scaling on pXRD-Bench-20. Columns follow the fixed panel order. Solve rows report \(E\) when \texttt{selected.cif} is exact, \(A\) when only a declared alternative is exact, and \(M\) when no submitted CIF is exact; \(\text{--}\) denotes a case already terminated by the gate. Review rows report \(S\) (stop) or \(C\) (continue), with \(\checkmark\) or \(\times\) indicating retrospective agreement or disagreement, respectively, with candidate-set exactness at that stage. \(S_0\) is the initial blind solve, \(C_1\) the evidence-conditioned continuation, \(S_2\) and \(S_3\) fresh solves, and \(J_t\) the intervening reviews.}
    \label{tab:verifier-gated-case-matrix}
    \begin{tabular}{@{}l*{20}{c}@{}}
        \toprule
        \textbf{Stage} & \textbf{01} & \textbf{02} & \textbf{03} & \textbf{04} & \textbf{05}
            & \textbf{06} & \textbf{07} & \textbf{08} & \textbf{09} & \textbf{10}
            & \textbf{11} & \textbf{12} & \textbf{13} & \textbf{14} & \textbf{15}
            & \textbf{16} & \textbf{17} & \textbf{18} & \textbf{19} & \textbf{20} \\
        \midrule
        \(S_0\) & E & E & M & E & M & M & E & E & E & E & M & E & M & M & E & E & E & E & M & E \\
        \(J_0\) & \(C^{\times}\) & \(C^{\times}\) & \(C^{\checkmark}\) & \(S^{\checkmark}\) & \(C^{\checkmark}\) & \(C^{\checkmark}\) & \(S^{\checkmark}\) & \(S^{\checkmark}\) & \(S^{\checkmark}\) & \(S^{\checkmark}\) & \(C^{\checkmark}\) & \(S^{\checkmark}\) & \(C^{\checkmark}\) & \(C^{\checkmark}\) & \(C^{\times}\) & \(S^{\checkmark}\) & \(S^{\checkmark}\) & \(C^{\times}\) & \(C^{\checkmark}\) & \(C^{\times}\) \\
        \addlinespace[1pt]
        \(C_1\) & E & E & E & -- & A & M & -- & -- & -- & -- & M & -- & E & M & A & -- & -- & M & E & E \\
        \(J_1\) & \(C^{\times}\) & \(C^{\times}\) & \(C^{\times}\) & -- & \(S^{\checkmark}\) & \(C^{\checkmark}\) & -- & -- & -- & -- & \(C^{\checkmark}\) & -- & \(S^{\checkmark}\) & \(C^{\checkmark}\) & \(S^{\checkmark}\) & -- & -- & \(C^{\checkmark}\) & \(C^{\times}\) & \(C^{\times}\) \\
        \addlinespace[1pt]
        \(S_2\) & E & E & E & -- & -- & M & -- & -- & -- & -- & E & -- & -- & M & -- & -- & -- & M & M & E \\
        \(J_2\) & \(C^{\times}\) & \(S^{\checkmark}\) & \(C^{\times}\) & -- & -- & \(C^{\checkmark}\) & -- & -- & -- & -- & \(C^{\times}\) & -- & -- & \(S^{\times}\) & -- & -- & -- & \(C^{\checkmark}\) & \(C^{\checkmark}\) & \(S^{\checkmark}\) \\
        \addlinespace[1pt]
        \(S_3\) & A & -- & M & -- & -- & M & -- & -- & -- & -- & M & -- & -- & -- & -- & -- & -- & M & M & -- \\
        \bottomrule
    \end{tabular}
\end{table*}

The initial solve covered 13/20 problems. The evidence-conditioned continuation recovered four additional cases, and the first verifier-authorized fresh solve recovered one more; the final fresh-solve stage added no further match. All 20 execution chains completed, yielding cumulative candidate-set coverage of 18/20.

\subsubsection{Stop/continue policy prompts}
\label{app:verifier-prompts}

The 102-word seed and 415-word selected policies compared in Figure~\ref{fig:verifier-gated-scaling}(a) are reproduced below verbatim apart from Markdown-to-LaTeX formatting. A fixed 51-word wrapper supplied the evidence paths and \texttt{decision.json} schema. The wrapper, reflection and evaluator instructions, continuation prompt, and hashes of all frozen prompt artifacts accompany the released experiment.

\paragraph{Seed policy.}
\begin{quote}
\footnotesize\itshape
Act as a inference-time verifier for exact crystal-structure recovery. Decide whether the submitted candidate set is sufficiently established or whether another solution attempt is warranted.

Consider the diffraction fit, physical plausibility, and what the trajectory actually instantiated and compared. Distinguish support for the cell, Z, and average framework from support for the submitted symmetry, coordinates, occupancies, and element ordering. A small unresolved discrete ambiguity should not be considered resolved merely because an averaged or convenient realization fits well.

Decide continue when the diffraction fit is dubious or a materially different, scientifically motivated structure that could plausibly improve the candidate set seems likely.
\end{quote}

\paragraph{Selected policy (\texttt{judge-v1}).}
\begin{quote}
\footnotesize\itshape
Act as an inference-time verifier for exact crystal-structure recovery. Use only the blind evidence. Decide whether the submitted CIF set already contains a sufficiently established realization or whether another solution attempt is likely to add a materially different viable candidate.

Audit claims against the actual trace and CIFs:
\begin{itemize}
    \item Confirm that the primary CIF parses and symmetry-expands to the stated composition, atom count, occupancies, cell, and ordering. Reduce or primitive-standardize cells before treating different settings, supercells, or reported Z values as different structures.
    \item Require strong indexing coverage and no important unexplained reflections. Then require atom-coordinate evidence: a whole-profile or integrated-intensity comparison of the instantiated CIF, not peak positions alone. Consider fit relative to noise, a Pawley/profile lower bound, or competing models when available. Absence of a separately supplied comparable score is not itself a reason to continue if the trace contains auditable equivalent evidence.
    \item Determine what was actually solved. Favor convergence from independent maps/searches, chemically assigned scattering peaks, coordinate refinement, extinction or symmetry tests, and constrained-versus-unconstrained comparisons. A familiar prototype name alone is weak evidence.
    \item Check periodic distances, composition/charge balance, bond-valence consistency, and coordination topology. Use chemistry as a veto for demonstrable failures such as overlaps, pervasive impossible bonds, wrong stoichiometry, or incompatible occupancies---not as a preference for conventional density, packing, coordination number, or literature form. Inspect complete neighbor shells; a cutoff can mislabel long or distorted coordination as undercoordination.
    \item Track diffraction-relevant discrete ambiguities such as element ordering, symmetry lowering, superstructures, occupancies, and distinct frameworks. An ambiguity is adequately handled when alternatives are explicitly submitted or when data, multiplicities, maps, refinement, or physical impossibility quantitatively reject it.
\end{itemize}

Do not continue merely because an untested polymorph, integer-multiple Z, denser packing, or more conventional coordination can be imagined. Such an alternative must be concrete: compatible with indexed reflections and systematic absences, composition and site multiplicities, and supported by a plausible periodic realization or a genuinely comparable fit. Conversely, a low residual cannot rescue a flexible profile fit whose atomic model was never instantiated or whose decisive ordering/symmetry choices remain unsupported.

Decide \textbf{continue} for an invalid CIF, dubious line coverage or atomic-intensity fit, a concrete surviving incompatible realization absent from the set, or a material exact-structure choice supported only by assumption. Decide \textbf{stop} when at least one explicit ordered realization has convergent diffraction support and passes validity/chemistry checks, while credible remaining ambiguities are covered or rejected. Base the reason on the decisive evidence and set \texttt{proposed\_branch} only when a specific viable branch is identifiable.
\end{quote}

\FloatBarrier

\subsection{Profile-residual diagnostics}
\label{app:profile-residual-distributions}

We analyze candidate-level profile agreement across both benchmark corpora, labeling each parseable, forward-simulable structure as an exact match or miss using the benchmark \texttt{StructureMatcher}; omitted candidates remain failures in the primary recovery metric. For a shared cross-method diagnostic, the hidden reference \(s^\star\) and candidate \(s\) are independently forward-simulated and maximum-normalized under the clean GSAS-II \texttt{fixedparams-v1} model \citep{toby2013gsasii}. We then compute the unit-weight, fixed-parameter profile residual
\begin{equation}
R_{\mathrm{wp}}(s,s^\star)
=
100
\sqrt{
\frac{
\sum_i
\left[
\widetilde{y}_i(s^\star)-\widetilde{y}_i(s)
\right]^2
}{
\sum_i \widetilde{y}_i(s^\star)^2
}
}.
\label{eq:fixed-profile-residual}
\end{equation}
which is the standard \(R_{\mathrm{wp}}\) expression with all profile weights set to one. Unit weights are used because these simulated profiles provide no counting uncertainties. No noise is added, and no nuisance, lattice, or structural parameters are fitted. Consequently, this shared fixed-parameter \(R_{\mathrm{wp}}\) is an evaluation-time oracle diagnostic rather than a conventional refined Rietveld result; unlike the bounded-fit \(R_{\mathrm{wp}}\) used for candidate selection, it requires the hidden reference and is unavailable during blind inference. Throughout this section, ``fixed-parameter,'' ``bounded-fit,'' and ``native-profile'' identify the scoring protocol rather than distinct residual definitions.

\begin{figure*}[p]
    \centering
    \includegraphics[width=\textwidth]{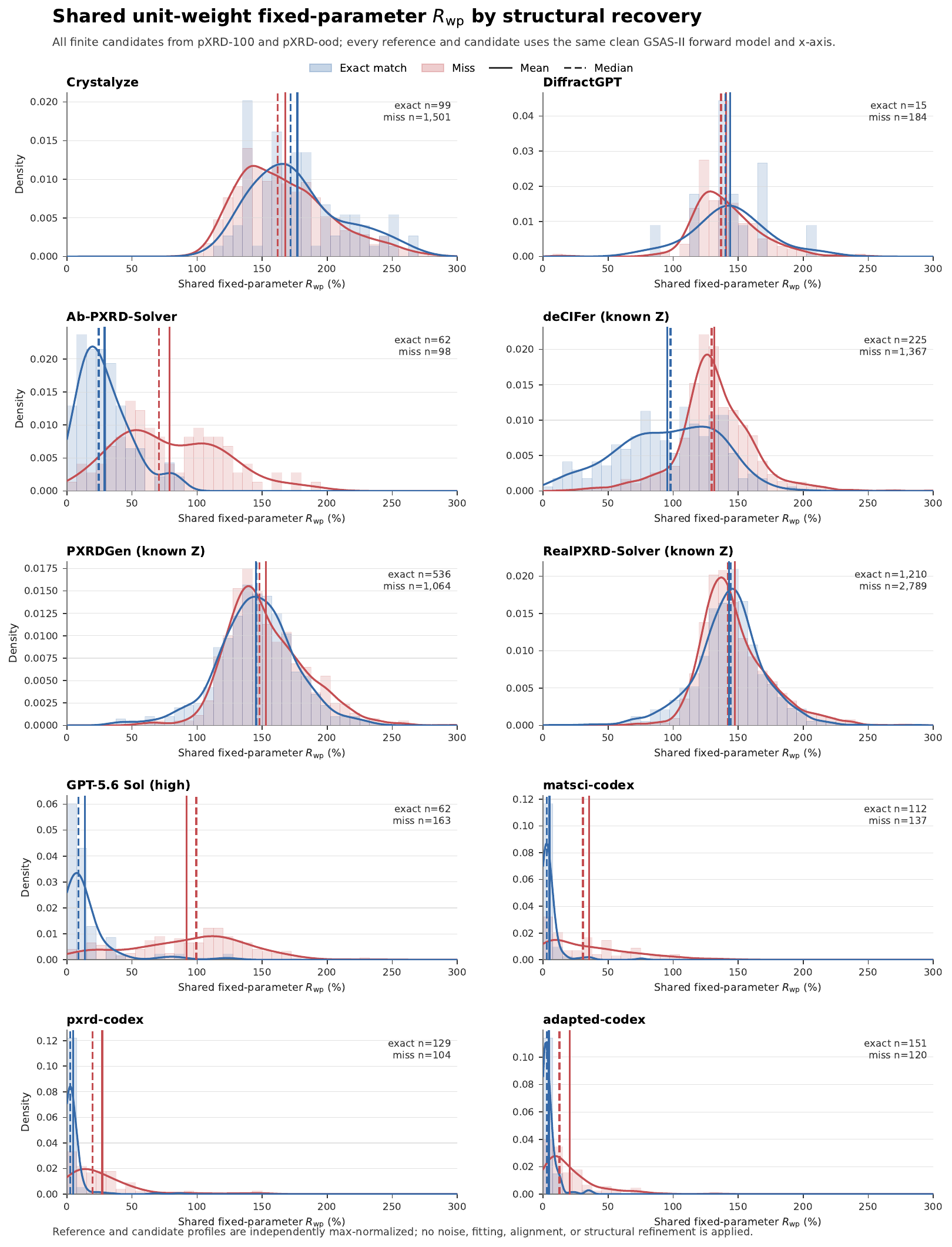}
    \caption{Candidate-level fixed-parameter \(R_{\mathrm{wp}}\) distributions under the shared \texttt{fixedparams-v1} GSAS-II forward model. Blue and red distributions denote candidates accepted and rejected, respectively, by the benchmark structure matcher. Solid lines indicate means and dashed lines indicate medians. Both reference and candidate profiles are clean, independently maximum-normalized simulations; no fitting, alignment, or refinement is performed. The horizontal scale and forward model are shared across all panels.}
    \label{fig:fixedparams-profile-residuals}
\end{figure*}

Under the shared fixed-parameter protocol, \(R_{\mathrm{wp}}\) strongly separates matches from misses for Ab-PXRD-Solver (median \(24.5\%\) versus \(70.8\%\)) and the agent conditions (match medians \(2.4\%\)--\(8.8\%\), versus \(12.6\%\)--\(99.4\%\) for misses), but less consistently for the other generative systems. Because \texttt{StructureMatcher} tolerates finite geometric deviations while this diagnostic uses narrow, unaligned peaks, an accepted structure can still have a large residual; profiles with comparable squared intensity but little peak overlap approach \(100\sqrt{2}\approx141\%\). This diagnostic therefore probes parameter precision but does not replace exact structural recovery. We also compute framework-native \(R_{\mathrm{wp}}\) using each method's frozen input representation listed in Table~\ref{tab:baseline-protocols} \citep{riesel2024crystalyze,choudhary2025diffractgpt,su2026abpxrd,johansen2026decifer,li2025pxrdgen}. RealPXRD-Solver is excluded only from this native-profile analysis because its released variable-length sparse-peak representation has no fixed-grid \(R_{\mathrm{wp}}\) without an additional binning or broadening convention; its candidate CIFs remain included in the shared analysis above.

\begin{figure*}[p]
    \centering
    \includegraphics[width=\textwidth]{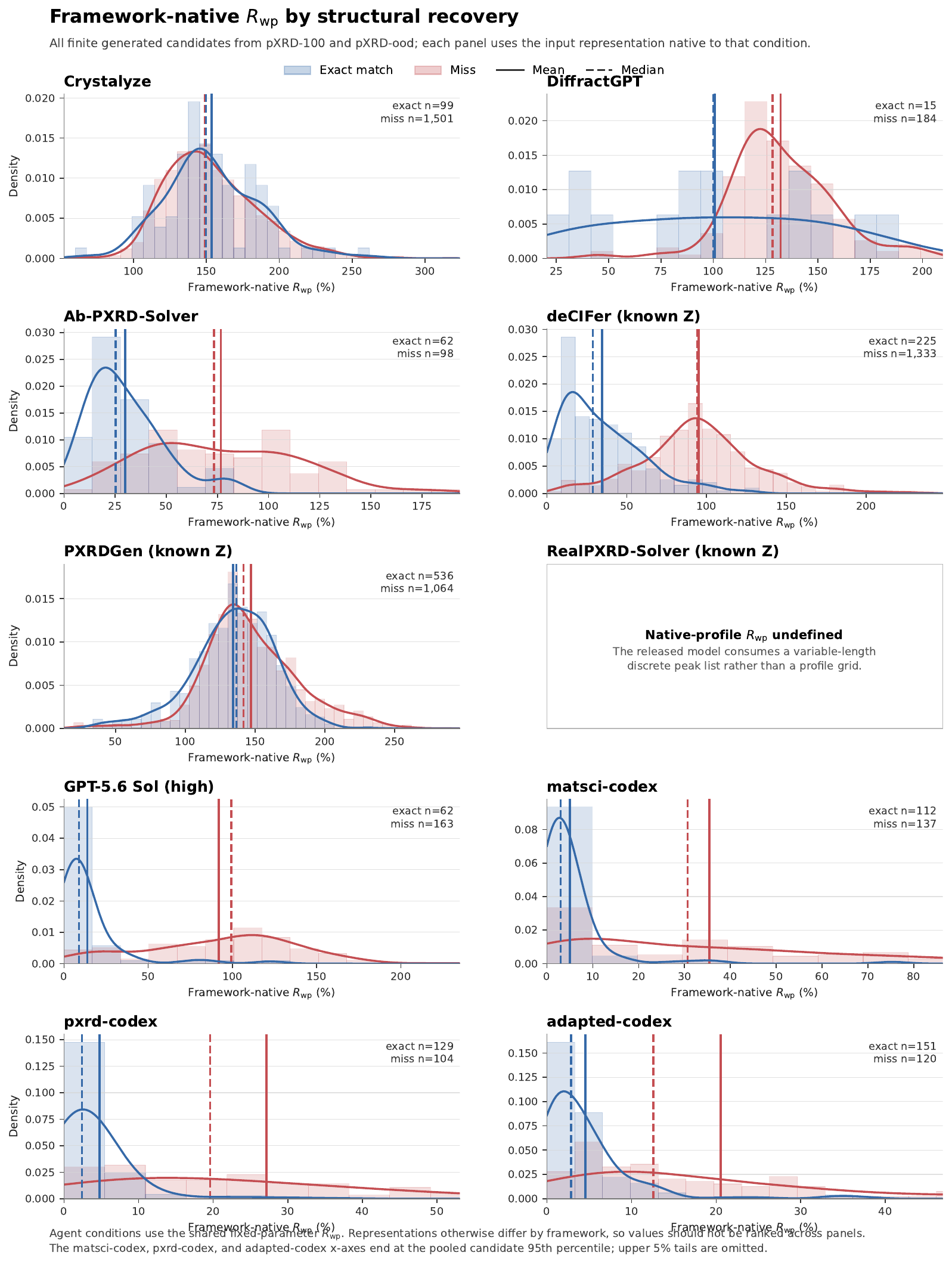}
    \caption{Candidate-level framework-native \(R_{\mathrm{wp}}\) distributions. Colors and vertical-line styles follow Figure~\ref{fig:fixedparams-profile-residuals}. Because the angular or reciprocal-space grids, peak shapes, broadening models, noise, and discretization vary between panels, absolute values should be compared only within a panel and not used to rank frameworks. For \texttt{matsci-codex}, \texttt{pxrd-codex}, and \texttt{adapted-codex}, the displayed range ends at the pooled candidate 95th percentile; the upper 5\% tails remain included in the reported means and medians. The RealPXRD-Solver panel records that a fixed-grid native residual is undefined for its released sparse-peak input representation.}
    \label{fig:native-profile-residuals}
\end{figure*}

Native-profile \(R_{\mathrm{wp}}\) likewise separates matches from misses for Ab-PXRD-Solver (median \(25.3\%\) versus \(73.3\%\)), deCIFer (\(28.7\%\) versus \(94.0\%\)), and the agent conditions, but overlaps substantially for Crystalyze and PXRDGen. It can therefore support candidate ranking within some frameworks, but is neither calibrated across frameworks nor guaranteed to track structural correctness; its usefulness depends on the profile representation and structural-verifier tolerances.

\FloatBarrier

\subsection{Baseline condition, synthetic curriculum, and evolution process for experimental multiphase pXRD}
\label{app:multiphase-evolution}

This section specifies the baseline solver, synthetic generator, scoring objective, and evolution protocol used in Sections~\ref{sec:multiphase-framework} and~\ref{sec:multiphase-november}.

\paragraph{Baseline condition and task specification.}
The baseline uses GPT-5.6 Sol at maximum reasoning effort in Codex CLI 0.154.0, with a terminal, a scientific Python environment, GSAS-II, DICVOL14, and the frozen filtered COD 2024 reference library. Each attempt starts in a fresh workspace. Measured-pattern attempts receive two CPU cores and 5 GiB of memory, with no wall-time limit. Synthetic training attempts have a four-hour limit, with up to ten solvers running concurrently. The baseline receives neither learned guidance nor access to the evolution framework's source code.

The solver inputs are \path{inputs/pattern.csv}, containing \(2\theta\) and observed counts, and \path{inputs/metadata.json}, containing the precursor formulas, firing temperature, and radiation and instrument information. For measured patterns, the counts are supplied without smoothing, background subtraction, normalization, or resampling. Product composition, phase count, and a calibrated instrument profile are unspecified. Synthetic ground truth and deposited experimental interpretations remain unavailable to the solver. The same baseline task prompt is used for synthetic and experimental solves:

\begin{quote}
\small
Determine the crystalline phases and their weight fractions from \path{inputs/pattern.csv}. Read \path{inputs/metadata.json} for the chemical context and measurement information. The product composition and number of phases are unknown.

Use the available scientific tools to determine and refine a structural model for each phase. Save one CIF per phase, the combined observed/calculated/difference profile as \path{fit.csv}, and \path{result.json} listing each phase, its CIF path, its weight fraction normalized over the modeled crystalline phases, and the fit statistics. Explain the evidence and remaining uncertainty briefly in \path{report.md}; state when a structure or fraction cannot be determined reliably.

Save the final GSAS-II project used to produce your submitted CIFs, phase fractions and \path{fit.csv} as \path{final.gpx}.
\end{quote}

The evolved v6 condition appends its learned guidance to this prompt and provides read-only access to its prototype-search helper.

\paragraph{Synthetic problem construction.}
The evolver selects source structures, positive crystalline mass fractions summing to one, precursor metadata, count level, and a random seed for each case. It records a scientific rationale and the intended reference availability: the full library, all entries of one reduced composition withheld, or specified COD records withheld. The initial six cases are listed in Table~\ref{tab:multiphase-curriculum}. The next batch adds an Mg partial reaction and a four-phase Ba mixture; the third adds a Ni source-quality problem and a Cu mixture without the nominal \(\mathrm{CuWO_{4}}\) product. Unresolved earlier cases remain in the active six-case panel.

\begin{table}[htbp]
  \centering\small
  \caption{\textbf{Initial synthetic curriculum.} Generating phase identities and fractions are visible to the evolver and evaluator, but not supplied to the solver. Withholding removes the specified references from the solver's library; chemically related structures may remain available.}
  \label{tab:multiphase-curriculum}
  \begin{tabularx}{\textwidth}{@{}l p{0.40\textwidth} X@{}}
    \toprule
    Case & Generating phase inventory & Intended difficulty \\
    \midrule
    c01 & \(\mathrm{CaWO_{4}}\) & Single-phase control with an available parent reference. \\
    c02 & \(\mathrm{ZnWO_{4}}\), \(\mathrm{ZnO}\), \(\mathrm{WO_{3}}\) & Low counts; a 7.3 wt\% residual precursor. \\
    c03 & \(\mathrm{SrWO_{4}}\), \(\mathrm{SrCO_{3}}\), \(\mathrm{WO_{3}}\) & All \(\mathrm{SrWO_{4}}\) entries withheld; carbonate-derived mixture. \\
    c04 & \(\mathrm{MnWO_{4}}\), \(\mathrm{Mn_{3}O_{4}}\), \(\mathrm{WO_{3}}\) & All \(\mathrm{MnWO_{4}}\) entries withheld; weak residual oxide. \\
    c05 & \(\mathrm{CaWO_{4}}\), \(\mathrm{Ca_{3}WO_{6}}\), \(\mathrm{WO_{3}}\) & Withheld 10.7 wt\% intermediate and overlapping reflections. \\
    c06 & \(\mathrm{PbWO_{4}}\), \(\mathrm{Pb_{3}O_{4}}\), \(\mathrm{PbO}\), \(\mathrm{WO_{3}}\) & Four phases; weak PbO and specimen broadening. \\
    \bottomrule
  \end{tabularx}
\end{table}

The generator makes symmetry-preserving cell and coordinate perturbations to the registered structures, then calculates their mass-normalized GSAS-II profiles. Their weighted sum is scaled to the selected peak count, a positive background is added, and independent Poisson counts are drawn once from this noise-free mean. Instrument-profile parameters are shared within a measurement session; angular zero, background, specimen effects, and phase-specific effects use separate seeded random streams. Table~\ref{tab:multiphase-generator} gives the frozen laboratory settings. These are synthetic-distribution parameters, not calibrated bounds for experimental refinement. Case c01 fixes the angular offset to zero and disables specimen effects, using a \(1~\mu\mathrm{m}\) domain size and \(U_{\mathrm{iso}}=0.005~\text{\AA}^2\); instrument uncertainty and structural perturbations remain.

\begin{table}[htbp]
  \centering\small
  \setlength{\tabcolsep}{4pt}
  \renewcommand{\arraystretch}{1.12}
  \caption{\textbf{Frozen synthetic-generator settings.} \(\mathcal U[a,b]\) and \(\mathcal L[a,b]\) denote uniform and log-uniform draws. Probabilities apply per phase for strain and texture, and per scan for roughness and background. Instrument coefficients use GSAS-II's native parameter convention. Control recipes can disable effects or specify a background family.}
  \label{tab:multiphase-generator}
  \begin{tabularx}{\textwidth}{@{}p{0.24\textwidth} X@{}}
    \toprule
    Component & Setting \\
    \midrule
    Radiation and geometry & Cu K\(\alpha_1/\alpha_2\): \(1.540598/1.544426~\text{\AA}\), intensity ratio \(I_2/I_1=0.5\); Bragg--Brentano reflection, radius \(200~\mathrm{mm}\). \\
    Angular grid & \(9.90343322^\circ\)--\(100.09455322^\circ\) in \(2\theta\), step \(0.0108664^\circ\). \\
    Count level and noise & Curator-selected signal maximum of 5,000, 15,000, or 40,000 counts, before background; independent Poisson noise on signal plus background. \\
    Instrument profile & \(U\sim\mathcal U[0,40]\), \(W\sim\mathcal U[2,40]\), \(X\sim\mathcal L[0.1,6]\); \(V=Y=Z=0\). \(SH/L\sim\mathcal L[0.005,0.10]\), polarization \(\sim\mathcal U[0.5,1]\). \\
    Angular corrections & Zero offset \(\sim\mathcal U[-0.25,0.25]^\circ\); specimen displacement \(\sim\mathcal U[-800,800]~\mu\mathrm{m}\). \\
    Size and thermal motion & Crystallite size \(\sim\mathcal L[0.04,1.5]~\mu\mathrm{m}\); per-phase \(U_{\mathrm{iso}}\sim\mathcal L[0.003,0.02]~\text{\AA}^2\). \\
    Microstrain & Present with probability 0.6, magnitude \(\mathcal L[300,8000]\) ppm. Conditional on strain, probability 0.4 of uniaxial anisotropy with axial/equatorial ratio \(\mathcal L[0.25,4]\), along [001], [100], or [010]. \\
    Preferred orientation & Probability 0.35; March--Dollase ratio \(\mathcal L[0.7,1.4]\), along [001], [100], [010], [110], or [111]. \\
    Surface roughness & Probability 0.5; Suortti parameters \(A\sim\mathcal U[0.05,0.6]\), \(B\sim\mathcal L[0.02,0.5]\), rejecting draws with an intensity deficit above 50\% at \(2\theta=20^\circ\). \\
    Structural perturbations & Maximum principal cell strain \(\mathcal L[0.0002,0.005]\); requested coordinate RMS \(\mathcal L[0.01,0.04]~\text{\AA}\), with oxygen-specific \(\mathcal L[0.01,0.05]~\text{\AA}\). Symmetry and occupied composition are preserved. \\
    Background family & Simple polynomial, exponential tail, or power-law tail, with probabilities 0.1, 0.6, and 0.3 when selected automatically. \\
    Background amplitudes & Correlated Bernstein floor coefficients in 200--1,200 counts; with probability 0.15, 2,500--3,500 counts. Tail amplitude \(\mathcal U[1000,3000]\) counts, or \(\mathcal U[0,700]\) for the high-floor branch. \\
    Background shape & Simple floor degree 0--2; quadratic floor for either tail family. Tail decay length \(\mathcal L[5,20]^\circ\); power-law exponent \(\mathcal U[1,3]\). \\
    \bottomrule
  \end{tabularx}
\end{table}

The tail functions are \(\exp[-(x-x_{\min})/\ell]\) and \([1+(x-x_{\min})/\ell]^{-p}\), where \(x=2\theta\), \(\ell\) is the decay length, and \(p\) the power-law exponent. Each frozen case retains the recipe, realized parameters, perturbed CIFs, clean profile, noisy observation, and source hashes. Generation checks record known-phase fraction recovery and the residual from omitting each phase. The diagnostic strict-recovery screen uses a maximum absolute fraction error of 0.03 and an omission residual of at least 0.25 noise units. These checks describe quantifiability under the generating model; they are separate from solver scoring and do not establish a unique structural solution.

\FloatBarrier
\paragraph{Profile score and project verification.}
The frozen scorer is version 7.0.1. It reopens \path{final.gpx} and calculates the profile on the original observation grid, retaining the submitted structures, scales, background, instrument calibration, and specimen parameters. CIFs and reported fractions must agree with the project within the recorded numerical and export-rounding tolerances. Saved fit exclusions do not remove scoring bins, and the evaluator optimizes no parameters. Only after this prediction is fixed does it read the hidden generating mean.

Let \(\mu_b\) be the noise-free generating count, including background, and \(\hat y_b\) the replayed prediction at bin \(b\), for \(B\) bins. The profile error and combined training score are
\begin{equation}
  D=\frac{2}{B}\sum_{b=1}^{B}
      \left[\hat y_b-\mu_b+\mu_b\log\!\left(\frac{\mu_b}{\hat y_b}\right)\right],
  \qquad
  Q=100\left[\frac{0.6}{1+D}+0.4S\right],
  \label{eq:multiphase-synthetic-score}
\end{equation}
where \(S\in[0,1]\) is the structural score defined below, and \(0\log 0=0\). A zero prediction at a bin with positive \(\mu_b\) gives \(D=\infty\). Near the generating mean, \(D\) approximates the mean squared residual in Poisson-noise units; \(D=1\) earns half of the profile credit. Scoring against \(\mu_b\) rather than the particular noisy observation discourages fitting counting noise. Observed-pattern \(R_{\mathrm{wp}}\) is recorded separately as a diagnostic.

\paragraph{Structural score.}
The structural term uses continuous agreement between phase inventories, atomic models, and mass fractions, rather than a binary StructureMatcher match. For each reference/submission pair \((i,j)\), the scorer enumerates compatible common cells and searches origin shifts, using a 20\% lattice-length tolerance and \(5^\circ\) angle tolerance. Same-element atomic populations are matched subject to their occupancy capacities. Periodic distances are measured in the mean of the two lattice metrics. Define \(K_r(z)=[1+(z/r)^2]^{-1}\). For transported population \(t_\ell\) of element \(e_\ell\), its normalized coefficient is
\begin{equation}
  \begin{gathered}
  c_\ell =
  \frac{Z_{e_\ell}t_\ell}
       {\sum_e Z_e\max(N^{\mathrm{ref}}_e,N^{\mathrm{sub}}_e)},\\[4pt]
  q_{ij}=K_{0.03}(\epsilon_{ij})\sum_\ell c_\ell K_{0.15\,\text{\AA}}(d_\ell),
  \qquad
  A_{ij}=\sum_\ell c_\ell K_{0.03\,\text{\AA}}(d_\ell).
  \end{gathered}
  \label{eq:multiphase-structural-pair}
\end{equation}
Here \(Z_e\) is atomic number, \(N_e\) the occupied population in the registered common cell, \(d_\ell\) the corresponding periodic distance, and \(\epsilon_{ij}\) the root-mean-square logarithmic lattice strain. Atomic-number weighting is a proxy for X-ray scattering sensitivity. The registration search maximizes \(q_{ij}(0.5+0.5A_{ij})\) over the tested cell/origin mappings and population transports. It is a finite deterministic search; unfinished registrations retain explicit score bounds.

For reference mass fractions \(w_i\) and submitted fractions \(v_j\), let \(g(w)=1-\exp(-w/0.05)\). For a fully quantified submission, the inventory, mass-overlap, and fraction-precision credits are
\begin{equation}
  I_{ij}=\frac{2q_{ij}\min\{g(w_i),g(v_j)\}}
                    {\sum_a g(w_a)+\sum_b g(v_b)},\qquad
  M_{ij}=\frac{q_{ij}\min(w_i,v_j)}{\max(1,\sum_b v_b)},\qquad
  F_{ij}=K_{0.05}(w_i-v_j).
  \label{eq:multiphase-inventory-credit}
\end{equation}
The final structural score maximizes total credited agreement over one-to-one phase assignments \(\mathcal A\):
\begin{equation}
  S=\max_{\mathcal A}\sum_{(i,j)\in\mathcal A}
      \left(0.5I_{ij}+0.5M_{ij}\right)
      \left(0.5+0.5A_{ij}\right)
      \left(0.75+0.25F_{ij}\right).
  \label{eq:multiphase-structural-score}
\end{equation}
The inventory term gives credit to weak phases while penalizing additional unmatched assertions; the mass term emphasizes abundant phases. Site and fraction precision modulate this recovered-inventory credit. Lattice agreement enters \(q_{ij}\), with no additional fine-scale lattice-precision term.

\paragraph{Incomplete submissions and physical checks.}
Numerically duplicate phase exports are merged. An invalid individual model remains an unmatched assertion, while valid submitted structures retain their credit. An unknown fraction receives inventory credit with \(g(v_j)=1\), but no mass or fraction-precision credit. Explicitly unresolved material is not renormalized away. Without a complete verified project, the profile contribution is zero and independently evaluable structural credit remains. Malformed top-level submissions and hard physical failures score zero; examples of the latter include invalid occupancies and fully occupied interatomic contacts below \(0.8~\text{\AA}\). Close O--O contacts below \(1.8~\text{\AA}\), partially occupied short contacts, and questionable profile parameters trigger review rather than automatic rejection. Reported means retain these provisional scores and scientific failures. Evaluator errors remain distinct from solver failures, and unresolved structural registration is reported as an interval.

\paragraph{Evolution, promotion, and stopping rules.}
The evolver uses the same GPT-5.6 Sol model alias and maximum reasoning effort as the solvers. It can inspect synthetic ground truth, trajectories, scripts, and saved projects, and run diagnostic calculations or procedure tests on completed attempts. Such oracle-assisted tests inform package design but are recorded separately from fresh solver evaluations. A package contains at most 600 words of guidance plus optional executable helpers; every evaluated revision is frozen before launch.

The original run followed the written procedure below. The controller enforced freezing, access boundaries, and solve-budget accounting; promotion was an explicit evolver decision with a saved rationale, not an automatic statistical test.
\begin{enumerate}
  \item \textbf{Baseline.} Curate six cases and collect three independent baseline attempts per case.
  \item \textbf{Screen and revise.} Run one attempt per case with a new package. Revise on unresolved cases, temporarily setting aside successful cases. The operational target was the intended scientific behavior and scores around 80 or higher. Before repeated evaluation, test one unchanged package across all six cases, including the set-aside cases.
  \item \textbf{Evaluate and promote.} Complete the promising package to three attempts per case, including its screening attempt. The intended success condition was most attempts near the noise floor, most case means above 80, and every case mean above baseline. At the 30-development-solve limit, complete the current or most promising package to three attempts per case and retain it if it improves on the baseline or incumbent, carrying unresolved cases forward.
  \item \textbf{Test transfer.} Add new cases to restore a six-case active panel. On each new case, compare the unchanged baseline and incumbent with three attempts each before using those results for further adaptation. Report these new-case comparisons separately from carried-over cases.
\end{enumerate}

The 30-solve limit counts nonbaseline development attempts when queued. Baseline attempts, prospective transfer comparisons, and the second and third attempts used to complete a package's three-repeat evaluation are excluded. Once the limit is reached, only one package can be completed to three attempts per already-attempted case in that cycle. Completion of the next prospective comparison starts a new cycle. There is no fixed total-run solve budget: the outer loop continues until stopped by the operator or interrupted by service or infrastructure limits. The later framework-access campaign's numerical promotion gate is a separate protocol, specified in Appendix~\ref{app:multiphase-framework-evolution}.

\paragraph{Retained package and recorded outcomes.}
The archived run contains 135 completed solver attempts on ten synthetic cases and 15 revisions; v6 was the only promoted package. Its 331 words of guidance and prototype-search helper address source selection, missing-reference recovery, reconsideration of weak phases, instrument/specimen modeling, and saved-project checks. On the initial six cases, its three-repeat mean score was 74.07 versus 50.04 for the baseline, with all six case means improved and complete generating-phase recovery in 17 of 18 attempts. Only two case means exceeded 80, so the evolver promoted v6 under the budget-exhaustion rule. The next two two-case prospective comparisons improved from 58.79 to 72.44 and from 43.02 to 68.97. No later revision replaced v6. The package text and helper description appear in Appendix~\ref{app:multiphase-framework-evolution}.

\paragraph{Reproduction records.}
The frozen run is retained under \path{results/experimental-evolution/divalent-tungstates/} in the research repository. \path{evolve/LAB.md} and \path{runtime/generator_defaults.json} define the generator; batch manifests record selected phases, fractions, seeds, and withholding rules; \path{scorer/config.json} and its source files define score version 7.0.1. \path{evolve/PROMPT.md}, \path{evolve/settings.json}, and \path{control/progression.py} record the evolution instructions and budget accounting. Case records, package hashes, solver outputs, scores, promotion rationale, and evolver-written investigation notes preserve the executed run. The measured-pattern evaluation and its limitations are described in Appendices~\ref{app:multiphase-experimental-results} and~\ref{app:multiphase-framework-evolution}.

\FloatBarrier

\subsection{Experimental profile comparisons and per-attempt results}
\label{app:multiphase-experimental-results}

\paragraph{Profile evaluation.}
The experimental comparisons use the same measured counts, evaluation window, and residual definition for each solution. Agent profiles are calculated by reopening the submitted GSAS-II project without further optimization. The replay retains the agent's phase models, fractions, background, instrument parameters, and specimen effects. Exported profiles and reported scalar residuals are checked against this calculation. Deposited PG profiles are taken from the saved calculated curves accompanying each measurement and interpolated onto the observation grid where necessary. They are rescored directly, without a new refinement or conversion to an agent's model. All residuals use
\begin{equation}
  R_{\mathrm{wp}}=100\sqrt{
    \frac{\sum_{i\in\mathcal I} w_i(y_i-\hat y_i)^2}
         {\sum_{i\in\mathcal I} w_i y_i^2}},
  \qquad w_i=\frac{1}{\max(y_i,1)},
  \label{eq:multiphase-observed-rwp}
\end{equation}
where \(y_i\) and \(\hat y_i\) are the measured and calculated counts, respectively, and \(\mathcal I\) contains points from \(10.1^\circ\) to \(99.9^\circ\) in \(2\theta\). The calculated intensity includes the fitted background. This common residual measures agreement with the observation; the fitted model and its flexibility remain specific to each solution.

For aggregate scores, we average the three attempts within each measurement, then calculate the mean or median of those 15 values. All 45 attempts in each of the five Sol-max conditions have verified profiles. The additional Sr follow-ups are reported separately in Table~\ref{tab:multiphase-recovery}.

\begin{figure}[tbp]
  \centering
  \includegraphics[width=\textwidth]{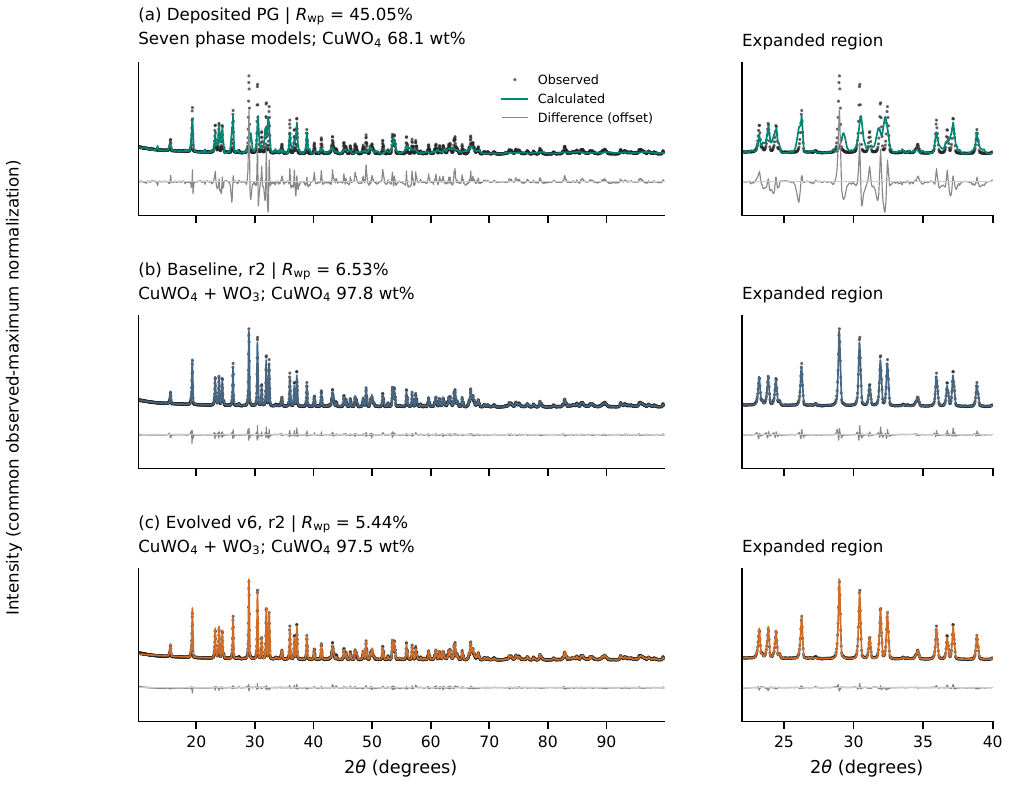}
  \caption{\textbf{Deposited and agent solutions for the Cu measurement (006).} The deposited PG calculated profile is compared with the median-residual attempt in each agent condition (r2 for both baseline and v6). The left panels show the full evaluation window and the right panels expand \(22^\circ\)--\(40^\circ\). Observations are black points, calculations are colored lines, and differences are gray with a vertical offset. All rows share the observed maximum as a display normalization and identical vertical limits within each column. Residuals use the original counts over the full evaluation window. PG's saved calculated curve is rescored; agent curves are independently replayed from their final projects. No additional fitting is performed for this figure.}
  \label{fig:multiphase-cu-deposited}
\end{figure}

\paragraph{Cu: a simpler phase inventory with better profile agreement.}
Figure~\ref{fig:multiphase-cu-deposited} compares the PG deposit with baseline and v6 solutions for case 006. The deposited model contains \(\mathrm{CuWO_{4}}\) (68.1 wt\%), \(\mathrm{CuO}\) (9.9\%), \(\mathrm{Cu_{7}(W_{3}O_{19})_{4}}\) (6.3\%), \(\mathrm{Cu_{2}WO_{4}}\) (5.9\%), \(\mathrm{WO_{2}}\) (3.6\%), and two \(\mathrm{WO_{3}}\) models (3.2\% and 3.1\%). Each of the six baseline/v6 attempts instead submits two models, \(\mathrm{CuWO_{4}}\) and \(\mathrm{WO_{3}}\), with 96.8--97.9 wt\% assigned to \(\mathrm{CuWO_{4}}\). The agent interpretations reproduce the measured profile more closely under the shared residual definition. The disagreement in phase inventory and fractions warrants independent experimental review, particularly because the deposited and agent analyses use different fitted models.

\paragraph{All initial attempts and model/compute comparisons.}
Table~\ref{tab:multiphase-all-attempts} reports every initial full-panel attempt for the five Sol-max conditions in Table~\ref{tab:multiphase-conditions}, the PG deposits, and two earlier comparisons using the baseline task without learned guidance. The Sol-high comparison uses \texttt{gpt-5.6-sol} at high reasoning effort in Codex 0.154, with three completed attempts per measurement. The Opus comparison uses \texttt{claude-opus-5} at high effort in Claude Code 2.1.299, with one attempt per measurement. These runs use the same measured-input hashes, filtered COD library, scientific software, two-CPU/five-GiB worker allocation, and no solver time limit. The Sol-high panel includes a documented replacement for a user-cancelled case-015 r3 that produced no complete submission; the replacement is marked in the table. Ten Opus sessions were resumed after a subscription usage limit, preserving their workspaces and conversation state.

Mean residuals are 8.98\% for Sol-high, 7.90\% for the main Sol-max baseline, and 6.32\% for Opus. Recorded mean runtimes are 13.9, 27.0, and 86.5 minutes per completed attempt, respectively. The higher-effort Sol condition and the Opus model/harness combination therefore provide useful performance comparisons, with different compute use and execution dates; the reasoning-effort labels are not calibrated across providers. All Sol-high profiles pass the project-replay check. Opus uses a declared secondary grid tolerance of \(10^{-6}\) degrees, accommodating \(2\theta\) coordinates exported to six decimal places; all 15 profiles pass, compared with four at the stricter \(10^{-7}\)-degree tolerance. Measured counts and sample counts must still agree, and the residual formula and evaluation window are unchanged.

The Opus Sr solution illustrates why a low residual need not imply phase recovery. It reaches \(R_{\mathrm{wp}}=9.74\%\) while representing 12 unresolved reflections as independent background peaks, without recovering the \(\mathrm{Sr_{2}WO_{5}}\) structure. Its reported two-phase fit without those peaks has a residual of approximately 15.8\%. This comparison provides context for another model/harness combination rather than isolating model strength.

\FloatBarrier
% Generated by scripts/plot_experimental_transfer.py.
\newcommand{\experimentalTransferAllAttemptRows}{%
001 (BaCO$_{3}$) & 1 & 8.00 & \textbf{8.73} & 7.22 & \textbf{7.15} & \textbf{5.85} & 6.87 & 6.67 & 7.30 \\*
 & 2 &  & 8.73 & 7.22 & 7.23 & 6.08 & \textbf{6.65} & \textbf{5.83} &  \\*
 & 3 &  & 8.82 & \textbf{7.15} & 7.58 & 5.92 & 6.79 & 5.96 &  \\
\addlinespace[3pt]
002 (BaO$_{2}$) & 1 & 7.20 & 8.87 & 7.59 & 8.73 & 8.51 & 7.06 & \textbf{7.12} & 8.30 \\*
 & 2 &  & \textbf{7.68} & 12.67 & 7.65 & 7.24 & \textbf{5.78} & 7.21 &  \\*
 & 3 &  & 7.80 & \textbf{7.12} & \textbf{7.10} & \textbf{7.21} & 8.29 & 7.60 &  \\
\addlinespace[3pt]
003 (CaCO$_{3}$) & 1 & 18.28 & \textbf{9.00} & 7.79 & 8.09 & 7.31 & 7.37 & 7.34 & 7.23 \\*
 & 2 &  & 9.28 & 8.39 & 7.88 & 7.26 & 7.76 & 7.45 &  \\*
 & 3 &  & 10.47 & \textbf{7.48} & \textbf{7.56} & \textbf{7.09} & \textbf{5.30} & \textbf{7.34} &  \\
\addlinespace[3pt]
004 (CoO) & 1 & 6.37 & 6.79 & 7.36 & 5.66 & \textbf{4.72} & 4.96 & 5.63 & 5.36 \\*
 & 2 &  & \textbf{6.41} & 6.96 & 4.95 & 6.19 & \textbf{4.77} & 5.25 &  \\*
 & 3 &  & 7.07 & \textbf{6.12} & \textbf{4.90} & 4.82 & 5.21 & \textbf{5.08} &  \\
\addlinespace[3pt]
005 (Co$_{3}$O$_{4}$) & 1 & 3.78 & 4.20 & \textbf{3.37} & 3.37 & 3.42 & 3.34 & \textbf{3.06} & 2.79 \\*
 & 2 &  & 3.61 & 3.56 & 3.36 & 3.13 & \textbf{3.07} & 3.19 &  \\*
 & 3 &  & \textbf{3.42} & 3.44 & \textbf{3.22} & \textbf{3.07} & 3.26 & 3.17 &  \\
\addlinespace[3pt]
006 (CuO) & 1 & 45.05 & \textbf{7.01} & \textbf{6.48} & 6.46 & \textbf{3.51} & 6.89 & 5.14 & 3.78 \\*
 & 2 &  & 7.37 & 6.53 & 5.44 & 3.89 & 6.88 & 7.19 &  \\*
 & 3 &  & 7.17 & 7.39 & \textbf{4.30} & 3.88 & \textbf{6.81} & \textbf{4.17} &  \\
\addlinespace[3pt]
007 (MgO) & 1 & 13.12 & 10.18 & 9.68 & 9.31 & \textbf{5.73} & \textbf{5.73} & 8.35 & 8.24 \\*
 & 2 &  & 13.21 & 8.91 & 6.74 & 8.29 & 5.81 & \textbf{7.64} &  \\*
 & 3 &  & \textbf{9.88} & \textbf{8.42} & \textbf{6.04} & 5.83 & 7.52 & 8.34 &  \\
\addlinespace[3pt]
008 (MnO) & 1 & 10.22 & \textbf{6.60} & 8.45 & \textbf{4.49} & 3.98 & 6.16 & \textbf{4.03} & 3.98 \\*
 & 2 &  & 9.12 & 7.07 & 6.36 & 3.92 & 4.47 & 6.37 &  \\*
 & 3 &  & 7.23 & \textbf{6.75} & 6.75 & \textbf{3.85} & \textbf{3.91} & 4.25 &  \\
\addlinespace[3pt]
009 (Mn$_{3}$O$_{4}$) & 1 & 11.38 & 12.26 & 5.23 & 5.30 & \textbf{3.74} & 5.30 & 5.02 & 4.95 \\*
 & 2 &  & 9.63 & \textbf{5.09} & 5.42 & 3.81 & \textbf{4.10} & \textbf{3.95} &  \\*
 & 3 &  & \textbf{7.31} & 5.27 & \textbf{5.06} & 4.30 & 5.03 & 4.97 &  \\
\addlinespace[3pt]
010 (Mn$_{2}$O$_{3}$) & 1 & 8.27 & 8.49 & 7.46 & 7.41 & \textbf{5.89} & 6.46 & 6.80 & 6.67 \\*
 & 2 &  & \textbf{6.71} & 6.86 & 7.14 & 6.09 & 6.85 & 6.25 &  \\*
 & 3 &  & 7.10 & \textbf{6.83} & \textbf{6.92} & 6.24 & \textbf{6.17} & \textbf{6.13} &  \\
\addlinespace[3pt]
011 (MnO$_{2}$) & 1 & 7.43 & 8.37 & 6.12 & 5.84 & 5.49 & 5.79 & 5.57 & 6.84 \\*
 & 2 &  & 7.81 & 6.21 & \textbf{5.73} & 5.56 & \textbf{5.42} & 6.39 &  \\*
 & 3 &  & \textbf{7.81} & \textbf{5.99} & 5.88 & \textbf{5.43} & 5.78 & \textbf{5.54} &  \\
\addlinespace[3pt]
012 (NiO) & 1 & 6.74 & 5.17 & \textbf{4.31} & 4.57 & 4.14 & \textbf{4.19} & 4.10 & 4.30 \\*
 & 2 &  & \textbf{4.93} & 4.59 & 4.79 & \textbf{4.08} & 4.21 & 4.13 &  \\*
 & 3 &  & 6.40 & 4.36 & \textbf{4.48} & 4.19 & 4.29 & \textbf{4.04} &  \\
\addlinespace[3pt]
013 (PbO) & 1 & 11.54 & 16.06 & 14.70 & \textbf{11.70} & 11.47 & 10.52 & 9.72 & 10.61 \\*
 & 2 &  & 18.35 & 12.58 & 13.01 & \textbf{8.02} & \textbf{10.47} & \textbf{8.84} &  \\*
 & 3 &  & \textbf{12.99} & \textbf{10.25} & 12.97 & 8.29 & 13.13 & 9.09 &  \\
\addlinespace[3pt]
014 (SrCO$_{3}$) & 1 & 15.81 & 18.13 & \textbf{12.89} & 14.66 & 14.65 & 15.85 & \textbf{6.64} & 9.74 \\*
 & 2 &  & \textbf{14.70} & 16.48 & \textbf{7.12} & \textbf{8.98} & 13.04 & 15.23 &  \\*
 & 3 &  & 15.70 & 16.72 & 7.88 & 11.73 & \textbf{7.53} & 15.59 &  \\
\addlinespace[3pt]
015 (ZnO) & 1 & 15.70 & \textbf{6.38} & \textbf{5.49} & \textbf{4.82} & 6.21 & 5.58 & 6.96 & 4.74 \\*
 & 2 &  & 11.72 & 11.25 & 5.89 & \textbf{4.89} & \textbf{5.26} & \textbf{5.33} &  \\*
 & 3 &  & 9.28$^{\dagger}$ & 13.64 & 5.00 & 5.16 & 6.27 & 6.24 &  \\
\midrule
\multicolumn{2}{l}{Mean} & 12.59 & 8.98 & 7.90 & 6.71 & 5.98 & 6.49 & 6.44 & 6.32 \\
\multicolumn{2}{l}{Median} & 10.22 & 8.12 & 7.20 & 5.86 & 5.49 & 6.36 & 6.15 & 6.67 \\
}

\begingroup
\footnotesize
\setlength{\tabcolsep}{2.5pt}
\renewcommand{\arraystretch}{1.06}
\begin{longtable}{@{}lccccccccc@{}}
  \caption{\textbf{All initial experimental profile scores, \(R_{\mathrm{wp}}\) (\%).} Bold marks the lowest residual among three attempts within each condition and case, before rounding. Baseline and the four adapted/context conditions use Sol-max. Single PG and Opus scores appear only on the first row of each case. Aggregates are over measurement-level means. Every reaction also includes \(\mathrm{WO_{3}}\). \(^{\dagger}\)Documented replacement for a user-cancelled attempt.}
  \label{tab:multiphase-all-attempts}\label{tab:multiphase-scan-means}\\
  \toprule
  Case & Run & PG & \shortstack{Sol\\high} & \shortstack{Baseline\\Sol-max} & \shortstack{Evolved\\v6}
    & \shortstack{Framework\\access} & \shortstack{Framework\\+ v6}
    & \shortstack{Evolved\\framework\\v5} & Opus \\
  \midrule
  \endfirsthead
  \multicolumn{10}{l}{\tablename~\thetable\ (continued)}\\
  \toprule
  Case & Run & PG & \shortstack{Sol\\high} & \shortstack{Baseline\\Sol-max} & \shortstack{Evolved\\v6}
    & \shortstack{Framework\\access} & \shortstack{Framework\\+ v6}
    & \shortstack{Evolved\\framework\\v5} & Opus \\
  \midrule
  \endhead
  \midrule
  \multicolumn{10}{r}{Continued on next page}\\
  \endfoot
  \bottomrule
  \endlastfoot
  \experimentalTransferAllAttemptRows
\end{longtable}
\endgroup

\paragraph{Reproduction.}
The manuscript's \path{scripts/plot_experimental_transfer.py} builds the comparison figures and tables from compact snapshots of the saved evaluation records. These snapshots retain per-attempt residuals, reported phases and fractions, and source-file hashes; the Cu profile arrays are included separately. \path{data/experimental_transfer/model_comparison.json} retains the earlier Sol-high and Opus scores, checked against the archived replay outputs and comparison report, including the replacement's provenance. The script checks plotted residuals against the saved evaluation results when refreshing inputs and launches no solving or refinement jobs. Full trajectories, projects, and evaluator outputs are retained under \path{results/experimental-eval/divalent-tungstates-PG15/} in the research repositories.

\FloatBarrier
\subsection{Additional multiphase controls and evolution records}
\label{app:multiphase-framework-evolution}
\label{sec:multiphase-supporting-notes}

\subsubsection{Framework-access evolution protocols}

Both campaigns started from framework access and retained it as the incumbent. They used different synthetic curricula and acceptance rules. The earlier campaign, \path{divalent-tungstates-fw}, used selective single-attempt screening, as in the original baseline-evolution run. After three baseline attempts on each of six cases, the evolver screened v1 once on every case, then revised the package on unresolved cases. A score of approximately 80 or higher, together with the intended scientific behavior, qualified a case to be set aside. Before promotion, a single package was to complete a full-panel screen followed by two additional attempts per case; at the 30-development-solve limit, the most promising package was to receive the repeated evaluation and be promoted if it improved on the baseline. This run stopped before reaching either route to promotion.

The run completed 18 baseline and 17 revision attempts. v1 passed two cases; v2 resolved two more; and v3 resolved a fifth. The remaining case, containing a withheld \(\mathrm{Ca_{3}WO_{6}}\) intermediate, received revisions through v8. The best of these single attempts scored 70.09 under the early run's v5. v7 failed to produce final deliverables and receives zero; v9 was stopped before scoring. Scores from successful screens of different versions were not pooled to claim a validated package. The early v5 is a different artifact from the later \emph{Evolved framework v5} tested on experimental patterns.

The later campaign, \path{divalent-tungstates-access}, replaced screening with three attempts per case for every candidate package. Each active panel contained six cases. The evolver could test up to three versions before forming the next panel, carrying forward unresolved cases and adding new problems. Ten distinct synthetic cases were baselined, requiring 30 attempts. Seven revisions required a further \(7\times6\times3=126\) attempts. The active panel changed after v3 and after v6.

For each revision, the controller compared its mean score with the framework baseline on the same cases. Promotion required
\begin{equation}
  \Delta\bar Q\geq\max(4,2\,\mathrm{SE}),
  \qquad \min_c\Delta\bar Q_c\geq-15,
  \label{eq:framework-promotion}
\end{equation}
with three scored attempts per condition on at least six cases. Here \(\Delta\bar Q_c\) is the difference between case means and \(\mathrm{SE}\) is the controller's estimate from pooled within-case replicate variance. With six cases and three attempts per condition, \(\mathrm{SE}^2=\sum_c(s_{c,\mathrm{candidate}}^2+s_{c,\mathrm{baseline}}^2)/(3\times6^2)\), where each \(s^2\) is a sample variance. This is an operational promotion criterion, not a multiple-testing-adjusted significance test.

\begin{table}[htbp]
  \centering\small
  % Generated by scripts/plot_framework_evolution.py.
\newcommand{\frameworkEvolutionRows}{%
v1 & -11.81 & 10.13 & -39.66 \\
v2 & -3.55 & 13.94 & -32.62 \\
v3 & +4.16 & 11.95 & -21.83 \\
v4 & -6.68 & 8.46 & -38.34 \\
v5 & +2.96 & 10.06 & -37.62 \\
v6 & -6.04 & 8.43 & -35.93 \\
v7 & [-4.45, -4.41] & [10.80, 10.81] & -20.57 \\
}

  \caption{\textbf{Later framework-access evolution.} Each row compares 18 candidate attempts with baseline attempts on the same six cases. Positive changes improve the synthetic score. No version satisfies either the required mean gain or the permitted worst-case regression.}
  \label{tab:multiphase-fw-versions}
  \begin{tabular}{@{}lrrr@{}}
    \toprule
    Version & Mean change & Required mean gain & Worst case change \\
    \midrule
    \frameworkEvolutionRows
    \bottomrule
  \end{tabular}
  \begin{minipage}{\textwidth}\footnotesize
  \vspace{4pt}
  One v7 attempt has a bounded structural-registration score. Its interval is propagated through the mean and estimated threshold; Figure~\ref{fig:multiphase-framework-evolution} plots their midpoints. The regression limit is violated throughout the interval. The live controller leaves that case unqualified for promotion until registration is resolved. Provisional physical-review scores are retained, and invalid submissions remain in the denominator.
  \end{minipage}
\end{table}

\begin{figure}[tbp]
  \centering
  \includegraphics[width=\textwidth]{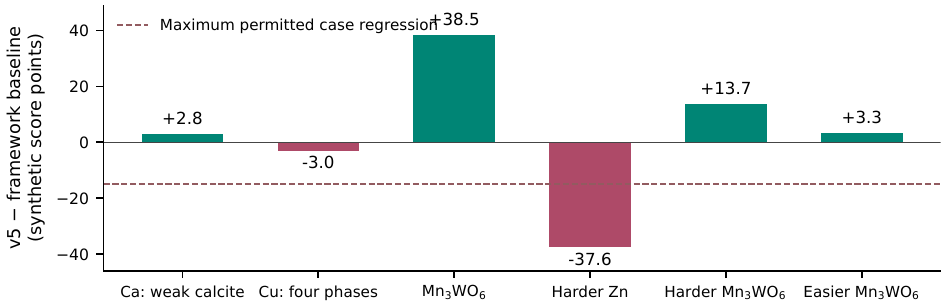}
  \caption{\textbf{Case-level gains and regressions for the later framework v5.} Bars compare three v5 attempts with three framework-baseline attempts. The mean improvement of 2.96 score points combines a 38.50-point gain on a missing-\(\mathrm{Mn_{3}WO_{6}}\) case with a 37.62-point loss on the harder Zn problem. The dashed line marks the permitted case-regression limit.}
  \label{fig:multiphase-v5}
\end{figure}

The later v5 contains 205 words of guidance and no added helper. It prioritizes a complete physical checkpoint, distinguishes numerical failure from scientific rejection, permits cross-chemistry transfer for chemically supported unassigned reflections, and stops further library search once the inventory explains the coherent reflection families. Its subsequent refinement instructions cover coordinates, specimen effects, and phase-profile terms. Figure~\ref{fig:multiphase-v5} shows why its positive aggregate change was insufficient for promotion.

\paragraph{Reproduction.}
\path{scripts/plot_framework_evolution.py} rebuilds the two-campaign figure and version table from the manuscript's compact job-level snapshot, retaining score intervals, invalid submissions, and the unscored v9 attempt. Source hashes identify the run status, evolver instructions, frozen scorer configuration, and later promotion-rule implementation. Full trajectories, package revisions, procedure tests, and evolver-written diagnoses are retained in the corresponding run directories. The two campaigns have different cases and budgets; their version numbers do not identify comparable stages on a shared learning curve.

\subsubsection{Contents of the original v6 package}

The 331-word v6 package combines the following guidance with one formula-seeded cross-chemistry prototype-search helper. The helper accepts a formula, reference index and CIF root, with optional crystallographic filters. The command invocation and Markdown formatting are abbreviated below; the released package records the exact text and tool implementation.

\begin{quote}\small
\textbf{Build and revisit the inventory.}
Screen occupied composition, symmetry, source conditions, disorder, and geometry before fitting. A poor fit or zero scale rejects that record, not a chemically plausible phase: compare a physical alternate record or polymorph under the same restrained nuisance model. Keep plausible weak phases supported by coherent unique or partly overlapped families on a watch list.

If no exact entry exists, search cross-chemistry prototypes with the stoichiometric-analogue helper. Optionally add an indexed space group and cell. Inspect several parents, justify any site substitution, physical-screen the model, and refine scale before cell.

After any phase or source change, reconverge a smooth low-order background, all scales, one angular correction, and cells. Once the provisional inventory is coherent, establish a clean common profile and re-challenge the watch list under it. If inventory changes, rebuild that profile before judging phases.

\textbf{Require a physically adequate baseline.}
Neutralize unused phase size/strain, roughness, and texture. Fit one zero or specimen-displacement correction and a nonnegative low-dimensional U/W/X profile; release SH/L when asymmetry repeats across peaks. Judge adequacy from observed-versus-calculated positions, FWHM, and asymmetry for isolated families of the same phase at low, middle, and high angle, plus a smooth plausible background---not from Rwp, interior values, or covariance alone. If a systematic trend remains, or all phase-width trials collapse or hit bounds, return to the baseline and inventory rather than adding nuisance terms.

Then compare size and strain from neutral values, one phase at a time. Combine only complementary replay-stable terms. If hkl families remain differently broad, challenge symmetry-appropriate anisotropy. Before texture, decide whether any coherent cross-phase intensity trend is background, polarization, or low-angle roughness; reject scale/background surrogates. Compare no texture with symmetry-distinct axial and diagonal low-index axes, requiring support from multiple isolated families.

After each accepted change, save, reopen, and replay the full range. Prune unsupported or bound terms and verify geometry, fractions, and exact project/export agreement.
\end{quote}

Guidance-only and helper-only variants have not been evaluated separately. The observed phase-recovery gain therefore concerns the delivered package as a whole.

\subsubsection{Experimental scope and submitted-model diagnostics}

\begin{figure}[tbp]
  \centering
  \includegraphics[width=\textwidth]{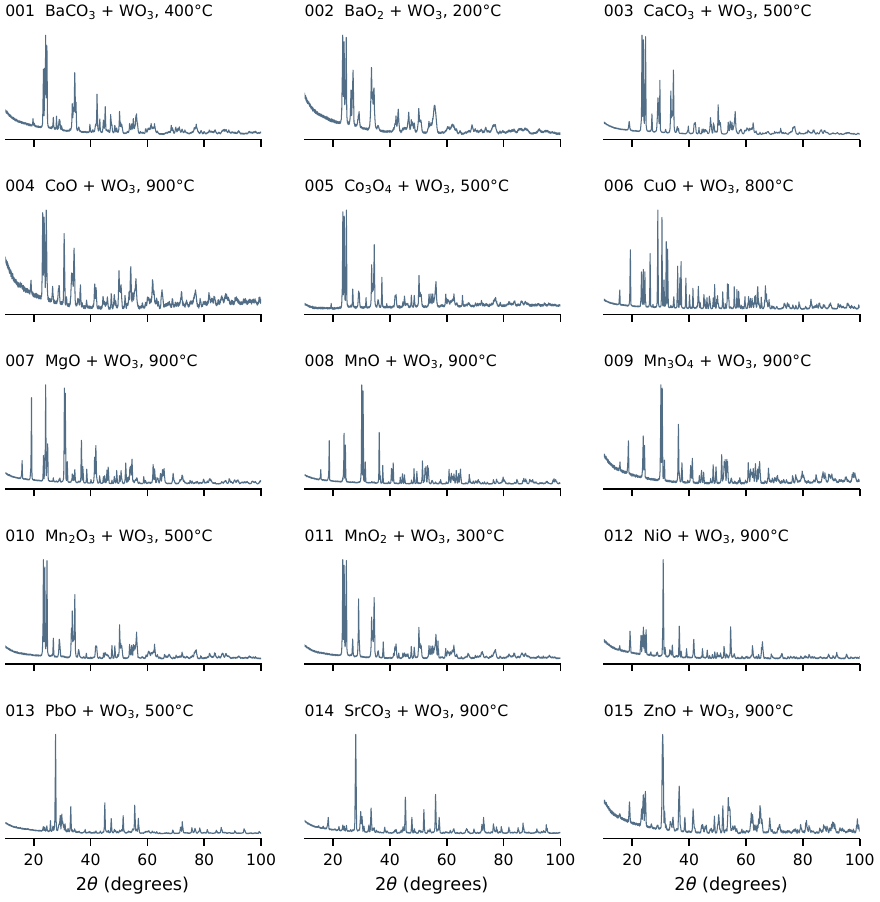}
  \caption{\textbf{Experimental input panel.} Each scan is independently max-normalized for display; solvers receive the original counts. Titles give the precursor pair and firing temperature. Background, peak widths, relative intensities, and phase overlap vary across the measurements. The arrangement does not define a train/test partition.}
  \label{fig:multiphase-patterns}
\end{figure}

The 15 measured patterns form a chemically related family selected from the Precursor Genome ledger (Figure~\ref{fig:multiphase-patterns}). They informed earlier human decisions about the framework and synthetic generator, making this a developmental experimental evaluation. The frozen evolver runs did not receive the measured targets or their deposited solutions. Conditions used a mutable model alias on different dates and consumed different amounts of inference-time compute. Additional Sr repeats were selected after inspecting the initial outcomes and concern the same specimen. These design choices support comparisons on this panel; new specimens and matched-compute controls would test broader transfer and isolate the contributions of context and computation.

The Pb/Sr structural checks compare submitted CIFs with canonical deposited structures using symmetric pymatgen \texttt{StructureMatcher}: \texttt{ltol}=0.2, \texttt{stol}=0.3, and \texttt{angle\_tol}=5 degrees, with primitive reduction and supercell matching enabled, lattice rescaling and subset matching disabled, and elemental comparison. Before matching, each site's occupied species are retained but occupancy magnitudes are replaced by equal fractions summing to one. A match identifies structural recovery within these tolerances; refined occupancies, phase fractions, and the rest of the mixture require separate assessment.

\begin{table}[htbp]
  \centering\small
  \setlength{\tabcolsep}{4pt}
  \caption{\textbf{Diagnostics of the 45 submitted projects per condition.} Background terms are polynomial or interpolation coefficients, excluding fixed-background curves and auxiliary peaks; entries give the median and range. Nondefault \(SH/L\) means a submitted value different from 0.002. Phase counts are means over all attempts.}
  \label{tab:multiphase-diagnostics}
  \begin{tabularx}{\textwidth}{@{}Xccc@{}}
    \toprule
    Condition & Background terms & Nondefault \(SH/L\) & Phases \\
    \midrule
    Baseline & 15 (6--20) & 12/45 & 2.47 \\
    Evolved v6 & 8 (6--12) & 17/45 & 2.91 \\
    Framework access & 16 (10--100) & 44/45 & 2.64 \\
    Framework + v6 & 8 (3--33) & 45/45 & 2.80 \\
    Evolved framework v5 & 12 (8--30) & 45/45 & 2.51 \\
    \bottomrule
  \end{tabularx}
\end{table}

These diagnostics describe final parameter values; parameters may have been fitted earlier and fixed before export. A CIF contact/occupancy screen flags at least one model in 20/45 baseline, 9/45 v6, 9/45 framework-access, 6/45 combined, and 3/45 evolved-framework-v5 attempts. Flags require chemical adjudication: short O--O contacts can be appropriate in a peroxide such as \(\mathrm{BaO_{2}}\), while other recurrent flags arise from coarse coordinates in a supplied \(\mathrm{WO_{3}}\) record. They are retained as diagnostics rather than counted automatically as invalid solutions.

Disagreements with deposited models remain useful questions for experimental reanalysis. Agents assign more \(\mathrm{CaCO_{3}}\) on case 003, favor a simpler \(\mathrm{CuWO_{4}}\)-dominated inventory on case 006, and omit the small deposited \(\mathrm{SrCO_{3}}\) contribution in successful evolved Sr fits. Profile agreement, recovered structures, and chemical plausibility supply complementary evidence for assessing these interpretations.

\bibliographystyle{plainnat}
\bibliography{references}

\end{document}